\documentclass[10pt,twocolumn,letterpaper]{article}

\PassOptionsToPackage{table}{xcolor}  %
  \usepackage[pagenumbers]{paperstyle} %

\makeatletter
\def\abstract{%
   \iftoggle{pstylepagenumbers}{}{\thispagestyle{empty}}%
   \centerline{\large\bf Abstract}%
   \par\vskip 7pt\noindent%
   \it\ignorespaces%
}

\makeatother

\usepackage[utf8]{inputenc}  %
\usepackage[T1]{fontenc}     %
\usepackage{amsfonts}        %
\usepackage{nicefrac}        %
\usepackage{microtype} %
\definecolor{cppgreen}{HTML}{A6D189}
\definecolor{cppteal}{HTML}{81C8BE}
\definecolor{cppgreenteal}{HTML}{8CCBAE}  %
\definecolor{cppsky}{HTML}{A0CCD4}  %
\definecolor{cppyellow}{HTML}{F0CB85}  %
\definecolor{cppflamingo}{HTML}{E9C3C3}  %
\definecolor{cpppeach}{HTML}{EF9F76}  %
\definecolor{cppmaroon}{HTML}{E2A1A3}  %
\definecolor{cpprosewater}{HTML}{F2D5CF}
\definecolor{cppblue}{HTML}{8CAAEE}
\definecolor{cpplavender}{HTML}{BABBF1}
\definecolor{cppgray}{HTML}{B9B9B9}  %
\definecolor{tabgray}{HTML}{D4D4D4}       %
\definecolor{tabgreenteal}{HTML}{94C9B0}  %
\definecolor{tablightblue}{HTML}{BBCEDB}  %
\definecolor{tabapgreen}{HTML}{6FB58F}    %
\definecolor{tabwarmgray}{HTML}{C3C0BA}  %
\definecolor{tabcream}{HTML}{E1DECF}     %
\definecolor{mistralcream}{HTML}{F5F4EF}  %
\definecolor{mistralsand}{HTML}{EBE9E0}   %
\definecolor{tabcreamgray}{HTML}{EAEAE9}  %
\definecolor{tabgreenfig}{HTML}{94CAA5}   %
\definecolor{tabblue}{HTML}{003EAC}      %
\newcommand{\dashedmidrule}[1]{%
  \noalign{\vskip\aboverulesep}%
  \multispan{#1}\vrule width 2pt height 0.4pt depth 0pt\xleaders\hbox to 3.5pt{\hfil\vrule width 2pt height 0.4pt depth 0pt}\hfill\\%
  \noalign{\vskip\belowrulesep}}
\newlength{\methodcol}\newlength{\tmpcolw}
\AtBeginDocument{%
  \settowidth{\methodcol}{\scriptsize Gemini Robotics ER 2}%
  \settowidth{\tmpcolw}{\scriptsize WildDet3D + sensor depth}\ifdim\tmpcolw>\methodcol\setlength{\methodcol}{\tmpcolw}\fi
  \settowidth{\tmpcolw}{\scriptsize WildDet3D + UniDepthV2}\ifdim\tmpcolw>\methodcol\setlength{\methodcol}{\tmpcolw}\fi
  \settowidth{\tmpcolw}{\scriptsize Claude Opus 5.5 + distance}\ifdim\tmpcolw>\methodcol\setlength{\methodcol}{\tmpcolw}\fi
  \addtolength{\methodcol}{2pt}%
}
\usepackage{multirow}
\usepackage{capt-of}  %
\usepackage{tikz}  %
\usepackage{fvextra}  %
\DefineVerbatimEnvironment{promptverb}{Verbatim}{breaklines,breaksymbolleft={},fontsize=\scriptsize,bgcolor=tabcreamgray!60,bgcolorpadding=3pt}
\usepackage{wrapfig}
\usepackage{tabularx}
\usepackage{array}

\newcolumntype{C}{>{\centering\arraybackslash}X}
\newcolumntype{L}{>{\raggedright\arraybackslash}X}

\newcommand{\ms}[2]{#1\hspace{0.5pt}\textcolor{gray}{\fontsize{5}{6}\selectfont\ifstrequal{#2}{--}{\phantom{\textpm0.0}}{\textpm#2}}}
\newcommand{\markcode}[2]{\if b#1\textbf{#2}\else\if u#1\underline{#2}\else#2\fi\fi}
\newcommand{\cmark}[3]{\markcode{#1}{#3}}
\newcolumntype{A}{>{\centering\arraybackslash}p{33pt}}

\definecolor{revpurple}{HTML}{8B2FA0}

\makeatletter
\renewcommand{\paragraph}{%
  \@startsection{paragraph}{4}{\z@}%
  {1.5ex \@plus 0.5ex \@minus 0.2ex}%
  {-0.6em}%
  {\normalsize\bf}%
}
\makeatother

\renewcommand{\ie}{i.e.,\xspace}
\renewcommand{\eg}{e.g.,\xspace}

\renewcommand{\vs}{vs.\@\xspace}

\newcommand{\IoUtwoD}{\ensuremath{\text{IoU2D}}\xspace}
\newcommand{\IoUthreeD}{\ensuremath{\text{IoU3D}}\xspace}
\newcommand{\NCD}{\ensuremath{\text{NCD}}\xspace}
\newcommand{\APtwoD}{\ensuremath{\text{AP}_{\text{IoU2D}}}\xspace}
\newcommand{\APthreeD}{\ensuremath{\text{AP}_{\text{IoU3D}}}\xspace}
\newcommand{\APncd}{\ensuremath{\text{AP}_{\text{NCD}}}\xspace}
\newcommand{\APtwoDat}[1]{\ensuremath{\text{AP}_{\text{IoU2D}}\text{@#1}}\xspace}
\newcommand{\APthreeDat}[1]{\ensuremath{\text{AP}_{\text{IoU3D}}\text{@#1}}\xspace}

\newcommand{\bhat}{\ensuremath{\smash[t]{\hat{b}}}}

\definecolor{pstyleblue}{rgb}{0.21,0.49,0.74}
\usepackage[pagebackref,breaklinks,colorlinks,allcolors=pstyleblue]{hyperref}
\hypersetup{urlcolor=black}  %

\def\confName{}
\def\confYear{}

\title{BOP-Refer: Benchmark for Monocular 3D and 2D Object Grounding}

\newcommand{\authorsep}{\hspace{0.75em}}
\newcommand{\affilsep}{\hspace{0.55em}}  %
\author{%
Tomas Hodan$^{1}$ \authorsep Vineet Bhat$^{2}$ \authorsep Martin Sundermeyer$^{3}$ \authorsep Jonathan Tremblay$^{4}$ \authorsep Yash Patel$^{5}$ \\
M\'ed\'eric Fourmy$^{5}$ \authorsep Stephen Tyree$^{4}$ \authorsep Junwen Huang$^{6}$ \authorsep Varun Burde$^{5}$ \authorsep Martin C\'ifka$^{5}$ \\
Eric Brachmann$^{7}$ \authorsep Vincent Lepetit$^{8}$ \authorsep Stan Birchfield$^{4}$ \authorsep Jiri Matas$^{5}$ \authorsep Carsten Rother$^{9}$ \\[0.82em]
{\normalsize $^{1}$Mistral \affilsep $^{2}$NYU \affilsep $^{3}$Google \affilsep $^{4}$NVIDIA \affilsep $^{5}$CTU Prague \affilsep $^{6}$TU Munich \affilsep $^{7}$Niantic Spatial \affilsep $^{8}$ParisTech \affilsep $^{9}$Heidelberg Uni.} \\[0.82em]
{\normalsize \textbf{Online evaluation system: \href{https://bop.felk.cvut.cz/refer}{\textcolor{black}{\underline{https://bop.felk.cvut.cz/refer}}}}}
}

\newsavebox{\authorblockbox}\newsavebox{\anonblockbox}\newlength{\anonblockpad}
\newlength{\anontoppad}\newlength{\anonbottompad}
\makeatletter
\newcommand{\measureanonpad}{\iftoggle{pstylefinal}{\setlength{\anonblockpad}{0pt}\setlength{\anontoppad}{0pt}\setlength{\anonbottompad}{0pt}}{%
  \setlength{\anontoppad}{24.26pt}\setlength{\anonbottompad}{16pt}%
  \sbox{\authorblockbox}{\large\lineskip .5em\begin{tabular}[t]{c}\@author\end{tabular}}%
  \sbox{\anonblockbox}{\large\lineskip .5em\begin{tabular}[t]{c}Anonymous \confName~submission\end{tabular}}%
  \setlength{\anonblockpad}{\dimexpr\ht\authorblockbox+\dp\authorblockbox-\ht\anonblockbox-\dp\anonblockbox-6.8pt\relax}}}
\patchcmd{\@maketitle}{\vspace*{24pt}}{\vspace*{\dimexpr 24pt+\anontoppad\relax}}{}{\PackageWarningNoLine{main}{Could not patch the title block}}
\makeatother

\begin{document}
\twocolumn[{%
\renewcommand\twocolumn[1][]{#1}%
\measureanonpad
\maketitle
\vspace*{\dimexpr\anonblockpad-\anontoppad-\anonbottompad\relax}%
\vspace{-26.0pt}  %
\begin{center}
\setlength{\tabcolsep}{1.5pt}
\newlength{\teaserw}\setlength{\teaserw}{\dimexpr(\textwidth-8\tabcolsep)/5\relax}
\newlength{\teaserpad}\setlength{\teaserpad}{3pt}
\newlength{\teaserqh}%
\sbox0{\footnotesize\itshape\parbox{5cm}{Tg\\Tg}}\setlength{\teaserqh}{\dimexpr\ht0+\dp0\relax}
\newsavebox{\teaserimgbox}\newlength{\teaserih}\newlength{\teaserch}
\newcommand{\teasercell}[2]{%
  \sbox{\teaserimgbox}{\includegraphics[width=\teaserw]{figures/#1}}%
  \setlength{\teaserih}{\ht\teaserimgbox}%
  \setlength{\teaserch}{\dimexpr\teaserih+\teaserqh+2\teaserpad\relax}%
  \begin{tikzpicture}[baseline={(0,-\teaserih)}]
    \begin{scope}
      \clip[rounded corners=2.2pt] (0,0) rectangle (\teaserw,-\teaserch);
      \node[inner sep=0,anchor=north west] at (0,0) {\usebox{\teaserimgbox}};
    \end{scope}
    \draw[black,line width=0.4pt,rounded corners=2.2pt] (0.2pt,-0.2pt) rectangle ({\teaserw-0.2pt},{-\teaserch+0.2pt});
    \node[inner sep=0,anchor=north west] at (\teaserpad,{-\teaserih-\teaserpad})
      {\footnotesize\itshape\parbox[t][\teaserqh][t]{\dimexpr\teaserw-2\teaserpad\relax}{\microtypesetup{expansion=false}\centering #2}};
  \end{tikzpicture}}
\begin{tabular}{@{}ccccc@{}}
\teasercell{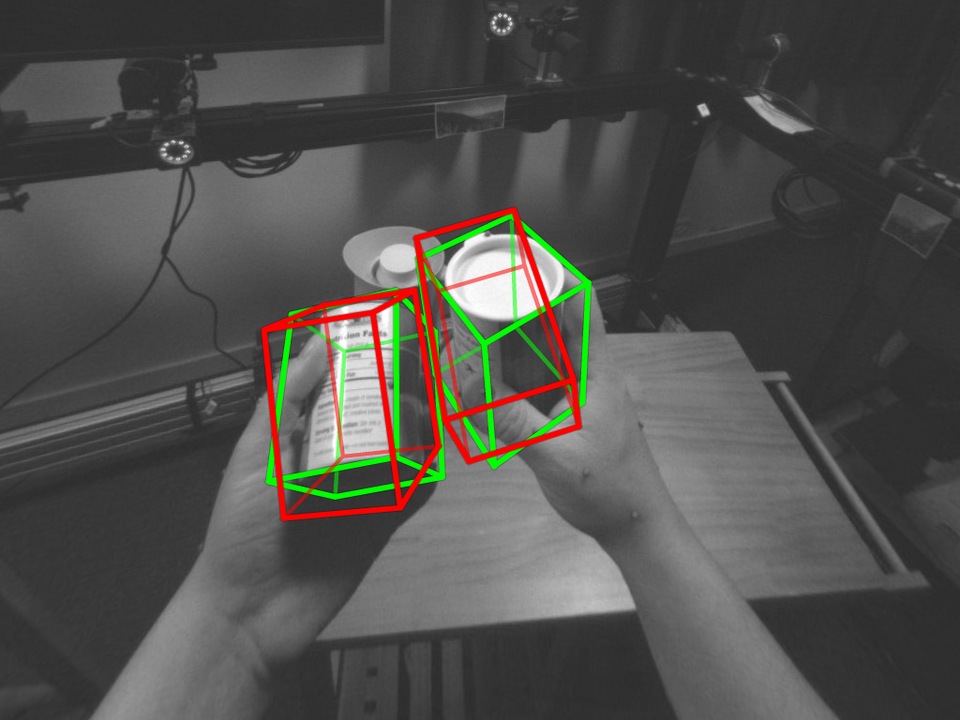}{the items held in the user's hands} &
\teasercell{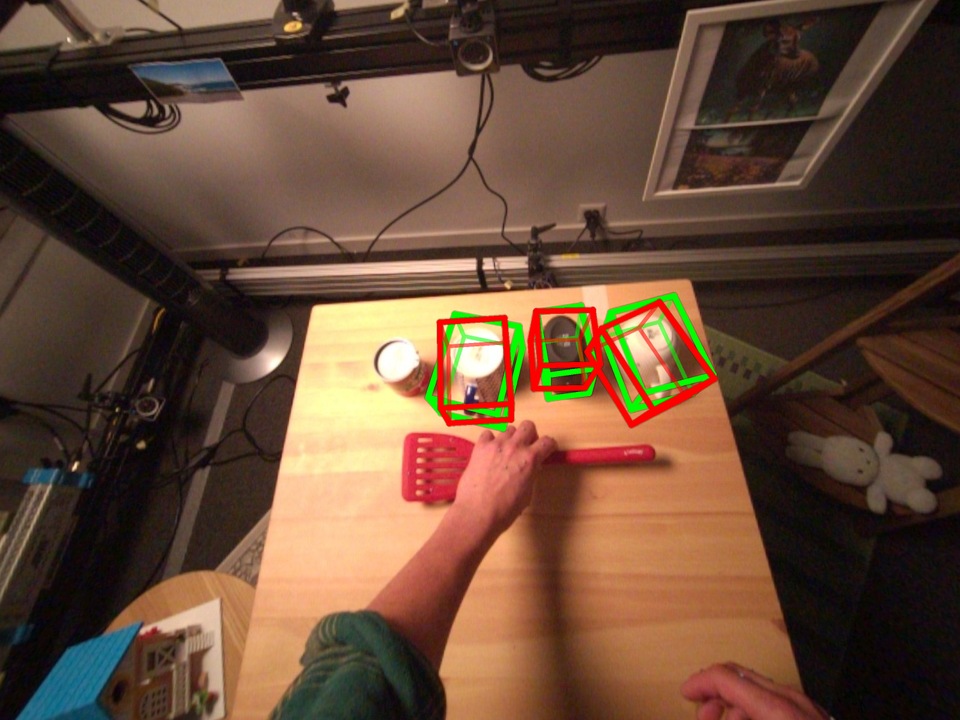}{the three ceramic vessels behind the spatula} &
\teasercell{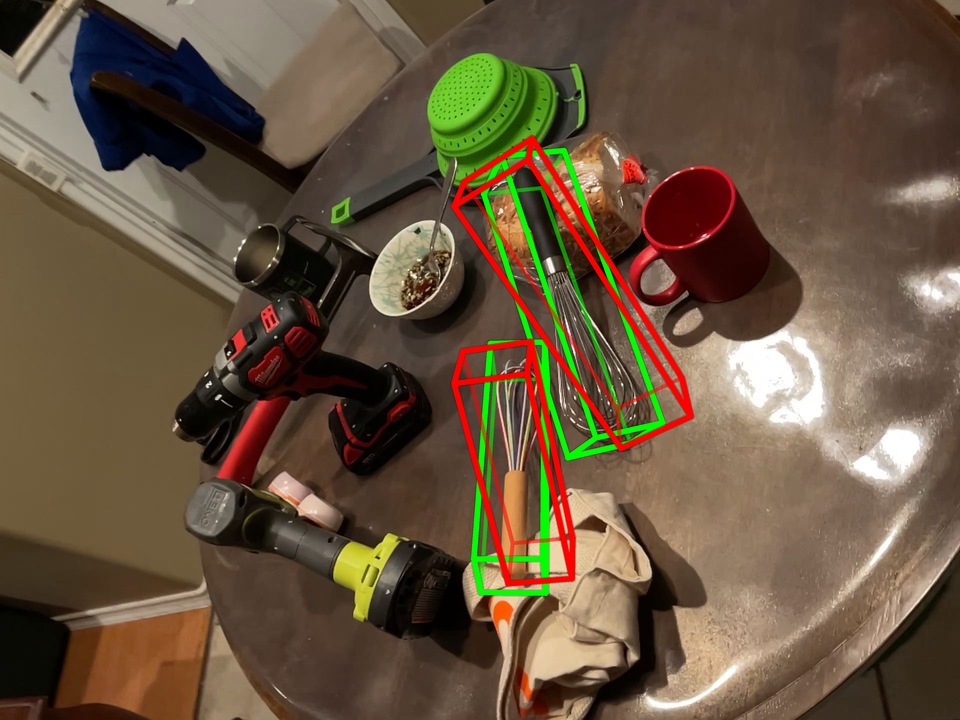}{the whisks with wire heads} &
\teasercell{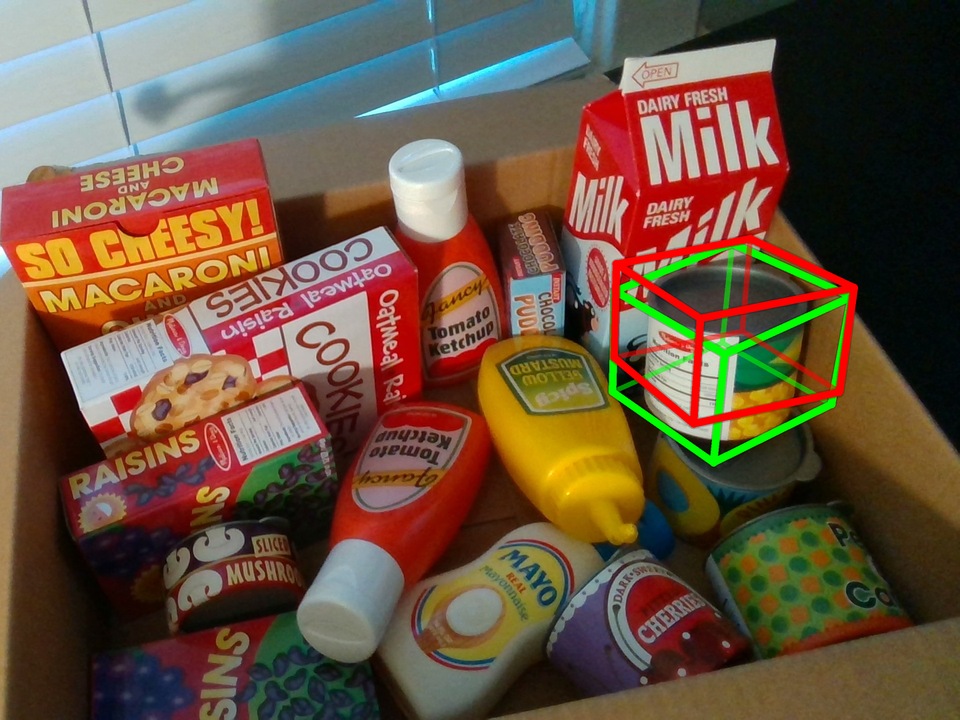}{the can on top of the pineapple can} &
\teasercell{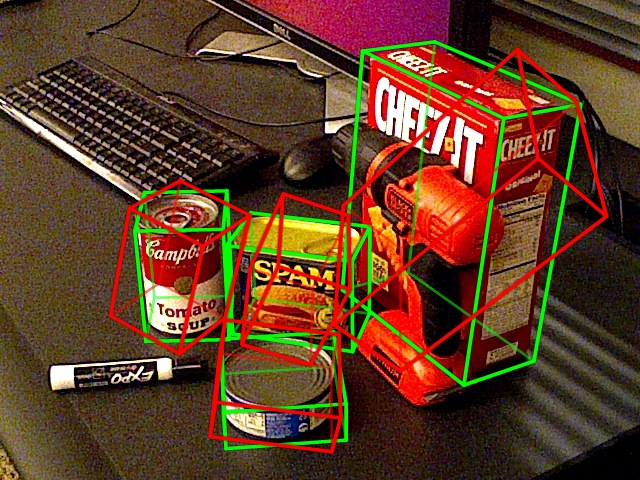}{all the pantry food items on the table} \\
\noalign{\vskip 6pt}  %
\teasercell{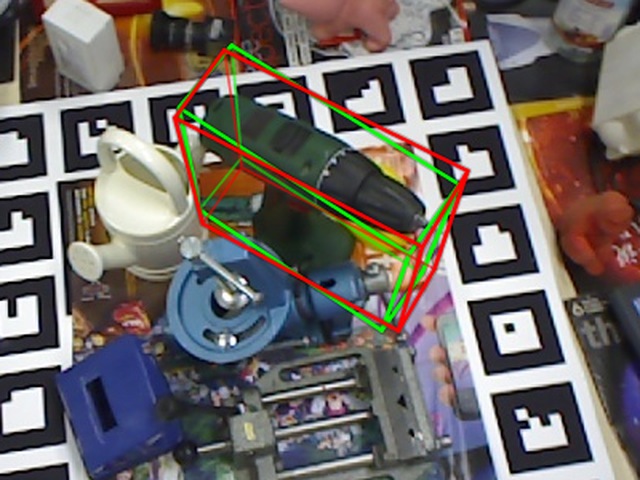}{the cordless power drill} &
\teasercell{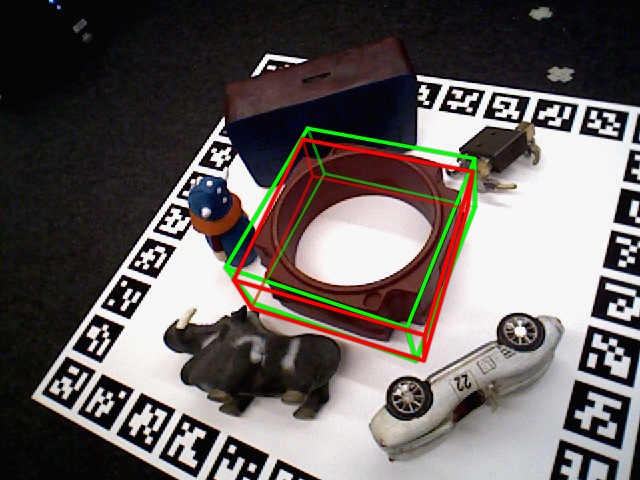}{the large object in front of the radio} &
\teasercell{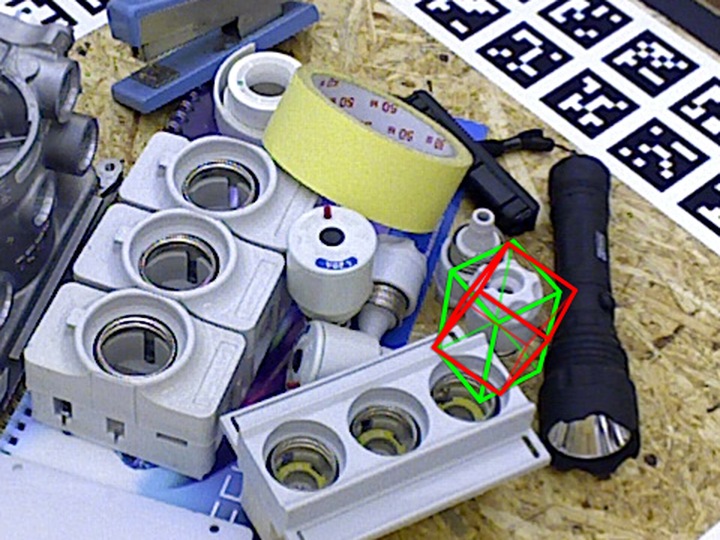}{the small ceramic insert nearest the flashlight} &
\teasercell{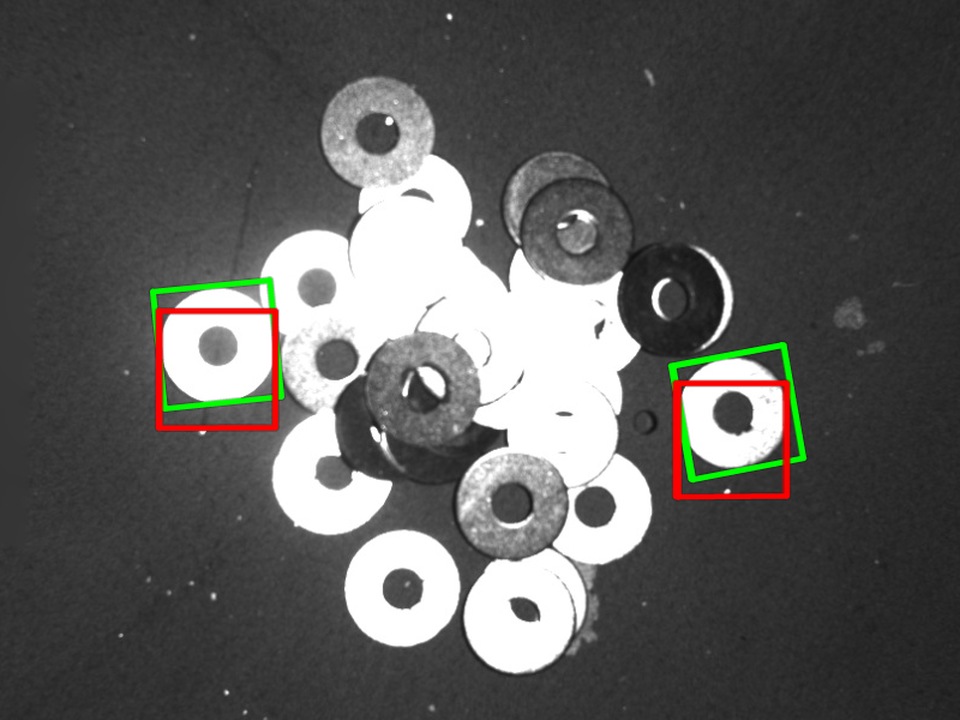}{the pair farthest apart horizontally} &
\teasercell{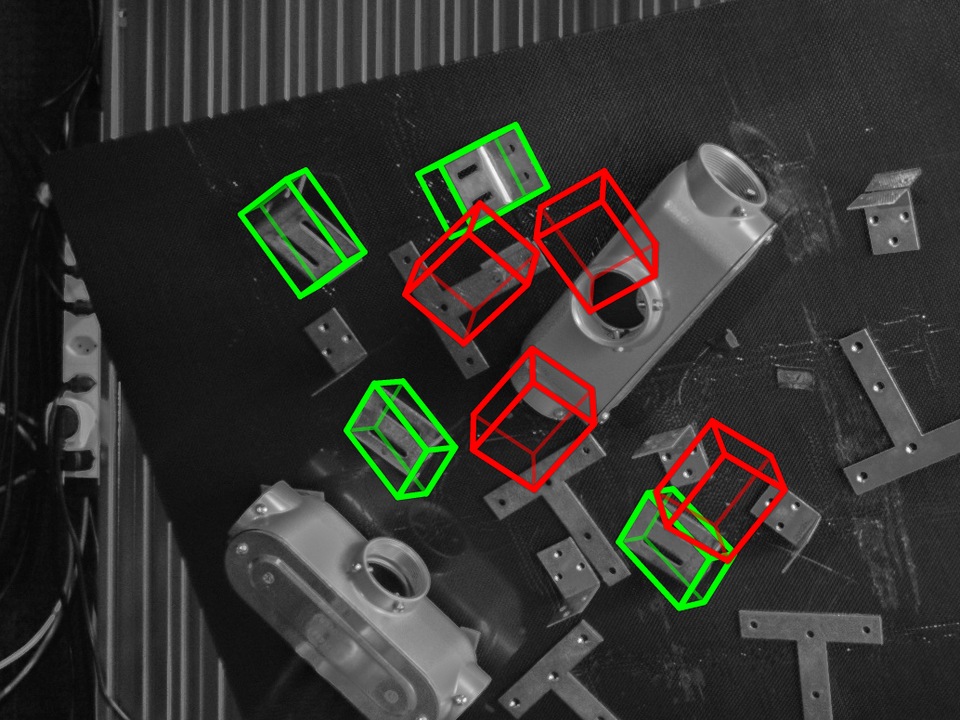}{the four slotted metal mounting brackets} \\
\end{tabular}
\vspace{-2pt}
\captionof{figure}{\textbf{Monocular 3D object grounding on BOP-Refer.} Given an image with known intrinsics and a query referring to potentially multiple objects, the task is to predict their 3D bounding boxes (2D grounding can be evaluated too). This figure shows sample images with the ground-truth boxes in green and predictions in red by Gemini~4 Carbon, currently the best in 3D and 2D of the 17 VLMs we evaluated.}
\label{fig:teaser}
\end{center}
\vspace{8pt}

\vspace*{\anonbottompad}%
}]

\begin{abstract}

We introduce \mbox{BOP-Refer}, the first benchmark for monocular 3D and 2D grounding of everyday and industrial objects sourced from 9~diverse BOP datasets.
We pair images from these datasets with VLM-generated and human-verified referring expressions, and derive accurate ground-truth 3D and 2D bounding boxes of the referred objects from their 3D models and 6DoF pose annotations.
Our systematic evaluation of 17 frontier VLMs across the major proprietary and open-weight families (Gemini, GPT, Claude, Qwen, Gemma)
shows that the best model, Gemini~4 Carbon, reaches 72.6\% \APtwoD in 2D but only 12.4\% \APthreeD in 3D,
despite the more lenient IoU thresholds in 3D. 
The predicted 3D boxes are usually centered on the target objects in the image but often off in depth by more than the object's size, suggesting that metric scale estimation, which needs reliable object size priors, is currently by far the main source of error.

\end{abstract}
\vfill\newpage  %

\section{Introduction}
\label{sec:intro}
3D object grounding, which is the task of 3D localization of objects referred to by a natural-language expression from visual or 3D input,
is important for systems that act in the physical world~\citep{chen2020scanrefer,achlioptas2020referit3d}.
Robots need 3D information for grasping, AR systems for augmenting objects with virtual content, and navigation systems for reaching named targets.
We focus on the monocular variant~\citep{zhan2024mono3dvg}, predicting 3D bounding boxes from a single image with known intrinsics rather than from the reconstructed point cloud assumed by most 3D grounding methods. This variant needs only a calibrated camera, as found on phones, robots, and AR/VR headsets.

While 2D grounding has been studied extensively, with task-specific detectors~\citep{kamath2021mdetr,liu2024grounding} 
and more recently general-purpose vision-language models (VLMs)~\citep{peng2024kosmos2,bai2023qwenvl,google2024gemini2spatial},
the 3D counterpart remains underexplored. Three factors make it harder: (1) the expected output is an oriented 3D box rather than an axis-aligned rectangle, (2) ground truth is hard to obtain at scale, and (3) a single monocular view leaves object scale and distance jointly underdetermined.
Existing 3D grounding benchmarks focus on room-scale indoor or traffic scenes~\citep{chen2020scanrefer,achlioptas2020referit3d,zhan2024mono3dvg,wang2025argusnet}, open-vocabulary 3D detectors are evaluated with category-level queries rather than free-form referring expressions~\citep{yao2024ovmono3d,huang2026wilddet3d,nvidia2025locateanything3d}, and few evaluate 2D and 3D grounding together~\citep{cho2025cubellm,daxberger2025mmspatial}, leaving it unclear whether errors arise in localizing the referred objects in the image or in estimating their 3D position, size, or orientation.

To address these limitations, we introduce BOP-Refer, a benchmark for 3D and 2D object grounding built on top of BOP (Benchmark Object Pose)~\citep{hodan2018bop,hodan2024bop}, an established benchmark suite for 6DoF object pose estimation that includes a wide range of datasets{} with accurate 3D object models and pose annotations. Our key contributions are:

\begin{enumerate}
  \item \textbf{First benchmark for monocular 3D and 2D grounding of diverse objects from referring expressions.}
   BOP-Refer covers 225 diverse everyday and industrial objects from 9 BOP datasets, with 1{,}400 validation and 1{,}400 test image-query pairs. 
   We provide VLM-generated, human-verified referring expressions (Fig.~\ref{fig:teaser}) with high-quality ground-truth 3D and 2D boxes, for direct comparison of 3D and 2D grounding on the same data.
  \item \textbf{Comprehensive evaluation of frontier VLMs.} We evaluate 12 proprietary and 5 open-weight VLMs from the Gemini, GPT, Claude, Qwen, and Gemma families, revealing that monocular 3D grounding remains far from solved, whereas the strongest models already localize most referred objects well in 2D. Our analysis points to metric scale estimation (the distance and size of the objects) as the main source of error in 3D grounding.
  \item \textbf{Symmetry-aware 3D evaluation.} We introduce two symmetry-aware 3D box comparison functions that treat symmetry-related 3D boxes as equivalent: a variant of the popular \IoUthreeD, and \NCD (Normalized Corner Distance), which can distinguish near and far misses.
  \item \textbf{Online evaluation system.} We developed an online evaluation system that scores submissions on the test split with private ground-truth annotations and provides a public leaderboard of submitted methods. The validation split has public ground truth for development.
\end{enumerate}

\section{Related work}
\label{sec:related}

\begin{table*}[!t]
\centering
\footnotesize
\renewcommand{\arraystretch}{1.1}
\newcommand{\benchtab}{%
\begin{tabular}{lcccccc}
\toprule
\textbf{Benchmark} & \textbf{Domain} & \textbf{Input} & \textbf{Output} & \textbf{Queries} & \textbf{Targets} & \textbf{GT src.} \\
\midrule
\multicolumn{7}{l}{\textit{2D object grounding}} \\
RefCOCO/+/g~\citep{kazemzadeh2014refcoco,yu2016refcoco+,mao2016refcocog} & Web (COCO)                    & RGB, query           & 2D box           & Human             & Single & Human    \\
Flickr30K Ent.~\citep{plummer2015flickr30k} & Web (Flickr30K)                 & RGB, query           & 2D box           & Human             & Multi  & Human    \\
\dashedmidrule{7}
\multicolumn{7}{l}{\textit{Spatial reasoning}} \\
RoboSpatial~\citep{liang2024robospatial} & Indoor, objects (5 sources)     & RGB, query           & VQA              & Templated         & --     & Mixed    \\
3DSRBench~\citep{ma20243dsrbench} & Web, indoor (COCO, HSSD)      & RGB, query           & VQA              & Human             & --     & Human    \\
BOP-ASK~\citep{bhat2025bopask} & Objects (3 BOP)              & RGB, $\mathbf{K}$, query      & Pose, grasp, VQA & Templ.+verif.%
& --     & Model+pose \\

\dashedmidrule{7}
\multicolumn{7}{l}{\textit{3D object grounding}} \\
ScanRefer~\citep{chen2020scanrefer} & Indoor (ScanNet)                & Point cloud, query   & 3D box           & Human             & Single & Mesh     \\
ReferIt3D~\citep{achlioptas2020referit3d} & Indoor (ScanNet)                & Point cloud, query   & 3D box           & Human, templ.\  & Single & Mesh     \\
Mono3DRefer~\citep{zhan2024mono3dvg} & Driving (KITTI)                 & RGB, $\mathbf{K}$, query      & 3D box           & LLM+verif.          & Single & LiDAR    \\
\rowcolor{tabcreamgray!86} \textbf{BOP-Refer} & Objects (9 BOP) & RGB, $\mathbf{K}$, query & 3D/2D box & VLM+verif. & Multi & Model+pose \\
\bottomrule
\end{tabular}}
\sbox0{\setlength{\tabcolsep}{0pt}\benchtab}
\setlength{\tabcolsep}{\dimexpr(\textwidth-\wd0)/14\relax}
\benchtab
\setlength{\abovecaptionskip}{7pt}
\caption{
\textbf{BOP-Refer compared with related benchmarks.} 
Among these benchmarks, only BOP-Refer unifies 3D and 2D grounding on the same data and features diverse everyday and industrial objects.
Column \emph{GT src.}\ indicates the source of ground truth -- \emph{Human}: manual annotation,
\emph{Mesh}: segmentation of reconstructed scene meshes~\citep{dai2017scannet}, \emph{LiDAR}: boxes manually fitted to point clouds, \emph{Model+pose}:
boxes derived from 3D object models (CAD or scanned) in ground-truth 6DoF poses.}
\label{tab:benchmark_comparison}
\end{table*}

\subsection{2D grounding}

Classical 2D grounding benchmarks such as RefCOCO~\citep{kazemzadeh2014refcoco}, RefCOCO+~\citep{yu2016refcoco+}, and RefCOCOg~\citep{mao2016refcocog}, derived from MSCOCO~\citep{lin2014coco}, pair each expression with a single region, whereas Flickr30k Entities~\citep{plummer2015flickr30k} and gRefCOCO~\citep{liu2023gres,ding2026grex} also include multi-target expressions, as BOP-Refer does. \citet{chen2024refl4} report labeling error rates of up to 24\% in the RefCOCO family; BOP-Refer minimizes such errors by deriving ground truth from 3D object models in annotated 6DoF poses and by verifying every query by hand. Open-vocabulary detectors such as MDETR~\citep{kamath2021mdetr}, GLIP~\citep{li2022glip}, Grounding DINO~\citep{liu2024grounding}, and OWLv2~\citep{minderer2023owlv2}, and general-purpose VLMs such as Kosmos-2~\citep{peng2024kosmos2}, the Qwen-VL series~\citep{bai2023qwenvl,qwen2025qwen3vl}, Gemini~\citep{google2024gemini2spatial}, and Molmo~\citep{deitke2024molmo} predict 2D boxes or points from referring expressions. The strongest reach $\sim$90\% accuracy on RefCOCO, where scores are saturating~\citep{chen2024refl4}, motivating our harder monocular 3D setting.

\subsection{3D grounding}

ScanRefer~\citep{chen2020scanrefer}, ReferIt3D~\citep{achlioptas2020referit3d}, and the multi-target Multi3DRefer~\citep{zhang2023multi3drefer} ground expressions in reconstructed ScanNet~\citep{dai2017scannet} point clouds. The monocular setting is addressed by Mono3DRefer~\citep{zhan2024mono3dvg}, built on KITTI~\citep{geiger2012kitti} with LLM-generated, human-verified queries, and by MM3DRefer and MT3DRefer~\citep{wang2025argusnet}, which add multiple objects and the Rope3D~\citep{ye2022rope3d} dataset (see also MonoMulti3D-ROPE~\citep{guo2025monomulti3dvg}). All match BOP-Refer's setting: predicting a 3D box from a single image with known intrinsics and a referring expression. Of these, only Mono3DRefer is publicly available, making it BOP-Refer's closest released counterpart (Tab.~\ref{tab:benchmark_comparison}). BOP-Refer differs by covering everyday and industrial objects rather than driving scenes, unifying 2D and 3D evaluation, and deriving amodal ground truth from 3D object models in 6DoF poses rather than from LiDAR.

Specialist models such as OVMono3D~\citep{yao2024ovmono3d}, DetAny3D~\citep{zhang2025detany3d}, and WildDet3D~\citep{huang2026wilddet3d} detect objects in 3D from category labels or 2D prompts, and LocateAnything3D~\citep{nvidia2025locateanything3d}, a VLM-based detector that also supports grounding, is evaluated only with templated category and location prompts (\eg ``chair on the left'') rather than free-form referring expressions, all on Omni3D~\citep{brazil2023omni3d}, which has no per-object symmetry handling. Boxer~\citep{kirillov2025boxer}, SAM~3D~\citep{meta2025sam3d}, and GR3D~\citep{gr3d} lift 2D detections, masks, or grounded regions to 3D, and Oryon and Horyon~\citep{corsetti2024oryon,corsetti2025horyon} estimate a text-described object's pose from two RGB-D views. Other 3D work, such as the Ego-Exo4D~\citep{grauman2024egoexo4d} correspondence task, ExCap3D~\citep{yeshwanth2025excap3d}, Cubify Anything~\citep{lazarow2025cubifyanything}, and SpatialLM~\citep{mao2025spatiallm}, does not ground free-form text to 3D boxes in a single image.

\subsection{Spatial reasoning}

Beyond the grounding benchmarks above, spatial reasoning in VLMs is evaluated through visual question answering with text answers, \eg by SpatialVLM~\citep{chen2024spatialvlm}, 3DSRBench~\citep{ma20243dsrbench}, 3D-LLM~\citep{hong20233dllm}, and RoboSpatial~\citep{liang2024robospatial}, or through 2D pointing at objects and regions, \eg by PointArena~\citep{cheng2025pointarena} (spatial, affordance, and counting queries). SpatialRGPT~\citep{cheng2024spatialrgpt} and Q-Spatial Bench~\citep{liao2024qspatial} ask about metric distances and sizes, and the latter shows that reference objects of known size help VLMs. BOP-ASK~\citep{bhat2025bopask}, also built on BOP datasets, provides 33M question-answer pairs for object-interaction reasoning, including referred-object poses. BOP-Refer instead targets text-to-box grounding with human-verified queries, 2D and 3D ground-truth boxes, and symmetry-aware metrics, which broad VLM benchmarks such as MMMU~\citep{yue2024mmmu}, MMBench~\citep{liu2024mmbench}, and MMStar~\citep{chen2024mmstar} do not cover.

\subsection{Vision-language models}
\label{sec:related_vlms}

Many recent VLMs output 2D bounding boxes or points: the Gemini family~\citep{google2024gemini2, google2025gemini}, Gemini Robotics ER~2~\citep{google2026geminiroboticser2,google2026geminiroboticsdocs} (a Gemini variant tuned for embodied reasoning), Claude Opus 4.6 and 4.7~\citep{anthropic2026opus, anthropic2026opus47}, Gemma 4~\citep{google2026gemma4}, and the Qwen-VL family~\citep{qwen2025qwen3vl} document 2D boxes, and GPT-5.4 and 5.5~\citep{openai2026gpt54,openai2026gpt55} return them when asked for pixel coordinates. A 3D box format is documented only for Gemini and for Qwen since Qwen3-VL, carried over to Qwen3.5~\citep{qwen2026qwen35} and Qwen3.6~\citep{qwen2026qwen36}. Their monocular 3D grounding remains largely untested: the few works that benchmark frontier closed VLMs on it cover one to three such models each, with category-level queries in room-scale scenes or referring expressions in driving scenes. Gemini Robotics~\citep{geminiroboticsteam2025geminirobotics} and the Seed1.5-VL~\citep{seed2025seed15vl} and Qwen3-VL~\citep{qwen2025qwen3vl} reports evaluate Gemini models on SUN-RGBD, MM-Spatial~\citep{daxberger2025mmspatial} evaluates GPT-4V and GPT-4o on indoor 3D grounding, and MonoVLM~\citep{qu2026monovlm} and M3SunAgent~\citep{zhang2026m3sunagent} evaluate GPT-5 and Gemini 2.5 on Mono3DRefer~\citep{zhan2024mono3dvg}, where they trail specialized methods. Qwen3-VL is moreover trained on 3D data in Omni3D's camera convention~\citep{brazil2023omni3d}, so its Omni3D results are in-distribution.

\section{BOP-Refer benchmark}
\label{sec:benchmark}

This section defines the 3D and 2D grounding tasks and their metrics (Secs.~\ref{sec:task_3d} and~\ref{sec:task_2d}), describes the source BOP datasets and the data splits (Sec.~\ref{sec:dataset}), and explains how the text queries are generated and human-verified (Sec.~\ref{sec:queries}). Supp.\ Sec.~\ref{sec:gt} details how the ground-truth 3D and 2D bounding boxes are derived from the BOP annotations.

\subsection{3D object grounding}
\label{sec:task_3d}

\paragraph{Input and output.}

Given a monocular image $I$ (RGB or monochrome) from a pinhole camera with known intrinsic matrix $\mathbf{K}$, and a text query $q$ referring to potentially multiple visible objects, the goal is to estimate oriented 3D bounding boxes $\{b_{\mathrm{3D},i}\}_{i=1}^{N}$ of the referred objects.
Each 3D box is parameterized as $b_{\mathrm{3D},i} = (\mathbf{R}, \mathbf{t}, \mathbf{s})$, where $(\mathbf{R}, \mathbf{t}) \in SE(3)$ is the rigid transformation from the box's local frame $B$ (origin at the box center, axes aligned with the box edges) to the camera frame, and $\mathbf{s} \in \mathbb{R}^3_+$ are the full extents along the three box axes. A point $\mathbf{p}_B$ expressed in $B$ thus maps to the camera frame as $\mathbf{R}\,\mathbf{p}_B + \mathbf{t}$, and the 8 box corners are $(\pm s_x/2, \pm s_y/2, \pm s_z/2)$ in $B$. The bounding boxes
are amodal (covering the entire referred objects, including any occluded parts) and have axes aligned with the object's symmetries, if any (see Supp.\ Sec.~\ref{sec:model_bbox} for a precise definition).

\paragraph{Known intrinsics.}
In practice, intrinsics are readily available in target deployment scenarios such as smartphones, robots, and AR/VR headsets. Geometrically, the focal length, object size, and distance are jointly unconstrained from a single view without intrinsics. Providing intrinsics resolves the focal length and leaves only the size-versus-distance ambiguity. The remaining problem is ill-posed but, we hypothesize, largely resolvable through learned priors about typical object sizes and scene layouts~\citep{zeng2024rsa} that VLMs acquire from large-scale visual training data. Our error analysis shows that current VLMs do not yet resolve this ambiguity (Sec.~\ref{sec:discussion}).

\begin{figure}[t]
\centering
\includegraphics[width=\columnwidth]{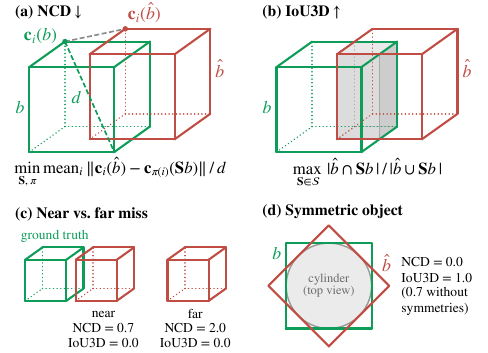}
\par\vspace{-1.5pt}
\caption{\textbf{Comparing 3D boxes.} The ground-truth box $b$ is green and the predicted box $\hat{b}$ red. (a)~\NCD is the mean distance between corresponding corners, such as $\mathbf{c}_i(b)$ and $\mathbf{c}_i(\hat{b})$ (gray line), divided by the diagonal $d$ of the ground-truth box. (b)~\IoUthreeD is the intersection volume (gray) divided by the union volume. $\mathbf{S}$ ranges over the symmetry transformations $\mathcal{S}$ of the target object and $\pi$ over the 24 corner labelings of the box~$b$ (Sec.~\ref{sec:task_3d}). Of the 225 BOP-Refer objects, 78 have an annotated discrete or continuous rotational symmetry. (c)~Once a prediction stops overlapping the ground truth, \IoUthreeD is zero, whereas \NCD still separates a near from a far miss. (d)~\NCD and \IoUthreeD are symmetry-aware: for a cylinder, a box $\hat{b}$ rotated by $45^\circ$ about its axis is equally valid and yields the optimal score $\IoUthreeD = 1.0$ and $\NCD = 0.0$.}
\label{fig:metrics}
\end{figure}

\paragraph{Measuring accuracy.}
On the 3D grounding task, we report two variants of Average Precision, \APncd and \APthreeD, which differ in the underlying box comparison function (Fig.~\ref{fig:metrics}) and thresholds of correctness.
\APncd scores predicted 3D boxes with our symmetry-aware Normalized Corner Distance (\NCD) defined below and applies thresholds $\delta \in \{0.2, 0.4, \ldots, 2.0\}$, where $\delta = 1$ corresponds to a mean corner error of one box diagonal.
\APthreeD is computed as in Omni3D~\citep{brazil2023omni3d} over thresholds $\tau \in \{0.05, 0.10, \ldots, 0.50\}$\footnote{\IoUthreeD thresholds are lower than the \IoUtwoD thresholds (Sec.~\ref{sec:task_2d}) because monocular 3D prediction is hard (see Fig.\ 11 in~\citep{brazil2023omni3d}).} but using our symmetry-aware variant of Intersection over Union (\IoUthreeD), also defined below.\footnote{A prediction is a true positive if $\NCD \leq \delta$ for \APncd and if $\IoUthreeD \geq \tau$ for \APthreeD, so a larger $\delta$ is looser and a larger $\tau$ stricter.
}
We introduce \NCD because \IoUthreeD cannot distinguish a near from a complete miss (Fig.~\ref{fig:metrics}c): once the predicted and ground-truth boxes stop overlapping, \IoUthreeD becomes zero, which is especially problematic for thin or small objects.
To prevent BOP datasets with more samples from dominating \APncd and \APthreeD, both metrics are computed per dataset and then averaged.
Details on the evaluation methodology are in Supp.\ Sec.~\ref{app:metrics}.

\paragraph{Normalized Corner Distance (\NCD).}
We define \NCD as the mean L2 distance between the 8 corresponding corners of the predicted box $\bhat_{\mathrm{3D}}$ and the ground-truth box $b_{\mathrm{3D}}$, normalized by the ground-truth box diagonal $d(b_{\mathrm{3D}})$, so that $1.0$ means the corners are off by one diagonal on average:
\begin{multline}
\text{NCD}(\hat{b}_{\mathrm{3D}}, b_{\mathrm{3D}}; \mathcal{S}) = \min_{\mathbf{S} \in \mathcal{S}} \, \min_{\pi \in \Pi} \\
\frac{1}{8\,d(b_{\mathrm{3D}})} \sum_{i=1}^{8} \big\lVert \mathbf{c}_i(\hat{b}_{\mathrm{3D}}) - \mathbf{c}_{\pi(i)}(\mathbf{S} \circ b_{\mathrm{3D}}) \big\rVert_2,
\label{eq:ncd}
\end{multline}%
where $\mathbf{c}_i(\cdot)$ is the $i$-th corner of a box.
We denote by $\mathcal{S}$ the set of symmetry transformations of the target object, given by the symmetry annotations of BOP~\citep{hodan2020bop} (available for all BOP objects, with continuous rotational symmetries discretized into a finite set) and including the identity. Applying any of them to the ground-truth box yields an equally valid box. The inner minimum resolves the ambiguity of corner labels. The same box can be described with its axes, and hence its corners, labeled in 24 different ways, and $\Pi$ is the set of the 24 corresponding corner permutations (Supp.\ Sec.~\ref{app:ncd}). Without this minimum, a prediction that describes the ground-truth box with differently ordered axes would get a nonzero \NCD. With it, \NCD depends on the two boxes only as sets of points, and $\NCD = 0$ if and only if the predicted box coincides with $\mathbf{S} \circ b_{\mathrm{3D}}$ for some $\mathbf{S} \in \mathcal{S}$. \IoUthreeD, introduced below, compares volumes, which do not depend on corner labels, so it maximizes over $\mathcal{S}$ only.

\paragraph{Intersection over Union (\IoUthreeD).}
For a predicted box $\bhat_{\mathrm{3D}}$ and a ground-truth box $b_{\mathrm{3D}}$, our symmetry-aware variant of \IoUthreeD maximizes the ratio of their intersection volume to their union volume over the symmetry transformations $\mathcal{S}$:
\begin{equation}
\text{IoU3D}(\hat{b}_{\mathrm{3D}}, b_{\mathrm{3D}}; \mathcal{S}) =
\max_{\mathbf{S} \in \mathcal{S}} \frac{\lvert \hat{b}_{\mathrm{3D}} \cap (\mathbf{S} \circ b_{\mathrm{3D}}) \rvert}{\lvert \hat{b}_{\mathrm{3D}} \cup (\mathbf{S} \circ b_{\mathrm{3D}}) \rvert},
\label{eq:iou3d}
\end{equation}
where $\lvert \cdot \rvert$ denotes volume.
We use this IoU3D variant as many BOP objects are symmetric and geometrically equivalent poses can therefore yield different box orientations.
As a concrete example, consider an upright cylinder. Any box obtained by rotating a valid box about the cylinder axis is equally valid, yet two such boxes with square cross-sections differing by $45^\circ$ have a volume ratio of only $0.71$ before the maximization, whereas $\IoUthreeD = 1$ correctly scores them as equivalent. Prior symmetry-aware variants of \IoUthreeD~\citep{wang2019nocs,ahmadyan2021objectron} consider only rotations about a single symmetry axis of rotationally symmetric object categories, whereas ours covers all discrete and continuous symmetries annotated for each object.
In BOP-Refer, 78 of the 225 objects have at least one discrete or continuous symmetry.

\subsection{2D object grounding}
\label{sec:task_2d}

\paragraph{Input and output.}
Given an image $I$ and a text query $q$ referring to one or more visible objects, the goal is to estimate their amodal 2D bounding boxes $\{b_{\mathrm{2D},i}\}_{i=1}^{N}$. Unlike in 3D, no camera intrinsics are given, as 2D boxes do not depend on them. Amodal means each box is the tightest axis-aligned rectangle around the whole object, including occluded parts. Each 2D box is $b_{\mathrm{2D},i} = (x_1, y_1, x_2, y_2)$, its top-left and bottom-right corners in pixels.

\paragraph{Measuring accuracy.}
We report the Average Precision (\APtwoD) computed with the same matching, pooling, and per-dataset macro-averaging protocol as \APthreeD (Sec.~\ref{sec:task_3d}) but using \IoUtwoD of axis-aligned boxes in place of \IoUthreeD.
Following COCO~\citep{lin2014coco}, we use \IoUtwoD thresholds $\tau \in \{0.50, 0.55, \ldots, 0.95\}$ and compute \APtwoD as the average over the thresholds (per-threshold values in Supp.\ Tab.~\ref{tab:ap_thresh_iou2d}).

\subsection{Benchmark dataset}
\label{sec:dataset}

\paragraph{Source datasets.} BOP-Refer is built on 9 BOP datasets: six everyday-object datasets (HOT3D~\citep{banerjee2024hot3d}, HANDAL~\citep{guo2023handal}, HOPEv2~\citep{tyree2022hope}, LM~\citep{hinterstoisser2012lm}, HB~\citep{kaskman2019homebreweddb}, YCB-V~\citep{xiang2018posecnn}) and three industrial-object datasets (T-LESS~\citep{hodan2017tless}, ITODD~\citep{drost2017itodd}, IPD~\citep{kalra2024ipd}), which together contain 227 unique objects, 225 of which are referred to by the BOP-Refer queries.
The datasets include accurate 3D object models and images of the objects in various scenes, annotated with 6DoF poses. We use the models and pose annotations to derive the ground-truth 3D and 2D bounding boxes (Supp.\ Sec.~\ref{sec:gt}) and to generate the referring text queries (Sec.~\ref{sec:queries}).
Fig.~\ref{fig:teaser} shows sample images and queries from each dataset, and Supp.\ Tab.~\ref{tab:datasets} lists per-dataset image counts and modality.

\paragraph{Image modality.} Images are RGB, except in ITODD, IPD, and the Quest~3 subset of HOT3D, which are monochrome. Although some datasets include depth images, we use only 2D images to keep the input modality consistent. Fisheye images from HOT3D were undistorted to a pinhole model.

\paragraph{Dataset samples.} Each sample consists of an image, camera intrinsics, a free-form natural-language query, and ground-truth 3D and 2D bounding boxes of the referred objects. A single query may refer to a single object (\eg ``the biggest mug'') or multiple objects (\eg ``all mugs on the table''). To support symmetry-aware evaluation, the ground-truth 3D bounding boxes are associated with discrete and continuous symmetry transformations of the objects.

\paragraph{Validation and test splits.} We provide validation and test splits, each containing 1{,}400 real images drawn from the 9 BOP datasets listed above, with one text query per image.
Ground-truth annotations are publicly available for the validation split but held private for the test split (accessible only through the online evaluation system).
We deliberately keep the splits compact to keep evaluation affordable even for large VLMs, and curate samples across datasets, query types, and target objects to ensure breadth. This matches popular grounding and VLM evaluation splits, which range from ${\sim}900$ to ${\sim}3{,}000$ samples (\eg 
PointArena~\citep{cheng2025pointarena}, BOP-ASK~\citep{bhat2025bopask}, MMMU~\citep{yue2024mmmu},
MMBench~\citep{liu2024mmbench}).

\paragraph{No dedicated training split.}
BOP-Refer is an evaluation-only benchmark and by design does not provide any training split, leaving the choice of training data to each evaluated approach.
Training data can be generated by applying our query-generation pipeline (Sec.~\ref{sec:queries}, code in the BOP-Refer toolkit) to any images with 3D object annotations, \eg the BOP training images of the same objects.

\subsection{Text query generation}
\label{sec:queries}

\paragraph{Generation procedure.} Text queries are generated by two VLMs (GPT-5.4~\citep{openai2026gpt54} and Gemini 3.1 Pro~\citep{google2026gemini31pro}) to
reduce model-specific bias.
For each image, we first construct a scene graph that encodes object attributes and pairwise spatial relations. Object attributes include the normalized 2D bounding box, depth, visibility fraction, apparent- and physical-size ranks, and a human-readable position description. Pairwise relations cover direction, depth ordering, proximity, occlusion, support, and relative size, each qualified, where applicable, by a small, moderate, or large margin.
We also pre-generate an appearance description of each object with the two VLMs, from 8 canonical views rendered from its 3D model.
Each of the generator VLMs is then prompted with the image, the camera intrinsics, the scene graph in YAML form, and the object appearance descriptions, and is asked to select one or more target objects and produce 5 candidate referring expressions covering a mix of strategies (appearance, spatial-relational, comparative, functional, and multi-object).
Note that the generators receive strictly more information than the models being evaluated, which receive only the image, the query, and (in the 3D task) the camera intrinsics. Details are in Supp.\ Sec.~\ref{app:query_prompts}.

\paragraph{Quality assurance.} Generated queries went through two filtering stages: (1)~automatic verification, where Claude Opus~4.7 checked each query against 14 criteria, including unambiguity, and rejected it if any failed,
and (2)~human verification, where 25 recruited reviewers inspected the remaining queries in a dedicated tool and rewrote or discarded those still judged incorrect or ambiguous. Finally, one query per image was kept, balancing object coverage and including multi-target queries.
Automatic verification keeps 49 to 71\% of the 21{,}190 candidate test queries, depending on the source dataset, and the human reviewers accept at least one query for 81\% of the inspected samples (Supp.\ Secs.~\ref{app:verification} and~\ref{app:manual_verification}).

\section{Experiments}
\label{sec:experiments}

We evaluate a diverse set of state-of-the-art VLMs from the Gemini, GPT, Claude, Qwen, and Gemma families, alongside four specialist 3D grounding pipelines (WildDet3D, Boxer, SAM~3D, and mask back-projection).
No model is fine-tuned on the benchmark.
We report test-split results here and validation-split results in Supp.\ Sec.~\ref{app:val_results}. Three key ablations are in Sec.~\ref{sec:ablations_main}, and further ones in Supp.\ Sec.~\ref{app:ablations}.

\subsection{Evaluated methods}
\label{sec:models}

\paragraph{VLMs.} We evaluate 12 proprietary and 5 open-weight models. The proprietary ones are Gemini Robotics ER~2~\citep{google2026geminiroboticser2}, four more Gemini models (3.1 Pro~\citep{google2026gemini31pro}, 3.8 Flash~\citep{google2026gemini38flash}, 4 Argon~\citep{google2026gemini4argon}, and 4 Carbon~\citep{google2026gemini4carbon}), four GPT models (5.4 and 5.5~\citep{openai2026gpt54,openai2026gpt55}, 6 Astra~\citep{openai2026gpt6astra}, and 6.1 Sol~\citep{openai2026gpt61sol}), and three Claude models (Opus~4.6~\citep{anthropic2026opus}, 4.7~\citep{anthropic2026opus47}, and~5.5~\citep{anthropic2026opus55}), all of which output 2D boxes (Sec.~\ref{sec:related_vlms}).
The open-weight ones are three successive Qwen releases with native 2D box prediction, Qwen3-VL-235B-A22B~\citep{qwen2025qwen3vl} and its natively multimodal successors Qwen3.5-397B-A17B~\citep{qwen2026qwen35} and Qwen3.6-35B-A3B~\citep{qwen2026qwen36}, as well as Qwen3.8-Flash-Next~\citep{qwen2026qwen38flashnext}, an open-weight preview of the architecture planned for Qwen4 (125B parameters, 6B active), and Gemma~4 31B~\citep{google2026gemma4}.
Each model is evaluated 3 times on both tasks 
with the same inputs and settings in every run. We report the mean and standard deviation over these runs in all result tables.

\begin{table*}[tp]
\centering
\scriptsize
\setlength{\tabcolsep}{0pt}
\renewcommand{\arraystretch}{1.15}  %
\begin{tabularx}{\textwidth}{@{}>{\raggedright\arraybackslash}p{\methodcol}@{\hspace{6pt}}C|*{9}{C}@{}}
\toprule
Method & \APncd~$\uparrow$ & HOT3D & HANDAL & HOPEv2 & YCB-V & LM & HB & T-LESS & ITODD & IPD \\
\midrule
\multicolumn{11}{@{}l}{\textit{Proprietary VLMs}} \\
Gemini 4 Carbon & \cellcolor{tabgreenfig!60}\ms{\textbf{60.3}}{0.4} & \cellcolor{tabcreamgray!70}\ms{\underline{58.2}}{2.5} & \cellcolor{tabcreamgray!86}\ms{\textbf{71.9}}{0.8} & \cellcolor{tabcreamgray!63}\ms{\textbf{52.6}}{0.9} & \cellcolor{tabcreamgray!97}\ms{\textbf{81.2}}{0.8} & \cellcolor{tabcreamgray!90}\ms{\underline{75.1}}{1.2} & \cellcolor{tabcreamgray!81}\ms{\textbf{67.7}}{2.5} & \cellcolor{tabcreamgray!75}\ms{\textbf{62.4}}{1.5} & \cellcolor{tabcreamgray!85}\ms{\textbf{71.1}}{1.6} & \cellcolor{tabcreamgray!3}\ms{\phantom{0}2.3}{1.2} \\
Claude Opus 5.5 & \cellcolor{tabgreenfig!47}\ms{\underline{46.5}}{0.2} & \cellcolor{tabcreamgray!67}\ms{56.2}{0.4} & \cellcolor{tabcreamgray!73}\ms{60.9}{1.3} & \cellcolor{tabcreamgray!43}\ms{36.3}{2.8} & \cellcolor{tabcreamgray!91}\ms{\underline{75.8}}{1.6} & \cellcolor{tabcreamgray!78}\ms{65.5}{2.1} & \cellcolor{tabcreamgray!67}\ms{55.8}{0.1} & \cellcolor{tabcreamgray!48}\ms{40.1}{1.5} & \cellcolor{tabcreamgray!25}\ms{21.1}{3.4} & \cellcolor{tabcreamgray!8}\ms{\phantom{0}7.0}{1.0} \\
Gemini 4 Argon & \cellcolor{tabgreenfig!46}\ms{45.7}{1.0} & \cellcolor{tabcreamgray!39}\ms{32.6}{1.0} & \cellcolor{tabcreamgray!80}\ms{\underline{66.6}}{0.9} & \cellcolor{tabcreamgray!59}\ms{\underline{48.9}}{2.0} & \cellcolor{tabcreamgray!81}\ms{67.3}{3.2} & \cellcolor{tabcreamgray!74}\ms{61.9}{0.4} & \cellcolor{tabcreamgray!73}\ms{\underline{61.0}}{1.4} & \cellcolor{tabcreamgray!42}\ms{35.3}{1.2} & \cellcolor{tabcreamgray!44}\ms{\underline{36.8}}{4.4} & \cellcolor{tabcreamgray!1}\ms{\phantom{0}0.7}{0.2} \\
GPT-6 Astra & \cellcolor{tabgreenfig!40}\ms{39.4}{0.7} & \cellcolor{tabcreamgray!62}\ms{51.5}{0.7} & \cellcolor{tabcreamgray!69}\ms{57.2}{0.7} & \cellcolor{tabcreamgray!12}\ms{10.4}{0.6} & \cellcolor{tabcreamgray!59}\ms{49.6}{0.9} & \cellcolor{tabcreamgray!73}\ms{61.1}{1.6} & \cellcolor{tabcreamgray!72}\ms{59.8}{2.8} & \cellcolor{tabcreamgray!53}\ms{\underline{44.4}}{2.5} & \cellcolor{tabcreamgray!19}\ms{15.8}{1.5} & \cellcolor{tabcreamgray!6}\ms{\phantom{0}4.8}{0.6} \\
GPT-6.1 Sol & \cellcolor{tabgreenfig!39}\ms{38.4}{1.2} & \cellcolor{tabcreamgray!63}\ms{52.3}{1.3} & \cellcolor{tabcreamgray!68}\ms{56.8}{1.4} & \cellcolor{tabcreamgray!13}\ms{10.7}{0.3} & \cellcolor{tabcreamgray!67}\ms{56.2}{2.6} & \cellcolor{tabcreamgray!70}\ms{58.5}{2.9} & \cellcolor{tabcreamgray!67}\ms{56.1}{2.9} & \cellcolor{tabcreamgray!47}\ms{39.4}{3.1} & \cellcolor{tabcreamgray!14}\ms{12.0}{1.7} & \cellcolor{tabcreamgray!4}\ms{\phantom{0}3.4}{0.5} \\
GPT-5.5 & \cellcolor{tabgreenfig!33}\ms{32.5}{0.4} & \cellcolor{tabcreamgray!64}\ms{53.6}{0.7} & \cellcolor{tabcreamgray!43}\ms{36.2}{1.0} & \cellcolor{tabcreamgray!8}\ms{\phantom{0}6.4}{0.7} & \cellcolor{tabcreamgray!65}\ms{54.2}{1.2} & \cellcolor{tabcreamgray!50}\ms{41.9}{0.9} & \cellcolor{tabcreamgray!52}\ms{43.6}{2.0} & \cellcolor{tabcreamgray!40}\ms{33.1}{2.6} & \cellcolor{tabcreamgray!10}\ms{\phantom{0}8.4}{0.2} & \cellcolor{tabcreamgray!18}\ms{\underline{15.1}}{1.2} \\
Gemini 3.8 Flash & \cellcolor{tabgreenfig!30}\ms{29.6}{0.7} & \cellcolor{tabcreamgray!21}\ms{17.3}{0.1} & \cellcolor{tabcreamgray!49}\ms{41.1}{0.8} & \cellcolor{tabcreamgray!4}\ms{\phantom{0}3.4}{0.4} & \cellcolor{tabcreamgray!63}\ms{52.4}{3.0} & \cellcolor{tabcreamgray!56}\ms{47.0}{2.0} & \cellcolor{tabcreamgray!62}\ms{51.9}{1.1} & \cellcolor{tabcreamgray!36}\ms{30.3}{1.3} & \cellcolor{tabcreamgray!27}\ms{22.4}{2.0} & \cellcolor{tabcreamgray!0}\ms{\phantom{0}0.3}{0.0} \\
Gemini Robotics ER 2 & \cellcolor{tabgreenfig!29}\ms{28.8}{0.4} & \cellcolor{tabcreamgray!48}\ms{39.9}{1.1} & \cellcolor{tabcreamgray!66}\ms{55.5}{1.4} & \cellcolor{tabcreamgray!33}\ms{27.2}{0.9} & \cellcolor{tabcreamgray!29}\ms{24.1}{0.5} & \cellcolor{tabcreamgray!35}\ms{29.1}{0.4} & \cellcolor{tabcreamgray!68}\ms{57.0}{0.5} & \cellcolor{tabcreamgray!8}\ms{\phantom{0}6.8}{0.2} & \cellcolor{tabcreamgray!24}\ms{19.8}{1.7} & \cellcolor{tabcreamgray!0}\ms{\phantom{0}0.0}{0.0} \\
Gemini 3.1 Pro & \cellcolor{tabgreenfig!25}\ms{24.6}{1.0} & \cellcolor{tabcreamgray!55}\ms{45.8}{3.1} & \cellcolor{tabcreamgray!57}\ms{48.0}{1.5} & \cellcolor{tabcreamgray!35}\ms{29.6}{0.8} & \cellcolor{tabcreamgray!49}\ms{40.9}{3.5} & \cellcolor{tabcreamgray!21}\ms{17.6}{1.4} & \cellcolor{tabcreamgray!21}\ms{17.3}{0.6} & \cellcolor{tabcreamgray!12}\ms{10.3}{2.0} & \cellcolor{tabcreamgray!15}\ms{12.2}{1.1} & \cellcolor{tabcreamgray!0}\ms{\phantom{0}0.1}{0.0} \\
GPT-5.4 & \cellcolor{tabgreenfig!25}\ms{24.5}{0.4} & \cellcolor{tabcreamgray!34}\ms{28.3}{0.3} & \cellcolor{tabcreamgray!54}\ms{45.2}{0.5} & \cellcolor{tabcreamgray!12}\ms{10.1}{0.1} & \cellcolor{tabcreamgray!46}\ms{38.0}{1.5} & \cellcolor{tabcreamgray!30}\ms{24.7}{0.6} & \cellcolor{tabcreamgray!45}\ms{37.5}{0.1} & \cellcolor{tabcreamgray!22}\ms{18.4}{0.5} & \cellcolor{tabcreamgray!21}\ms{17.9}{0.5} & \cellcolor{tabcreamgray!0}\ms{\phantom{0}0.0}{0.0} \\
Claude Opus 4.6 & \cellcolor{tabgreenfig!18}\ms{18.4}{0.1} & \cellcolor{tabcreamgray!15}\ms{12.4}{0.3} & \cellcolor{tabcreamgray!37}\ms{31.1}{0.3} & \cellcolor{tabcreamgray!23}\ms{18.8}{0.6} & \cellcolor{tabcreamgray!37}\ms{31.3}{0.5} & \cellcolor{tabcreamgray!39}\ms{32.3}{1.2} & \cellcolor{tabcreamgray!42}\ms{35.2}{0.3} & \cellcolor{tabcreamgray!5}\ms{\phantom{0}4.3}{0.3} & \cellcolor{tabcreamgray!0}\ms{\phantom{0}0.4}{0.1} & \cellcolor{tabcreamgray!0}\ms{\phantom{0}0.0}{0.0} \\
Claude Opus 4.7 & \cellcolor{tabgreenfig!18}\ms{18.3}{0.0} & \cellcolor{tabcreamgray!30}\ms{24.9}{0.7} & \cellcolor{tabcreamgray!44}\ms{36.7}{0.4} & \cellcolor{tabcreamgray!0}\ms{\phantom{0}0.4}{0.1} & \cellcolor{tabcreamgray!33}\ms{27.4}{1.4} & \cellcolor{tabcreamgray!21}\ms{17.2}{2.0} & \cellcolor{tabcreamgray!39}\ms{32.5}{0.7} & \cellcolor{tabcreamgray!27}\ms{22.4}{1.4} & \cellcolor{tabcreamgray!0}\ms{\phantom{0}0.0}{0.0} & \cellcolor{tabcreamgray!4}\ms{\phantom{0}3.2}{0.5} \\
\dashedmidrule{11}
\multicolumn{11}{@{}l}{\textit{Open-weight VLMs}} \\
Qwen3.8 Flash Next & \cellcolor{tabgreenfig!26}\ms{26.1}{0.5} & \cellcolor{tabcreamgray!23}\ms{19.1}{0.2} & \cellcolor{tabcreamgray!72}\ms{60.1}{1.0} & \cellcolor{tabcreamgray!38}\ms{31.4}{0.6} & \cellcolor{tabcreamgray!43}\ms{35.7}{4.3} & \cellcolor{tabcreamgray!35}\ms{29.6}{0.4} & \cellcolor{tabcreamgray!55}\ms{45.6}{2.1} & \cellcolor{tabcreamgray!9}\ms{\phantom{0}7.4}{0.2} & \cellcolor{tabcreamgray!7}\ms{\phantom{0}5.5}{0.5} & \cellcolor{tabcreamgray!0}\ms{\phantom{0}0.1}{0.1} \\
Qwen3-VL & \cellcolor{tabgreenfig!23}\ms{23.4}{0.3} & \cellcolor{tabcreamgray!12}\ms{10.0}{0.2} & \cellcolor{tabcreamgray!62}\ms{51.6}{0.4} & \cellcolor{tabcreamgray!38}\ms{32.1}{0.4} & \cellcolor{tabcreamgray!22}\ms{18.3}{1.2} & \cellcolor{tabcreamgray!61}\ms{50.8}{1.1} & \cellcolor{tabcreamgray!54}\ms{45.0}{0.3} & \cellcolor{tabcreamgray!3}\ms{\phantom{0}2.5}{0.1} & \cellcolor{tabcreamgray!0}\ms{\phantom{0}0.0}{0.0} & \cellcolor{tabcreamgray!0}\ms{\phantom{0}0.0}{0.0} \\
Qwen3.5 & \cellcolor{tabgreenfig!23}\ms{22.8}{0.1} & \cellcolor{tabcreamgray!15}\ms{12.2}{0.0} & \cellcolor{tabcreamgray!60}\ms{49.9}{0.2} & \cellcolor{tabcreamgray!16}\ms{13.4}{1.6} & \cellcolor{tabcreamgray!38}\ms{31.6}{2.2} & \cellcolor{tabcreamgray!46}\ms{38.2}{1.0} & \cellcolor{tabcreamgray!47}\ms{38.9}{1.8} & \cellcolor{tabcreamgray!17}\ms{14.1}{0.2} & \cellcolor{tabcreamgray!9}\ms{\phantom{0}7.4}{0.5} & \cellcolor{tabcreamgray!0}\ms{\phantom{0}0.0}{0.0} \\
Gemma 4 & \cellcolor{tabgreenfig!14}\ms{14.2}{0.7} & \cellcolor{tabcreamgray!13}\ms{10.5}{1.2} & \cellcolor{tabcreamgray!42}\ms{34.9}{1.0} & \cellcolor{tabcreamgray!12}\ms{\phantom{0}9.7}{1.0} & \cellcolor{tabcreamgray!20}\ms{16.5}{1.5} & \cellcolor{tabcreamgray!16}\ms{13.1}{1.2} & \cellcolor{tabcreamgray!40}\ms{33.6}{2.0} & \cellcolor{tabcreamgray!9}\ms{\phantom{0}7.6}{0.8} & \cellcolor{tabcreamgray!3}\ms{\phantom{0}2.1}{0.4} & \cellcolor{tabcreamgray!0}\ms{\phantom{0}0.0}{0.0} \\
Qwen3.6 & \cellcolor{tabgreenfig!14}\ms{13.9}{0.2} & \cellcolor{tabcreamgray!27}\ms{22.6}{0.1} & \cellcolor{tabcreamgray!49}\ms{41.1}{0.4} & \cellcolor{tabcreamgray!33}\ms{27.8}{1.8} & \cellcolor{tabcreamgray!10}\ms{\phantom{0}8.4}{0.6} & \cellcolor{tabcreamgray!17}\ms{13.9}{1.3} & \cellcolor{tabcreamgray!13}\ms{10.9}{0.9} & \cellcolor{tabcreamgray!0}\ms{\phantom{0}0.1}{0.1} & \cellcolor{tabcreamgray!0}\ms{\phantom{0}0.0}{0.0} & \cellcolor{tabcreamgray!0}\ms{\phantom{0}0.0}{0.0} \\
\dashedmidrule{11}
\multicolumn{11}{@{}l}{\textit{Other methods}} \\
Carbon2D + SAM~3 + DP2 & \cellcolor{tabgreenfig!46}\ms{45.7}{--} & \cellcolor{tabcreamgray!89}\ms{\textbf{74.7}}{--} & \cellcolor{tabcreamgray!77}\ms{64.5}{--} & \cellcolor{tabcreamgray!45}\ms{37.5}{--} & \cellcolor{tabcreamgray!69}\ms{58.0}{--} & \cellcolor{tabcreamgray!100}\ms{\textbf{83.5}}{--} & \cellcolor{tabcreamgray!61}\ms{51.0}{--} & \cellcolor{tabcreamgray!15}\ms{12.5}{--} & \cellcolor{tabcreamgray!0}\ms{\phantom{0}0.0}{--} & \cellcolor{tabcreamgray!36}\ms{\textbf{29.9}}{--} \\
WildDet3D + SAM~3 + DA3 & \cellcolor{tabgreenfig!20}\ms{19.6}{0.0} & \cellcolor{tabcreamgray!47}\ms{39.4}{0.0} & \cellcolor{tabcreamgray!55}\ms{45.8}{0.0} & \cellcolor{tabcreamgray!16}\ms{13.1}{0.0} & \cellcolor{tabcreamgray!7}\ms{\phantom{0}5.6}{0.0} & \cellcolor{tabcreamgray!60}\ms{50.4}{0.0} & \cellcolor{tabcreamgray!21}\ms{17.2}{0.0} & \cellcolor{tabcreamgray!0}\ms{\phantom{0}0.0}{0.0} & \cellcolor{tabcreamgray!0}\ms{\phantom{0}0.0}{0.0} & \cellcolor{tabcreamgray!6}\ms{\phantom{0}5.2}{0.0} \\
Boxer + SAM~3 + DA3 & \cellcolor{tabgreenfig!13}\ms{12.8}{0.0} & \cellcolor{tabcreamgray!18}\ms{15.0}{0.1} & \cellcolor{tabcreamgray!27}\ms{22.2}{0.0} & \cellcolor{tabcreamgray!21}\ms{17.7}{0.0} & \cellcolor{tabcreamgray!28}\ms{23.5}{0.2} & \cellcolor{tabcreamgray!14}\ms{12.0}{0.2} & \cellcolor{tabcreamgray!26}\ms{21.7}{0.0} & \cellcolor{tabcreamgray!1}\ms{\phantom{0}0.7}{0.0} & \cellcolor{tabcreamgray!3}\ms{\phantom{0}2.2}{0.0} & \cellcolor{tabcreamgray!1}\ms{\phantom{0}0.6}{0.0} \\
WildDet3D + SAM~3 & \cellcolor{tabgreenfig!13}\ms{12.8}{0.0} & \cellcolor{tabcreamgray!12}\ms{\phantom{0}9.7}{0.0} & \cellcolor{tabcreamgray!49}\ms{41.3}{0.0} & \cellcolor{tabcreamgray!0}\ms{\phantom{0}0.4}{0.0} & \cellcolor{tabcreamgray!30}\ms{24.9}{0.0} & \cellcolor{tabcreamgray!42}\ms{34.9}{0.0} & \cellcolor{tabcreamgray!3}\ms{\phantom{0}2.3}{0.0} & \cellcolor{tabcreamgray!0}\ms{\phantom{0}0.4}{0.0} & \cellcolor{tabcreamgray!0}\ms{\phantom{0}0.0}{0.0} & \cellcolor{tabcreamgray!1}\ms{\phantom{0}0.8}{0.0} \\
Boxer + SAM~3 & \cellcolor{tabgreenfig!12}\ms{11.5}{0.0} & \cellcolor{tabcreamgray!27}\ms{22.4}{0.0} & \cellcolor{tabcreamgray!26}\ms{21.4}{0.0} & \cellcolor{tabcreamgray!19}\ms{15.7}{0.0} & \cellcolor{tabcreamgray!17}\ms{14.6}{0.0} & \cellcolor{tabcreamgray!14}\ms{11.3}{0.0} & \cellcolor{tabcreamgray!17}\ms{14.5}{0.0} & \cellcolor{tabcreamgray!1}\ms{\phantom{0}0.9}{0.0} & \cellcolor{tabcreamgray!1}\ms{\phantom{0}0.6}{0.0} & \cellcolor{tabcreamgray!3}\ms{\phantom{0}2.1}{0.0} \\
SAM 3D + SAM~3 & \cellcolor{tabgreenfig!2}\ms{\phantom{0}1.9}{0.1} & \cellcolor{tabcreamgray!3}\ms{\phantom{0}2.8}{0.2} & \cellcolor{tabcreamgray!5}\ms{\phantom{0}4.3}{0.2} & \cellcolor{tabcreamgray!1}\ms{\phantom{0}0.6}{0.0} & \cellcolor{tabcreamgray!3}\ms{\phantom{0}2.2}{0.1} & \cellcolor{tabcreamgray!1}\ms{\phantom{0}0.5}{0.1} & \cellcolor{tabcreamgray!8}\ms{\phantom{0}6.4}{0.2} & \cellcolor{tabcreamgray!0}\ms{\phantom{0}0.0}{0.0} & \cellcolor{tabcreamgray!0}\ms{\phantom{0}0.0}{0.0} & \cellcolor{tabcreamgray!0}\ms{\phantom{0}0.0}{0.0} \\
\bottomrule
\end{tabularx}
\setlength{\abovecaptionskip}{7pt}
\caption{\textbf{3D object grounding results (\APncd in \%).} The overall score is the average of the per-dataset scores. \APthreeD is in Supp.~Tab.~\ref{tab:results_3d}.
}
\label{tab:results_ncd}
\end{table*}

\paragraph{VLM prompting strategy.} Each VLM receives the image, the text query, and a specification of the output format. In 2D, each family receives its documented prompt and box format where one exists; GPT, which documents none, receives a prompt similar to Claude's. In 3D, all models receive the same prompt, our own: it asks for the 9-DoF box format published for Gemini and Qwen (center, size, and roll/pitch/yaw angles), provides the camera intrinsics,
and includes 5 text-only in-context examples. Proprietary models run at medium reasoning effort, and Qwen3.5, Qwen3.6, and Qwen3.8 Flash Next with thinking on, except Qwen3.6 in 3D, where 62\% of its answers hit the output cap even when raised from 16K to 32K tokens.
As evaluation is costly, each model is run on the full test split in one configuration. We tried various prompts on the validation split and kept the elements with a clear effect on the scores: the camera intrinsics and the in-context examples, which improve format following for some models (Supp.\ Sec.~\ref{app:ablation_incontext_examples}). Further settings can be evaluated through the online evaluation system. Prompts, decoding parameters, and per-model details are in Supp.\ Sec.~\ref{app:prompts}.

\paragraph{Other methods.} Beyond end-to-end VLMs, we evaluate four additional 3D grounding methods, most in several variants.
First, WildDet3D~\citep{huang2026wilddet3d} is a promptable monocular 3D detector that can accept a text or 2D box/point prompt, and optionally a depth map. We evaluate it with its own SAM~3-based 2D detection~\citep{meta2025sam3} or ground-truth 2D boxes, and with no depth map, the sensor depth map, or one predicted by Depth Anything~3 (DA3)~\citep{depthanything3} or UniDepthV2~\citep{piccinelli2025unidepthv2}.
Second, Boxer~\citep{kirillov2025boxer} lifts a 2D bounding box to a 3D box, using the whole image and, optionally, 3D scene points. We evaluate it in the same settings as WildDet3D, with the gravity direction of each image estimated by GeoCalib~\citep{veicht2024geocalib} (Supp.\ Sec.~\ref{sec:supp_boxer}).
Third, SAM~3D~\citep{meta2025sam3d} reconstructs the 3D shape, pose, and texture of an object from an image and a 2D mask. It outputs no 3D box, so we take the axis-aligned box of the reconstructed mesh in its canonical frame, placed into the camera by the predicted pose and scale (Supp.\ Sec.~\ref{sec:supp_sam3d}).{} Fourth, mask back-projection lifts the SAM~3 mask~\citep{meta2025sam3} of each Gemini~4 Carbon 2D box with Depth Pro~2 (DP2) depth~\citep{hu2026depthpro2} and fits a box (Supp.\ Sec.~\ref{sec:supp_depthlift}). Like the VLMs, each variant is run 3 times{} (mask back-projection once).

\subsection{Results}
\label{sec:results}

\paragraph{3D grounding results (Tab.~\ref{tab:results_ncd} and Supp.\ Tab.~\ref{tab:results_3d}).}
Monocular 3D grounding is far from solved. The best VLM, Gemini~4 Carbon, reaches only 12.4\% \APthreeD and the next best, Claude Opus~5.5, 5.3\% (Supp.\ Tab.~\ref{tab:results_3d}), while the dedicated 3D pipelines WildDet3D and Boxer (run with SAM~3-based detection and Depth Anything~3 depth) score 1.6\% and 1.0\%.{} Mask back-projection (Carbon2D + SAM~3 + DP2), which lifts Carbon's 2D boxes with predicted depth, reaches 7.3\%, second only to Carbon's own 3D boxes. At such low scores, \APthreeD says little about how close the predictions come, as a complete miss and a box off by a few centimeters both score zero. \APncd (Tab.~\ref{tab:results_ncd}) separates them: it spreads the VLMs from 13.9\% to 60.3\%, the 12 VLMs below 2\% \APthreeD alone from 13.9\% to 32.5\%, and ranks them consistently with \APthreeD (Spearman's $\rho = 0.96$).

\APncd also shows clear progress across generations, from 18.4\% for Claude Opus~4.6 to 46.5\% for Opus~5.5, from 24.6\% for Gemini~3.1 Pro to 60.3\% for Gemini~4 Carbon, and from 24.5\% for GPT-5.4 to 39.4\% for GPT-6 Astra. Yet precise 3D boxes remain out of reach: even for Gemini~4 Carbon, AP at $\NCD \leq 1$ (a mean corner error of at most one box diagonal) is only 69.5\%. The best open-weight model reaches 26.1\% \APncd, less than half of the best proprietary score, whereas in 2D the best open-weight model reaches about 78\% of the best proprietary score (Tab.~\ref{tab:results_2d}).

Across datasets, the industrial IPD and ITODD, with small textureless parts in cluttered scenes, are the hardest for most models (no VLM exceeds 0.3\% \APthreeD on IPD), and YCB-V, LM, and HB are among the easiest. Gemini~4 Carbon gains most on ITODD (71.1\% \APncd \vs 21.1\% for Claude Opus~5.5): the camera distance barely varies there, possibly known from public ITODD data, and Carbon estimates it far more precisely (median error 2\% \vs 21\%). Over the other eight datasets, Carbon's mean \APncd is 58.9\% \vs 49.7\%. Per-threshold AP, including \APthreeDat{0.15} used in recent VLM evaluations~\citep{qwen2025qwen3vl}, is in Supp.\ Tabs.~\ref{tab:ap_thresh_ncd} and~\ref{tab:ap_thresh_iou3d}.

\medskip
\paragraph{2D grounding results (Tab.~\ref{tab:results_2d}).}
2D grounding is in much better shape. Gemini~4 Carbon leads with 72.6\% \APtwoD, ahead of Gemini~4 Argon (69.1\%) and Claude Opus~5.5 (69.0\%), 0.1 points apart and within the run-to-run standard deviation (at most $0.6$ points). The contrast with 3D is stark even at lenient thresholds: Claude Opus~5.5 reaches 91.1\% \APtwoDat{0.50} but only 17.6\% \APthreeDat{0.05} (Supp.\ Tabs.~\ref{tab:ap_thresh_iou2d} and~\ref{tab:ap_thresh_iou3d}). For Claude and GPT, the generational gains are even larger in 2D: \APtwoD grows from 1.3\% for Claude Opus~4.6 to 18.8\% for Opus~4.7 and 69.0\% for Opus~5.5, and from 4.6\% for GPT-5.4 to 25.2\% for GPT-5.5 and 65.5\% for GPT-6 Astra. This suggests that 2D grounding received little attention in earlier models but is a clear focus of recent releases. Among the open-weight models, which otherwise reach 44--57\%, Qwen3.5 stands out with 3.8\%. Its boxes are shifted by about 4\% of the image width to the right on every dataset, unlike those of Qwen3-VL and Qwen3.6 with the same prompt and parser, and we found no cause in our pipeline. Removing it raises its \APtwoD only to 8.6\%, so its low score is mostly genuine. As in 3D, IPD is the hardest dataset: Gemini 4 Carbon reaches 68.5\% \APtwoD on it, but only 11 of the 17 models exceed 25\% \APtwoD.

\begin{table*}[tp]
\centering
\scriptsize
\setlength{\tabcolsep}{0pt}
\renewcommand{\arraystretch}{1.15}  %
\begin{tabularx}{\textwidth}{@{}>{\raggedright\arraybackslash}p{\methodcol}@{\hspace{6pt}}C|*{9}{C}@{}}
\toprule
Method & \APtwoD~$\uparrow$ & HOT3D & HANDAL & HOPEv2 & YCB-V & LM & HB & T-LESS & ITODD & IPD \\
\midrule
\multicolumn{11}{@{}l}{\textit{Proprietary VLMs}} \\
Gemini 4 Carbon & \cellcolor{tabgreenfig!60}\ms{\textbf{72.6}}{0.1} & \cellcolor{tabcreamgray!86}\ms{70.6}{0.5} & \cellcolor{tabcreamgray!97}\ms{\textbf{79.7}}{0.7} & \cellcolor{tabcreamgray!79}\ms{65.1}{0.4} & \cellcolor{tabcreamgray!84}\ms{68.9}{1.9} & \cellcolor{tabcreamgray!89}\ms{\textbf{73.2}}{1.0} & \cellcolor{tabcreamgray!100}\ms{\textbf{82.1}}{0.5} & \cellcolor{tabcreamgray!89}\ms{\textbf{72.8}}{0.2} & \cellcolor{tabcreamgray!88}\ms{\textbf{72.2}}{1.5} & \cellcolor{tabcreamgray!83}\ms{\textbf{68.5}}{1.3} \\
Gemini 4 Argon & \cellcolor{tabgreenfig!58}\ms{\underline{69.1}}{0.5} & \cellcolor{tabcreamgray!80}\ms{65.8}{0.5} & \cellcolor{tabcreamgray!93}\ms{\underline{76.0}}{0.5} & \cellcolor{tabcreamgray!69}\ms{56.7}{0.8} & \cellcolor{tabcreamgray!83}\ms{68.2}{1.3} & \cellcolor{tabcreamgray!86}\ms{\underline{70.3}}{1.0} & \cellcolor{tabcreamgray!97}\ms{\underline{80.0}}{1.5} & \cellcolor{tabcreamgray!81}\ms{\underline{66.5}}{1.2} & \cellcolor{tabcreamgray!86}\ms{70.5}{1.6} & \cellcolor{tabcreamgray!83}\ms{\underline{68.2}}{2.8} \\
Claude Opus 5.5 & \cellcolor{tabgreenfig!58}\ms{69.0}{0.5} & \cellcolor{tabcreamgray!88}\ms{\textbf{72.6}}{0.4} & \cellcolor{tabcreamgray!90}\ms{73.5}{0.7} & \cellcolor{tabcreamgray!77}\ms{63.6}{1.0} & \cellcolor{tabcreamgray!87}\ms{71.8}{0.8} & \cellcolor{tabcreamgray!80}\ms{65.8}{2.0} & \cellcolor{tabcreamgray!95}\ms{78.3}{0.7} & \cellcolor{tabcreamgray!78}\ms{64.0}{0.2} & \cellcolor{tabcreamgray!86}\ms{\underline{70.9}}{1.1} & \cellcolor{tabcreamgray!74}\ms{60.9}{4.4} \\
GPT-6.1 Sol & \cellcolor{tabgreenfig!55}\ms{65.8}{0.6} & \cellcolor{tabcreamgray!88}\ms{\underline{72.5}}{0.5} & \cellcolor{tabcreamgray!92}\ms{75.7}{0.6} & \cellcolor{tabcreamgray!88}\ms{\underline{72.1}}{0.5} & \cellcolor{tabcreamgray!89}\ms{\underline{73.4}}{0.5} & \cellcolor{tabcreamgray!61}\ms{49.9}{2.1} & \cellcolor{tabcreamgray!85}\ms{69.5}{1.9} & \cellcolor{tabcreamgray!71}\ms{58.7}{2.3} & \cellcolor{tabcreamgray!72}\ms{59.5}{2.5} & \cellcolor{tabcreamgray!75}\ms{61.4}{1.8} \\
GPT-6 Astra & \cellcolor{tabgreenfig!55}\ms{65.5}{0.2} & \cellcolor{tabcreamgray!86}\ms{70.9}{0.4} & \cellcolor{tabcreamgray!91}\ms{74.6}{0.3} & \cellcolor{tabcreamgray!89}\ms{\textbf{73.0}}{0.9} & \cellcolor{tabcreamgray!92}\ms{\textbf{75.6}}{0.3} & \cellcolor{tabcreamgray!61}\ms{49.8}{1.0} & \cellcolor{tabcreamgray!85}\ms{69.8}{0.8} & \cellcolor{tabcreamgray!75}\ms{61.3}{2.0} & \cellcolor{tabcreamgray!70}\ms{57.2}{1.1} & \cellcolor{tabcreamgray!70}\ms{57.4}{0.6} \\
Gemini Robotics ER 2 & \cellcolor{tabgreenfig!53}\ms{63.1}{0.3} & \cellcolor{tabcreamgray!78}\ms{63.8}{0.6} & \cellcolor{tabcreamgray!77}\ms{63.5}{0.6} & \cellcolor{tabcreamgray!69}\ms{56.3}{0.6} & \cellcolor{tabcreamgray!82}\ms{67.7}{0.4} & \cellcolor{tabcreamgray!81}\ms{66.1}{2.4} & \cellcolor{tabcreamgray!89}\ms{73.1}{0.3} & \cellcolor{tabcreamgray!78}\ms{64.0}{1.5} & \cellcolor{tabcreamgray!81}\ms{66.7}{1.6} & \cellcolor{tabcreamgray!57}\ms{47.1}{2.2} \\
Gemini 3.8 Flash & \cellcolor{tabgreenfig!52}\ms{61.9}{0.1} & \cellcolor{tabcreamgray!76}\ms{62.0}{0.8} & \cellcolor{tabcreamgray!84}\ms{68.8}{0.3} & \cellcolor{tabcreamgray!67}\ms{55.4}{1.4} & \cellcolor{tabcreamgray!76}\ms{62.6}{0.5} & \cellcolor{tabcreamgray!69}\ms{56.7}{0.5} & \cellcolor{tabcreamgray!90}\ms{73.7}{1.4} & \cellcolor{tabcreamgray!71}\ms{58.5}{0.8} & \cellcolor{tabcreamgray!82}\ms{67.3}{0.2} & \cellcolor{tabcreamgray!64}\ms{52.2}{2.0} \\
Gemini 3.1 Pro & \cellcolor{tabgreenfig!51}\ms{61.0}{0.4} & \cellcolor{tabcreamgray!81}\ms{66.3}{0.4} & \cellcolor{tabcreamgray!75}\ms{61.9}{1.2} & \cellcolor{tabcreamgray!66}\ms{54.3}{1.7} & \cellcolor{tabcreamgray!80}\ms{65.8}{0.2} & \cellcolor{tabcreamgray!76}\ms{62.1}{1.4} & \cellcolor{tabcreamgray!89}\ms{72.9}{1.2} & \cellcolor{tabcreamgray!66}\ms{54.0}{0.5} & \cellcolor{tabcreamgray!80}\ms{65.4}{1.2} & \cellcolor{tabcreamgray!57}\ms{46.6}{0.3} \\
GPT-5.5 & \cellcolor{tabgreenfig!21}\ms{25.2}{0.5} & \cellcolor{tabcreamgray!28}\ms{22.7}{1.1} & \cellcolor{tabcreamgray!18}\ms{14.9}{0.3} & \cellcolor{tabcreamgray!29}\ms{23.6}{1.5} & \cellcolor{tabcreamgray!53}\ms{43.4}{1.9} & \cellcolor{tabcreamgray!10}\ms{\phantom{0}8.3}{0.7} & \cellcolor{tabcreamgray!39}\ms{32.4}{1.0} & \cellcolor{tabcreamgray!33}\ms{27.3}{0.7} & \cellcolor{tabcreamgray!52}\ms{42.6}{1.4} & \cellcolor{tabcreamgray!14}\ms{11.7}{0.8} \\
Claude Opus 4.7 & \cellcolor{tabgreenfig!16}\ms{18.8}{0.2} & \cellcolor{tabcreamgray!26}\ms{21.2}{0.4} & \cellcolor{tabcreamgray!7}\ms{\phantom{0}5.8}{0.3} & \cellcolor{tabcreamgray!9}\ms{\phantom{0}7.7}{0.1} & \cellcolor{tabcreamgray!53}\ms{43.6}{0.1} & \cellcolor{tabcreamgray!15}\ms{12.0}{1.1} & \cellcolor{tabcreamgray!37}\ms{30.7}{1.5} & \cellcolor{tabcreamgray!19}\ms{15.2}{0.5} & \cellcolor{tabcreamgray!38}\ms{31.6}{0.1} & \cellcolor{tabcreamgray!2}\ms{\phantom{0}1.8}{0.1} \\
GPT-5.4 & \cellcolor{tabgreenfig!4}\ms{\phantom{0}4.6}{0.1} & \cellcolor{tabcreamgray!1}\ms{\phantom{0}0.7}{0.0} & \cellcolor{tabcreamgray!2}\ms{\phantom{0}1.5}{0.2} & \cellcolor{tabcreamgray!8}\ms{\phantom{0}6.3}{0.2} & \cellcolor{tabcreamgray!7}\ms{\phantom{0}5.6}{0.9} & \cellcolor{tabcreamgray!1}\ms{\phantom{0}1.0}{0.0} & \cellcolor{tabcreamgray!4}\ms{\phantom{0}3.0}{0.2} & \cellcolor{tabcreamgray!4}\ms{\phantom{0}3.2}{0.4} & \cellcolor{tabcreamgray!24}\ms{19.8}{0.3} & \cellcolor{tabcreamgray!0}\ms{\phantom{0}0.0}{0.0} \\
Claude Opus 4.6 & \cellcolor{tabgreenfig!1}\ms{\phantom{0}1.3}{0.0} & \cellcolor{tabcreamgray!0}\ms{\phantom{0}0.4}{0.0} & \cellcolor{tabcreamgray!0}\ms{\phantom{0}0.3}{0.0} & \cellcolor{tabcreamgray!0}\ms{\phantom{0}0.2}{0.0} & \cellcolor{tabcreamgray!3}\ms{\phantom{0}2.6}{0.1} & \cellcolor{tabcreamgray!0}\ms{\phantom{0}0.3}{0.1} & \cellcolor{tabcreamgray!1}\ms{\phantom{0}0.7}{0.1} & \cellcolor{tabcreamgray!2}\ms{\phantom{0}2.0}{0.1} & \cellcolor{tabcreamgray!6}\ms{\phantom{0}5.1}{0.4} & \cellcolor{tabcreamgray!0}\ms{\phantom{0}0.0}{0.0} \\
\dashedmidrule{11}
\multicolumn{11}{@{}l}{\textit{Open-weight VLMs}} \\
Qwen3.8 Flash Next & \cellcolor{tabgreenfig!47}\ms{56.9}{0.4} & \cellcolor{tabcreamgray!82}\ms{67.3}{0.2} & \cellcolor{tabcreamgray!75}\ms{61.4}{0.4} & \cellcolor{tabcreamgray!67}\ms{55.1}{0.3} & \cellcolor{tabcreamgray!70}\ms{57.8}{0.7} & \cellcolor{tabcreamgray!51}\ms{42.2}{1.8} & \cellcolor{tabcreamgray!84}\ms{68.7}{0.8} & \cellcolor{tabcreamgray!55}\ms{45.0}{1.1} & \cellcolor{tabcreamgray!80}\ms{65.7}{1.3} & \cellcolor{tabcreamgray!60}\ms{48.9}{3.8} \\
Qwen3.6 & \cellcolor{tabgreenfig!38}\ms{45.6}{0.6} & \cellcolor{tabcreamgray!73}\ms{59.6}{1.0} & \cellcolor{tabcreamgray!54}\ms{44.4}{0.2} & \cellcolor{tabcreamgray!56}\ms{46.0}{0.1} & \cellcolor{tabcreamgray!66}\ms{54.0}{4.1} & \cellcolor{tabcreamgray!29}\ms{23.8}{0.3} & \cellcolor{tabcreamgray!65}\ms{53.0}{0.2} & \cellcolor{tabcreamgray!49}\ms{39.9}{1.7} & \cellcolor{tabcreamgray!69}\ms{56.6}{0.7} & \cellcolor{tabcreamgray!41}\ms{33.4}{2.2} \\
Qwen3-VL & \cellcolor{tabgreenfig!37}\ms{44.5}{0.4} & \cellcolor{tabcreamgray!70}\ms{57.7}{0.7} & \cellcolor{tabcreamgray!55}\ms{45.3}{1.4} & \cellcolor{tabcreamgray!55}\ms{45.3}{1.5} & \cellcolor{tabcreamgray!71}\ms{58.6}{1.3} & \cellcolor{tabcreamgray!45}\ms{37.2}{1.2} & \cellcolor{tabcreamgray!68}\ms{56.2}{0.9} & \cellcolor{tabcreamgray!43}\ms{35.0}{0.3} & \cellcolor{tabcreamgray!47}\ms{38.8}{1.2} & \cellcolor{tabcreamgray!32}\ms{26.6}{3.6} \\
Gemma 4 & \cellcolor{tabgreenfig!36}\ms{43.7}{0.5} & \cellcolor{tabcreamgray!64}\ms{52.2}{0.3} & \cellcolor{tabcreamgray!39}\ms{32.4}{0.9} & \cellcolor{tabcreamgray!40}\ms{32.6}{0.9} & \cellcolor{tabcreamgray!73}\ms{60.3}{0.1} & \cellcolor{tabcreamgray!55}\ms{45.3}{2.0} & \cellcolor{tabcreamgray!84}\ms{69.1}{1.9} & \cellcolor{tabcreamgray!50}\ms{41.4}{1.0} & \cellcolor{tabcreamgray!63}\ms{51.4}{0.4} & \cellcolor{tabcreamgray!10}\ms{\phantom{0}8.2}{0.5} \\
Qwen3.5 & \cellcolor{tabgreenfig!3}\ms{\phantom{0}3.8}{0.2} & \cellcolor{tabcreamgray!5}\ms{\phantom{0}4.2}{0.3} & \cellcolor{tabcreamgray!6}\ms{\phantom{0}4.8}{0.2} & \cellcolor{tabcreamgray!3}\ms{\phantom{0}2.4}{0.1} & \cellcolor{tabcreamgray!6}\ms{\phantom{0}5.0}{0.4} & \cellcolor{tabcreamgray!1}\ms{\phantom{0}0.6}{0.0} & \cellcolor{tabcreamgray!6}\ms{\phantom{0}5.0}{0.3} & \cellcolor{tabcreamgray!1}\ms{\phantom{0}0.9}{0.0} & \cellcolor{tabcreamgray!14}\ms{11.6}{0.6} & \cellcolor{tabcreamgray!0}\ms{\phantom{0}0.1}{0.0} \\
\bottomrule
\end{tabularx}
\setlength{\abovecaptionskip}{7pt}
\caption{\textbf{2D object grounding results (\APtwoD in \%).} The overall score is the average of the per-dataset scores.
}
\label{tab:results_2d}
\end{table*}

\subsection{Ablations}
\label{sec:ablations_main}
VLMs localize the referred objects far better in 2D than in 3D: Gemini~4 Carbon reaches 72.6\% \APtwoD but only 12.4\% \APthreeD on the same images and queries (Tab.~\ref{tab:results_2d} and Supp.\ Tab.~\ref{tab:results_3d}). To find where 3D grounding fails, we ran a range of ablations. Those on query cardinality, target visibility and size, camera intrinsics, few-shot prompting, and output representation are in Supp.\ Sec.~\ref{app:ablations}, and the three that stood out are reported here. Supp.\ Tab.~\ref{tab:ablations} lists the VLM ablations, and Supp.\ Tabs.~\ref{tab:results_ncd_oracle} and~\ref{tab:results_3d_oracle} the specialist pipelines given ground-truth 2D boxes. Each gives the method one piece of ground-truth information that it otherwise has to estimate: the distances between the camera center and the centers of the target objects, the targets' 2D boxes, or a depth map of the scene. Each ablation is run on the methods that natively accept its input. The distance and 2D-box ablations are run on three strong VLMs, one per lab (Gemini~4 Carbon, Claude Opus~5.5, and GPT-6 Astra), the 2D-box ablation also on WildDet3D, Boxer, and SAM~3D, and the depth ablation on WildDet3D and Boxer.

\paragraph{Given object distance.}
We add each target's distance from the camera center to the query. \APncd grows from 60.3\% to 87.5\% for Gemini~4 Carbon, from 46.5\% to 87.5\% for Claude Opus~5.5, and from 39.4\% to 90.6\% for GPT-6 Astra. AP at $\NCD \leq 1$ is 96\% or more, so nearly all boxes land within one box diagonal of the target, and \APthreeD increases to 58.7\%, 49.5\%, and 60.1\% (from 12.4\%, 5.3\%, and 3.0\%).

\paragraph{Given 2D box.}
We add the ground-truth 2D box of each target to the prompt, so only the 3D estimation step is left. The effect is small: \APncd changes from 60.3\%, 46.5\%, and 39.4\% to 63.4\%, 48.6\%, and 43.1\%. WildDet3D, Boxer, and SAM~3D take a 2D box as input anyway, from SAM~3 or the ground truth (Supp.\ Tab.~\ref{tab:results_ncd_oracle}). Ground-truth boxes help Boxer most, from 11.5\% to 28.1\% \APncd, partly because SAM~3 finds nothing for 656 of the 1{,}400 queries (Supp.\ Sec.~\ref{sec:supp_predictors}), whereas WildDet3D gains 1.5\,pp (6.2\,pp with DA3 depth) and SAM~3D 0.5\,pp. With ground-truth boxes but no sensor depth, their \APthreeD stays below 5\%.

\paragraph{Given depth map.}
WildDet3D and Boxer accept depth maps natively. Unlike the distance ablation, which gives a VLM one number per target, a depth map gives the distance of every pixel, from a sensor or a monocular depth estimator. With ground-truth 2D boxes and on the six datasets that provide sensor depth, predicted depth helps moderately and sensor depth much more. WildDet3D improves from 11.6\% \APncd without depth to 19.7\% with DA3 and 18.7\% with UniDepthV2, and to 33.2\% with sensor depth. Boxer improves from 22.4\% to 30.0\% and 30.8\% with predicted depth, and to 79.6\% with sensor depth, the only setting that also reaches a high \APthreeD (42.8\%, \vs at most 4.1\% otherwise).

\subsection{Discussion}
\label{sec:discussion}

\paragraph{Why does 3D grounding fail?}
The models usually find the referred objects but often misjudge their metric scale. For Gemini~4 Carbon, Claude Opus~5.5, and GPT-6 Astra, we pair each ground-truth box with the predicted box whose center is closest in the image (single-target test queries, Supp.\ Sec.~\ref{app:scale_analysis}). The predicted centers lie in almost exactly the right direction from the camera (median error at most 0.6$^\circ$), but the boxes have median errors of 7 to 18\% in distance and 7 to 15\% in size, errors coupled by the size-distance ambiguity of a single view. Such errors are costly in 3D: for the median object, an 8\,cm cube at 78\,cm, even a correctly sized and oriented box needs a distance error below 6.3\% to reach $\IoUthreeD \geq 0.25$, which holds for only 20\% (GPT-6 Astra) to 45\% (Gemini~4 Carbon) of the predictions. Over all queries, their 2D boxes miss only 1 to 2\% of the targets, but their 3D boxes miss 35 to 62\%. This is why the ground-truth distance or a depth map helps far more than the ground-truth 2D box (Sec.~\ref{sec:ablations_main}). Even given the distance, \APthreeD stays at 50 to 60\%, so size and orientation are hard to estimate too.

\paragraph{Is 3D grounding solvable?}
The low \APthreeD scores raise the question of whether precise monocular 3D grounding is tractable. Existing benchmarks suggest that it is for methods trained for the test setting: specialized models trained on the driving scenes of Mono3DRefer~\citep{zhan2024mono3dvg} and models trained on Omni3D~\citep{brazil2023omni3d}, including Qwen3-VL, reach high accuracy on these benchmarks (Supp.\ Sec.~\ref{app:omni3d_comparison}). None of the models evaluated in this paper was trained for BOP-Refer, whose objects, sizes, distances, and viewpoints differ from those benchmarks, and all of the models score low. We therefore expect the scores to improve in two ways: with training data closer to the BOP-Refer setting, and with VLMs that learn stronger general priors on object size.

\section{Conclusion}
\label{sec:conclusion}

We presented BOP-Refer, a benchmark for monocular 3D and 2D object grounding.
Built on 9 diverse BOP datasets, the benchmark pairs VLM-generated, human-verified referring expressions with accurate ground-truth boxes derived from 3D object models in annotated 6DoF poses. Our evaluation of 17 frontier VLMs shows that monocular 3D grounding remains a major open challenge, whereas the strongest models already localize most referred objects well in 2D. Most 3D failures thus come not from finding the referred objects but from estimating their metric distance and size, and giving the models the targets' distances brings by far the largest gains. As a single view constrains only the ratio of size to distance, an important direction for future work is to equip VLMs with stronger priors on the physical size of objects and the ability to infer it from context, \eg from nearby objects of known size~\citep{liao2024qspatial}.
Our query-generation pipeline applies to any images with 3D object annotations, so it can also produce matching training data, \eg from the BOP training images, and extend BOP-Refer beyond the BOP objects in future editions. The 3D object models could also enable text-to-shape tasks.
We hope BOP-Refer will help drive progress on 3D reasoning in VLMs.

\clearpage  %
\section*{Acknowledgments}
We thank Kirill Zaitsev, Kristina Cinova, Klara Janouskova, Mohak Bhardwaj, Andrew Dahdouh, Lela Eigenrauch, Felix Exner, Siqing Fan, Larissa Laich, Ekta Prashnani, and Shiji Xin, who together with the authors reviewed the generated queries. We are also grateful to NVIDIA and Google for providing compute for the evaluations.

\paragraph{LLM usage.} 
GPT-5.4 and Gemini~3.1 Pro were used to generate the object descriptions and candidate queries, Claude Opus~4.7 was used to check them against fixed criteria, and all retained queries were verified by human reviewers (Sec.~\ref{sec:queries}). No LLM was used to derive the ground-truth boxes. Claude Code (mostly with Opus~5.5) was used to edit and proofread the text, to search the literature, and to write and debug the toolkit, evaluation, and analysis code. All experiments, numbers, and claims were checked by the authors, who take full responsibility for the paper.

{
    \small
    \bibliographystyle{ieeenat_fullname}
    \bibliography{bibliography}
}

\appendix
\maketitlesupplementary
\setcounter{dbltopnumber}{3}  %

\section{Ablation studies}
\label{app:ablations}

Tab.~\ref{tab:ablations} reports 3D grounding ablations of Gemini~4 Carbon, Claude Opus~5.5, and GPT-6 Astra on the test split. Its first three groups score the default predictions on test-query subsets (Secs.~\ref{app:ablation_cardinality} and~\ref{app:ablation_vis_size}). Each subset score averages the datasets with at least one query in the subset (Sec.~\ref{app:metrics}), so the default score need not lie between the subset scores. The last two groups change the 3D prompt (Sec.~\ref{app:ablation_incontext_examples}) or add a ground-truth input (Sec.~\ref{sec:ablations_main}).

\begin{table*}[t]
\centering
\scriptsize
\setlength{\tabcolsep}{0pt}
\renewcommand{\arraystretch}{1.15}
\begin{tabularx}{\textwidth}{@{}l@{\hspace{8pt}}*{3}{C}|*{3}{C}|*{3}{C}@{}}
\toprule
& \multicolumn{3}{c|}{Gemini~4 Carbon} & \multicolumn{3}{c|}{Claude Opus~5.5} & \multicolumn{3}{c}{GPT-6 Astra} \\
Setting & \APncd~$\uparrow$ & \APthreeD~$\uparrow$ & \APtwoD~$\uparrow$ & \APncd~$\uparrow$ & \APthreeD~$\uparrow$ & \APtwoD~$\uparrow$ & \APncd~$\uparrow$ & \APthreeD~$\uparrow$ & \APtwoD~$\uparrow$ \\
\midrule
Default (Tabs.~\ref{tab:results_ncd}, \ref{tab:results_3d}, and~\ref{tab:results_2d}) & \cellcolor{tabcreamgray!82}\ms{60.3}{0.4} & \cellcolor{tabcreamgray!68}\ms{12.4}{0.5} & \cellcolor{tabcreamgray!93}\ms{72.6}{0.1} & \cellcolor{tabcreamgray!63}\ms{46.5}{0.2} & \cellcolor{tabcreamgray!29}\ms{\phantom{0}5.3}{0.3} & \cellcolor{tabcreamgray!88}\ms{69.0}{0.5} & \cellcolor{tabcreamgray!54}\ms{39.4}{0.7} & \cellcolor{tabcreamgray!16}\ms{\phantom{0}3.0}{0.1} & \cellcolor{tabcreamgray!84}\ms{65.5}{0.2} \\
\dashedmidrule{10}
\multicolumn{10}{@{}l}{\textit{Query cardinality}} \\
Single target & \cellcolor{tabcreamgray!84}\ms{61.9}{0.8} & \cellcolor{tabcreamgray!71}\ms{13.0}{0.5} & \cellcolor{tabcreamgray!92}\ms{72.0}{0.0} & \cellcolor{tabcreamgray!65}\ms{47.6}{0.7} & \cellcolor{tabcreamgray!28}\ms{\phantom{0}5.2}{0.4} & \cellcolor{tabcreamgray!89}\ms{69.5}{0.4} & \cellcolor{tabcreamgray!55}\ms{40.2}{0.7} & \cellcolor{tabcreamgray!15}\ms{\phantom{0}2.8}{0.1} & \cellcolor{tabcreamgray!84}\ms{65.3}{0.2} \\
Multiple targets & \cellcolor{tabcreamgray!75}\ms{55.2}{0.6} & \cellcolor{tabcreamgray!58}\ms{10.6}{0.8} & \cellcolor{tabcreamgray!93}\ms{72.7}{0.1} & \cellcolor{tabcreamgray!59}\ms{43.4}{1.9} & \cellcolor{tabcreamgray!32}\ms{\phantom{0}5.8}{1.3} & \cellcolor{tabcreamgray!88}\ms{69.0}{1.2} & \cellcolor{tabcreamgray!51}\ms{37.2}{1.7} & \cellcolor{tabcreamgray!17}\ms{\phantom{0}3.2}{0.5} & \cellcolor{tabcreamgray!89}\ms{69.7}{1.2} \\
\dashedmidrule{10}
\multicolumn{10}{@{}l}{\textit{Target visibility (visible fraction of the least visible target)}} \\
$<50\%$ & \cellcolor{tabcreamgray!80}\ms{58.9}{1.6} & \cellcolor{tabcreamgray!78}\ms{14.2}{2.6} & \cellcolor{tabcreamgray!73}\ms{57.3}{2.6} & \cellcolor{tabcreamgray!65}\ms{48.2}{2.0} & \cellcolor{tabcreamgray!48}\ms{\phantom{0}8.7}{2.3} & \cellcolor{tabcreamgray!64}\ms{50.2}{1.1} & \cellcolor{tabcreamgray!54}\ms{39.9}{3.0} & \cellcolor{tabcreamgray!19}\ms{\phantom{0}3.5}{2.8} & \cellcolor{tabcreamgray!73}\ms{57.2}{2.0} \\
$50$--$75\%$ & \cellcolor{tabcreamgray!80}\ms{59.0}{0.6} & \cellcolor{tabcreamgray!70}\ms{12.8}{2.0} & \cellcolor{tabcreamgray!73}\ms{56.7}{0.6} & \cellcolor{tabcreamgray!63}\ms{46.4}{2.7} & \cellcolor{tabcreamgray!25}\ms{\phantom{0}4.5}{0.8} & \cellcolor{tabcreamgray!67}\ms{52.3}{2.4} & \cellcolor{tabcreamgray!51}\ms{37.6}{2.8} & \cellcolor{tabcreamgray!21}\ms{\phantom{0}3.9}{1.7} & \cellcolor{tabcreamgray!69}\ms{54.2}{1.1} \\
$\geq 75\%$ & \cellcolor{tabcreamgray!83}\ms{60.8}{0.5} & \cellcolor{tabcreamgray!70}\ms{12.8}{0.8} & \cellcolor{tabcreamgray!99}\ms{77.7}{0.2} & \cellcolor{tabcreamgray!64}\ms{46.9}{0.5} & \cellcolor{tabcreamgray!31}\ms{\phantom{0}5.6}{0.3} & \cellcolor{tabcreamgray!93}\ms{73.1}{0.6} & \cellcolor{tabcreamgray!54}\ms{39.6}{0.5} & \cellcolor{tabcreamgray!17}\ms{\phantom{0}3.1}{0.1} & \cellcolor{tabcreamgray!87}\ms{67.9}{0.4} \\
\dashedmidrule{10}
\multicolumn{10}{@{}l}{\textit{Target size (2D box area of the largest target, relative to the image)}} \\
$<1\%$ & \cellcolor{tabcreamgray!67}\ms{49.4}{4.0} & \cellcolor{tabcreamgray!37}\ms{\phantom{0}6.8}{3.8} & \cellcolor{tabcreamgray!86}\ms{67.3}{0.4} & \cellcolor{tabcreamgray!42}\ms{31.1}{7.6} & \cellcolor{tabcreamgray!13}\ms{\phantom{0}2.3}{1.0} & \cellcolor{tabcreamgray!84}\ms{65.4}{1.9} & \cellcolor{tabcreamgray!35}\ms{25.8}{1.3} & \cellcolor{tabcreamgray!8}\ms{\phantom{0}1.5}{0.5} & \cellcolor{tabcreamgray!77}\ms{60.6}{0.8} \\
$1$--$5\%$ & \cellcolor{tabcreamgray!83}\ms{60.9}{0.4} & \cellcolor{tabcreamgray!64}\ms{11.7}{0.7} & \cellcolor{tabcreamgray!93}\ms{72.4}{0.1} & \cellcolor{tabcreamgray!61}\ms{45.1}{1.0} & \cellcolor{tabcreamgray!27}\ms{\phantom{0}4.9}{0.7} & \cellcolor{tabcreamgray!88}\ms{68.5}{0.6} & \cellcolor{tabcreamgray!53}\ms{38.8}{1.3} & \cellcolor{tabcreamgray!19}\ms{\phantom{0}3.5}{0.4} & \cellcolor{tabcreamgray!85}\ms{66.1}{0.6} \\
$>5\%$ & \cellcolor{tabcreamgray!100}\ms{73.6}{2.2} & \cellcolor{tabcreamgray!100}\ms{18.3}{3.1} & \cellcolor{tabcreamgray!100}\ms{78.2}{0.5} & \cellcolor{tabcreamgray!83}\ms{61.2}{1.2} & \cellcolor{tabcreamgray!36}\ms{\phantom{0}6.6}{0.8} & \cellcolor{tabcreamgray!95}\ms{74.6}{0.5} & \cellcolor{tabcreamgray!73}\ms{54.0}{0.6} & \cellcolor{tabcreamgray!27}\ms{\phantom{0}5.0}{0.3} & \cellcolor{tabcreamgray!93}\ms{72.4}{0.9} \\
\dashedmidrule{10}
\multicolumn{10}{@{}l}{\textit{3D prompt}} \\
Without camera intrinsics & \cellcolor{tabcreamgray!66}\ms{48.6}{0.8} & \cellcolor{tabcreamgray!38}\ms{\phantom{0}6.9}{0.3} & -- & \cellcolor{tabcreamgray!48}\ms{35.4}{0.3} & \cellcolor{tabcreamgray!13}\ms{\phantom{0}2.3}{0.2} & -- & \cellcolor{tabcreamgray!47}\ms{34.5}{0.1} & \cellcolor{tabcreamgray!13}\ms{\phantom{0}2.3}{0.1} & -- \\
0 in-context examples & \cellcolor{tabcreamgray!77}\ms{56.5}{0.4} & \cellcolor{tabcreamgray!52}\ms{\phantom{0}9.5}{0.5} & -- & \cellcolor{tabcreamgray!56}\ms{41.0}{0.4} & \cellcolor{tabcreamgray!25}\ms{\phantom{0}4.5}{0.2} & -- & \cellcolor{tabcreamgray!52}\ms{38.0}{0.9} & \cellcolor{tabcreamgray!15}\ms{\phantom{0}2.7}{0.5} & -- \\
10 in-context examples & \cellcolor{tabcreamgray!81}\ms{59.9}{0.3} & \cellcolor{tabcreamgray!72}\ms{13.1}{0.3} & -- & \cellcolor{tabcreamgray!66}\ms{48.5}{0.6} & \cellcolor{tabcreamgray!32}\ms{\phantom{0}5.9}{0.5} & -- & \cellcolor{tabcreamgray!57}\ms{42.1}{0.6} & \cellcolor{tabcreamgray!20}\ms{\phantom{0}3.7}{0.1} & -- \\
\dashedmidrule{10}
\multicolumn{10}{@{}l}{\textit{Ground-truth input (Sec.~\ref{sec:ablations_main})}} \\
+ object distance & \cellcolor{tabcreamgray!100}\ms{87.5}{0.8} & \cellcolor{tabcreamgray!100}\ms{58.7}{1.5} & -- & \cellcolor{tabcreamgray!100}\ms{87.5}{0.4} & \cellcolor{tabcreamgray!100}\ms{49.5}{0.6} & -- & \cellcolor{tabcreamgray!100}\ms{90.6}{0.1} & \cellcolor{tabcreamgray!100}\ms{60.1}{0.2} & -- \\
+ 2D box & \cellcolor{tabcreamgray!86}\ms{63.4}{0.5} & \cellcolor{tabcreamgray!78}\ms{14.2}{1.2} & -- & \cellcolor{tabcreamgray!66}\ms{48.6}{0.2} & \cellcolor{tabcreamgray!32}\ms{\phantom{0}5.8}{0.2} & -- & \cellcolor{tabcreamgray!59}\ms{43.1}{0.4} & \cellcolor{tabcreamgray!20}\ms{\phantom{0}3.6}{0.2} & -- \\
\bottomrule
\end{tabularx}
\setlength{\abovecaptionskip}{7pt}
\caption{\textbf{Ablations of three strong VLMs, one per lab (in \%).} The first row is the default setting of the main results. The next three groups score the default predictions on subsets of the test queries, and the last two change the 3D prompt or add a ground-truth input (Sec.~\ref{sec:ablations_main}). The 2D task uses no intrinsics, no in-context examples, and no ground-truth input (--). }
\label{tab:ablations}
\end{table*}

\subsection{Single-target \vs multi-target queries}
\label{app:ablation_cardinality}
Of the 1{,}400 test queries, 1{,}153 refer to a single object and 247 to several. In 2D, cardinality matters little for Gemini~4 Carbon and Claude Opus~5.5 (0.7 and 0.5\,pp \APtwoD), while GPT-6 Astra is better on multi-target queries (69.7\% \vs 65.3\%). In 3D, single-target queries are easier for all models: \APncd is 61.9\% \vs 55.2\% for Carbon, 47.6\% \vs 43.4\% for Claude Opus~5.5, and 40.2\% \vs 37.2\% for GPT-6 Astra. The \APthreeD differences are small for Claude Opus~5.5 and GPT-6 Astra (at most 0.6\,pp) and 2.4\,pp for Carbon.

\subsection{Object visibility and size}
\label{app:ablation_vis_size}
We bin each query by the visibility of its least visible target (the visible fraction of the projected object surface: heavily occluded $<\!50\%$, partially occluded $[50, 75)\%$, mostly visible $\geq\!75\%$) and by the size of its largest target (the 2D-box area as a fraction of the image area: small $<\!1\%$, medium $1$--$5\%$, large $>\!5\%$). The visibility bins hold 50, 195, and 1{,}155 test queries, and the size bins 214, 829, and 357. In 2D, all three models score highest on mostly visible targets, and most of the gain comes between partially occluded and mostly visible ones ($21.0$, $20.8$, and $13.7$\,pp \APtwoD). In 3D, visibility has no consistent effect: \APncd varies by at most $2.3$\,pp across the bins for each model, and the \APthreeD of Gemini~4 Carbon and Claude Opus~5.5 is even slightly higher on heavily occluded targets (14.2\% and 8.7\%) than on mostly visible ones (12.8\% and 5.6\%), although with a large run-to-run spread on these 50 queries. The 3D box error thus dominates the cost of occlusion.

Size has the strongest effect. From small to large targets, \APncd rises from 49.4\% to 73.6\% for Gemini~4 Carbon, from 31.1\% to 61.2\% for Claude Opus~5.5, and from 25.8\% to 54.0\% for GPT-6 Astra, while \APtwoD rises by only $10.9$, $9.2$, and $11.8$\,pp. Small targets also give the lowest \APthreeD of all bins (6.8\%, 2.3\%, and 1.5\%). Part of this effect likely comes from the metrics: \NCD and \IoUthreeD are both relative to the object size, so the same absolute position error costs more on a small object than on a large one.

\subsection{\texorpdfstring{Error analysis of the 3D predictions}{Error analysis of the 3D predictions}}
\label{app:scale_analysis}
This analysis separates finding the referred object from estimating its metric 3D box (Sec.~\ref{sec:discussion}). It uses the single-target queries of the test split and the three ablated VLMs (Sec.~\ref{sec:ablations_main}), with 3 runs per model (3{,}247 to 3{,}458 predictions each). The ground-truth box of each query is matched with the predicted box whose center is closest to the ground-truth center in the image. The two boxes are then compared in the camera frame by the angle between the directions from the camera center to their centers, the ratio of the distances of their centers from the camera center, and the ratio of their sizes (the geometric mean of the three extents). The error of a prediction is the relative error $|\rho - 1|$, where $\rho$ is the ratio of the predicted to the true value. Tab.~\ref{tab:scale_analysis} lists the results.

\paragraph{Required distance accuracy.} Consider a ground-truth cube of side $s$ and a predicted box with the correct size and orientation whose center is shifted by $D \leq 0.8\,s$. \IoUthreeD is then highest when the shift is along a box axis, where it equals $(s - D)/(s + D)$, so $\IoUthreeD \geq 0.25$ requires $D \leq 0.6\,s$ and $\IoUthreeD \geq 0.5$ requires $D \leq s/3$. Because the direction errors are small, the shift lies mostly along the viewing ray, and $D$ is close to the error of the predicted distance. The median object of the test split has a volume of 528\,cm$^3$, that of a cube with a side of 8.1\,cm, and the median distance of the objects from the camera is 77.6\,cm, so the two thresholds correspond to relative distance errors of 6.3\% and 3.5\%. Only 20 to 45\% and 11 to 29\% of the predictions meet them.

\paragraph{Per-dataset bias.} The median distance ratio varies by dataset (bottom of Tab.~\ref{tab:scale_analysis}). All models underestimate the distance on IPD (median ratios 0.81 to 0.84) and overestimate it on HOPEv2 (1.07 to 1.35) and HOT3D (1.09 to 1.20).

\begin{table}[!htbp]
\centering
\footnotesize
\setlength{\tabcolsep}{4pt}
\begin{tabular}{@{}lccc@{}}
\toprule
& Gemini & Claude & GPT-6 \\
& 4 Carbon & Opus~5.5 & Astra \\
\midrule
Median direction error ($^\circ$) & 0.6 & 0.6 & 0.5 \\
Direction error $< 5^\circ$ (\%) & 97 & 97 & 97 \\
Median distance ratio & 1.02 & 1.06 & 1.14 \\
Median distance error (\%) & 7.1 & 13.6 & 18.3 \\
Median size ratio & 1.02 & 1.02 & 1.06 \\
Median size error (\%) & 7.0 & 11.1 & 14.7 \\
Corr.\ of log distance and size ratios & 0.59 & 0.58 & 0.77 \\
Distance error $\leq 6.3\%$ (\%) & 45 & 26 & 20 \\
Distance error $\leq 3.5\%$ (\%) & 29 & 14 & 11 \\
Median distance error (cm) & 5.7 & 10.8 & 14.9 \\
Median distance error / object size & 0.65 & 1.17 & 1.57 \\
Distance error $>$ object size (\%) & 33 & 56 & 65 \\
\midrule
\multicolumn{4}{@{}l}{\textit{Targets of all queries}} \\
No 3D box overlaps the target (\%) & 35.1 & 52.3 & 62.0 \\
No 2D box overlaps the target (\%) & 1.2 & 2.1 & 1.3 \\
\midrule
\multicolumn{4}{@{}l}{\textit{Median distance ratio per dataset}} \\
HOT3D & 1.13 & 1.09 & 1.20 \\
HANDAL & 1.03 & 1.07 & 1.17 \\
HOPEv2 & 1.07 & 1.18 & 1.35 \\
YCB-V & 1.03 & 1.02 & 1.18 \\
LM & 0.96 & 1.04 & 0.94 \\
HB & 0.94 & 0.90 & 0.93 \\
T-LESS & 0.97 & 0.93 & 1.03 \\
ITODD & 0.99 & 1.02 & 1.14 \\
IPD & 0.81 & 0.84 & 0.81 \\
\bottomrule
\end{tabular}
\caption{Error analysis of the 3D predictions on the single-target test queries (3 runs per model). Ratios are predicted over ground truth, and errors are relative, $|\rho - 1|$. The correlation is computed over the predictions with a direction error below $5^\circ$. Rows 8 and 9 give the share of predictions whose distance error is small enough to reach $\IoUthreeD \geq 0.25$ and $\geq 0.5$ on the median object, and the next three rows give the distance error in absolute terms and relative to the object size. The middle part gives, over the targets of all test queries (single- and multi-target), the share of targets that no predicted 3D or 2D box of the model overlaps.}
\label{tab:scale_analysis}
\end{table}

\subsection{Prompt variants}
\label{app:ablation_incontext_examples}
We developed the 3D prompt on the validation split with several VLMs and kept the elements with a clear effect on the scores (Sec.~\ref{sec:models}). Beyond the 9-DoF JSON format of our 3D prompt, which only Qwen and Gemini document as their native 3D box format, we tried other output formats, including Spatial Cookbook--style prompts~\citep{google2024gemini2spatial} that place intermediate 2D anchors before the 3D box. None scored noticeably better, so the low scores do not appear to be an artifact of the output representation. The camera intrinsics helped some of the models, and the text-only in-context examples mainly improved format following: for one model, they raised the share of parsable answers from about 75\% to 97\% and fixed its camera-frame and unit conventions, whereas another parsed about 97\% even without them.

The 3D-prompt rows of Tab.~\ref{tab:ablations} measure both elements on the full test split, one at a time. Without the camera intrinsics, \APncd changes by $-11.7$, $-11.1$, and $-4.9$\,pp (Gemini~4 Carbon, Claude Opus~5.5, and GPT-6 Astra). Without in-context examples, it changes by $-3.8$, $-5.5$, and $-1.4$\,pp, and Carbon's share of parsable answers rises from 93.6\% to 98.9\%, as its longer 5-example prompt more often yields no answer (Sec.~\ref{app:output_parsing}), while the others stay at 99.8\% and 100.0\%. With 10 examples, it changes by $-0.4$, $+2.0$, and $+2.7$\,pp. Both elements thus also matter for the strong models, and 5 examples slightly underestimate Claude Opus~5.5 and GPT-6 Astra (by 2.0 and 2.7\,pp \APncd).

\subsection{Run-to-run variability}
\label{app:variability}
The VLMs are stochastic, so a single run cannot establish that one model outperforms another. We therefore ran the full evaluation of the test split 3 times for every model, on both tasks, and all result tables report the mean and one standard deviation over these runs. The standard deviation of the overall scores is at most $0.6$\,pp for \APtwoD, well below the gaps between the model groups of Sec.~\ref{sec:results}. For \APncd, it is at most $1.2$\,pp, well below the $13.8$\,pp lead of Gemini~4 Carbon and the gaps of $6.3$ and $5.9$\,pp that separate Claude Opus~5.5 and Gemini~4 Argon from GPT-6 Astra and GPT-6.1 Sol, and these from GPT-5.5, but not the gaps of at most $1.0$\,pp within these two pairs. For \APthreeD, it is at most $0.7$\,pp. Given the low absolute scores, small differences in \APthreeD, such as 2\% \vs 1\%, are not meaningful even where they exceed the run-to-run standard deviation.

\subsection{Query-generator bias}
\label{app:generator_bias}
Because the referring expressions are written by VLMs (GPT-5.4 and Gemini~3.1 Pro), a model might be favored by queries its own family generated. The generators produce only text: the ground-truth boxes come from the 3D object models and 6DoF pose annotations, so no information about the answer can leak through a query, and any advantage would have to come from familiarity with the phrasing. We see no sign of such an advantage for the generators themselves. In 3D, GPT-5.4 and Gemini~3.1 Pro reach only 24.5\% and 24.6\% \APncd, less than half the score of the best model (60.3\%, Gemini~4 Carbon). In 2D, GPT-5.4 reaches only 4.6\% \APtwoD, and Gemini~3.1 Pro (61.0\%) trails Claude Opus~5.5 (69.0\%), whose family generated no queries. Every query was also checked by an independent verifier (Claude Opus~4.7) and by human reviewers, who rewrote or discarded a large share of the candidates, which further dilutes the style of any single generator.

\section{Additional result tables}
\label{app:ap_thresholds}

Tab.~\ref{tab:results_3d} reports \APthreeD per dataset for the methods of Tab.~\ref{tab:results_ncd} (Sec.~\ref{sec:results}), and Tabs.~\ref{tab:results_ncd_oracle} and~\ref{tab:results_3d_oracle} report \APncd and \APthreeD for the specialist methods given ground-truth 2D boxes (Sec.~\ref{sec:ablations_main}). Tabs.~\ref{tab:ap_thresh_ncd}, \ref{tab:ap_thresh_iou3d}, and~\ref{tab:ap_thresh_iou2d} break \APncd, \APthreeD, and \APtwoD down by threshold (Sec.~\ref{app:metrics}), and the first two also list the three VLMs of Tab.~\ref{tab:ablations} given the ground-truth object distance or 2D box. Each AP is macro-averaged over the 9 datasets at its threshold, so the overall scores are the row means of these tables, and the single-threshold scores of other works are their columns, \eg \APthreeDat{0.15}~\citep{qwen2025qwen3vl} and \APtwoDat{0.50} and \APtwoDat{0.75}~\citep{lin2014coco}.

\begin{table*}[tp]  %
\begingroup
\renewenvironment{table*}[1][]{}{}%
\begin{table*}[tp]
\centering
\scriptsize
\setlength{\tabcolsep}{0pt}
\renewcommand{\arraystretch}{1.2}  %
\begin{tabularx}{\textwidth}{@{}>{\raggedright\arraybackslash}p{\methodcol}@{\hspace{6pt}}C|*{9}{C}@{}}
\toprule
Method & \APthreeD~$\uparrow$ & HOT3D & HANDAL & HOPEv2 & YCB-V & LM & HB & T-LESS & ITODD & IPD \\
\midrule
\multicolumn{11}{@{}l}{\textit{Proprietary VLMs}} \\
Gemini 4 Carbon & \cellcolor{tabgreenfig!60}\ms{\textbf{12.4}}{0.5} & \cellcolor{tabcreamgray!18}\ms{\phantom{0}\cmark{u}{b}{6.2}}{0.4} & \cellcolor{tabcreamgray!17}\ms{\cmark{u}{b}{5.9}}{0.3} & \cellcolor{tabcreamgray!12}\ms{\textbf{4.3}}{0.4} & \cellcolor{tabcreamgray!100}\ms{\textbf{34.7}}{0.8} & \cellcolor{tabcreamgray!59}\ms{\cmark{u}{b}{20.5}}{1.2} & \cellcolor{tabcreamgray!31}\ms{\textbf{10.7}}{1.9} & \cellcolor{tabcreamgray!35}\ms{\textbf{12.0}}{1.7} & \cellcolor{tabcreamgray!48}\ms{\textbf{16.8}}{1.1} & \cellcolor{tabcreamgray!0}\ms{0.0}{0.0} \\
Claude Opus 5.5 & \cellcolor{tabgreenfig!26}\ms{\phantom{0}\cmark{n}{u}{5.3}}{0.3} & \cellcolor{tabcreamgray!16}\ms{\phantom{0}\cmark{n}{u}{5.7}}{0.4} & \cellcolor{tabcreamgray!7}\ms{2.6}{0.0} & \cellcolor{tabcreamgray!4}\ms{1.4}{0.2} & \cellcolor{tabcreamgray!54}\ms{\underline{18.9}}{4.0} & \cellcolor{tabcreamgray!30}\ms{\cmark{n}{u}{10.4}}{1.1} & \cellcolor{tabcreamgray!13}\ms{\phantom{0}4.4}{1.3} & \cellcolor{tabcreamgray!10}\ms{\phantom{0}3.4}{0.2} & \cellcolor{tabcreamgray!1}\ms{\phantom{0}0.5}{0.2} & \cellcolor{tabcreamgray!1}\ms{\cmark{n}{u}{0.2}}{0.2} \\
Gemini 4 Argon & \cellcolor{tabgreenfig!20}\ms{\phantom{0}4.2}{0.7} & \cellcolor{tabcreamgray!2}\ms{\phantom{0}0.8}{0.2} & \cellcolor{tabcreamgray!8}\ms{2.9}{0.2} & \cellcolor{tabcreamgray!8}\ms{\underline{2.8}}{1.1} & \cellcolor{tabcreamgray!30}\ms{10.3}{3.6} & \cellcolor{tabcreamgray!19}\ms{\phantom{0}6.7}{1.5} & \cellcolor{tabcreamgray!19}\ms{\phantom{0}\cmark{n}{u}{6.5}}{0.8} & \cellcolor{tabcreamgray!8}\ms{\phantom{0}2.8}{0.3} & \cellcolor{tabcreamgray!14}\ms{\phantom{0}\underline{4.9}}{1.6} & \cellcolor{tabcreamgray!0}\ms{0.1}{0.1} \\
GPT-6 Astra & \cellcolor{tabgreenfig!15}\ms{\phantom{0}3.0}{0.1} & \cellcolor{tabcreamgray!10}\ms{\phantom{0}3.3}{0.4} & \cellcolor{tabcreamgray!5}\ms{1.6}{0.3} & \cellcolor{tabcreamgray!0}\ms{0.1}{0.0} & \cellcolor{tabcreamgray!7}\ms{\phantom{0}2.3}{0.7} & \cellcolor{tabcreamgray!26}\ms{\phantom{0}8.9}{1.5} & \cellcolor{tabcreamgray!14}\ms{\phantom{0}4.7}{1.4} & \cellcolor{tabcreamgray!16}\ms{\phantom{0}\underline{5.5}}{0.2} & \cellcolor{tabcreamgray!1}\ms{\phantom{0}0.5}{0.0} & \cellcolor{tabcreamgray!0}\ms{0.1}{0.1} \\
GPT-6.1 Sol & \cellcolor{tabgreenfig!14}\ms{\phantom{0}2.9}{0.4} & \cellcolor{tabcreamgray!9}\ms{\phantom{0}3.1}{0.1} & \cellcolor{tabcreamgray!5}\ms{1.6}{0.1} & \cellcolor{tabcreamgray!0}\ms{0.1}{0.0} & \cellcolor{tabcreamgray!12}\ms{\phantom{0}4.2}{1.2} & \cellcolor{tabcreamgray!23}\ms{\phantom{0}8.0}{0.8} & \cellcolor{tabcreamgray!15}\ms{\phantom{0}5.2}{1.8} & \cellcolor{tabcreamgray!10}\ms{\phantom{0}3.5}{0.4} & \cellcolor{tabcreamgray!1}\ms{\phantom{0}0.4}{0.2} & \cellcolor{tabcreamgray!0}\ms{0.0}{0.0} \\
GPT-5.5 & \cellcolor{tabgreenfig!8}\ms{\phantom{0}1.7}{0.1} & \cellcolor{tabcreamgray!11}\ms{\phantom{0}3.7}{0.2} & \cellcolor{tabcreamgray!1}\ms{0.4}{0.1} & \cellcolor{tabcreamgray!0}\ms{0.1}{0.0} & \cellcolor{tabcreamgray!13}\ms{\phantom{0}4.4}{0.5} & \cellcolor{tabcreamgray!6}\ms{\phantom{0}2.0}{0.5} & \cellcolor{tabcreamgray!6}\ms{\phantom{0}2.2}{0.6} & \cellcolor{tabcreamgray!5}\ms{\phantom{0}1.9}{0.1} & \cellcolor{tabcreamgray!1}\ms{\phantom{0}0.2}{0.1} & \cellcolor{tabcreamgray!1}\ms{\cmark{u}{b}{0.3}}{0.3} \\
Gemini 3.8 Flash & \cellcolor{tabgreenfig!6}\ms{\phantom{0}1.3}{0.0} & \cellcolor{tabcreamgray!0}\ms{\phantom{0}0.1}{0.0} & \cellcolor{tabcreamgray!1}\ms{0.5}{0.1} & \cellcolor{tabcreamgray!0}\ms{0.0}{0.0} & \cellcolor{tabcreamgray!5}\ms{\phantom{0}1.6}{0.1} & \cellcolor{tabcreamgray!8}\ms{\phantom{0}2.8}{0.2} & \cellcolor{tabcreamgray!14}\ms{\phantom{0}4.9}{0.6} & \cellcolor{tabcreamgray!2}\ms{\phantom{0}0.7}{0.1} & \cellcolor{tabcreamgray!2}\ms{\phantom{0}0.7}{0.3} & \cellcolor{tabcreamgray!0}\ms{0.0}{0.0} \\
Gemini 3.1 Pro & \cellcolor{tabgreenfig!5}\ms{\phantom{0}1.1}{0.3} & \cellcolor{tabcreamgray!7}\ms{\phantom{0}2.4}{0.3} & \cellcolor{tabcreamgray!3}\ms{1.1}{0.2} & \cellcolor{tabcreamgray!3}\ms{0.9}{0.2} & \cellcolor{tabcreamgray!14}\ms{\phantom{0}4.9}{2.4} & \cellcolor{tabcreamgray!1}\ms{\phantom{0}0.3}{0.3} & \cellcolor{tabcreamgray!0}\ms{\phantom{0}0.0}{0.0} & \cellcolor{tabcreamgray!1}\ms{\phantom{0}0.2}{0.1} & \cellcolor{tabcreamgray!1}\ms{\phantom{0}0.4}{0.1} & \cellcolor{tabcreamgray!0}\ms{0.0}{0.0} \\
Gemini Robotics ER 2 & \cellcolor{tabgreenfig!4}\ms{\phantom{0}0.9}{0.1} & \cellcolor{tabcreamgray!2}\ms{\phantom{0}0.7}{0.1} & \cellcolor{tabcreamgray!5}\ms{1.7}{0.2} & \cellcolor{tabcreamgray!1}\ms{0.4}{0.2} & \cellcolor{tabcreamgray!0}\ms{\phantom{0}0.0}{0.0} & \cellcolor{tabcreamgray!2}\ms{\phantom{0}0.7}{0.1} & \cellcolor{tabcreamgray!11}\ms{\phantom{0}3.8}{0.5} & \cellcolor{tabcreamgray!0}\ms{\phantom{0}0.0}{0.0} & \cellcolor{tabcreamgray!1}\ms{\phantom{0}0.5}{0.0} & \cellcolor{tabcreamgray!0}\ms{0.0}{0.0} \\
GPT-5.4 & \cellcolor{tabgreenfig!1}\ms{\phantom{0}0.3}{0.0} & \cellcolor{tabcreamgray!1}\ms{\phantom{0}0.3}{0.0} & \cellcolor{tabcreamgray!1}\ms{0.2}{0.0} & \cellcolor{tabcreamgray!0}\ms{0.0}{0.0} & \cellcolor{tabcreamgray!4}\ms{\phantom{0}1.4}{0.3} & \cellcolor{tabcreamgray!0}\ms{\phantom{0}0.1}{0.0} & \cellcolor{tabcreamgray!1}\ms{\phantom{0}0.4}{0.1} & \cellcolor{tabcreamgray!1}\ms{\phantom{0}0.2}{0.1} & \cellcolor{tabcreamgray!0}\ms{\phantom{0}0.0}{0.0} & \cellcolor{tabcreamgray!0}\ms{0.0}{0.0} \\
Claude Opus 4.7 & \cellcolor{tabgreenfig!0}\ms{\phantom{0}0.1}{0.0} & \cellcolor{tabcreamgray!1}\ms{\phantom{0}0.2}{0.1} & \cellcolor{tabcreamgray!0}\ms{0.1}{0.0} & \cellcolor{tabcreamgray!0}\ms{0.0}{0.0} & \cellcolor{tabcreamgray!1}\ms{\phantom{0}0.2}{0.0} & \cellcolor{tabcreamgray!0}\ms{\phantom{0}0.0}{0.0} & \cellcolor{tabcreamgray!1}\ms{\phantom{0}0.2}{0.0} & \cellcolor{tabcreamgray!1}\ms{\phantom{0}0.3}{0.1} & \cellcolor{tabcreamgray!0}\ms{\phantom{0}0.0}{0.0} & \cellcolor{tabcreamgray!0}\ms{0.0}{0.0} \\
Claude Opus 4.6 & \cellcolor{tabgreenfig!0}\ms{\phantom{0}0.1}{0.0} & \cellcolor{tabcreamgray!0}\ms{\phantom{0}0.0}{0.0} & \cellcolor{tabcreamgray!0}\ms{0.1}{0.0} & \cellcolor{tabcreamgray!0}\ms{0.1}{0.0} & \cellcolor{tabcreamgray!1}\ms{\phantom{0}0.2}{0.1} & \cellcolor{tabcreamgray!0}\ms{\phantom{0}0.1}{0.0} & \cellcolor{tabcreamgray!0}\ms{\phantom{0}0.1}{0.0} & \cellcolor{tabcreamgray!0}\ms{\phantom{0}0.0}{0.0} & \cellcolor{tabcreamgray!0}\ms{\phantom{0}0.0}{0.0} & \cellcolor{tabcreamgray!0}\ms{0.0}{0.0} \\
\dashedmidrule{11}
\multicolumn{11}{@{}l}{\textit{Open-weight VLMs}} \\
Qwen3-VL & \cellcolor{tabgreenfig!6}\ms{\phantom{0}1.2}{0.1} & \cellcolor{tabcreamgray!0}\ms{\phantom{0}0.0}{0.0} & \cellcolor{tabcreamgray!3}\ms{1.1}{0.1} & \cellcolor{tabcreamgray!5}\ms{1.6}{0.1} & \cellcolor{tabcreamgray!0}\ms{\phantom{0}0.0}{0.0} & \cellcolor{tabcreamgray!11}\ms{\phantom{0}3.9}{0.6} & \cellcolor{tabcreamgray!12}\ms{\phantom{0}4.2}{0.1} & \cellcolor{tabcreamgray!0}\ms{\phantom{0}0.0}{0.0} & \cellcolor{tabcreamgray!0}\ms{\phantom{0}0.0}{0.0} & \cellcolor{tabcreamgray!0}\ms{0.0}{0.0} \\
Qwen3.8 Flash Next & \cellcolor{tabgreenfig!3}\ms{\phantom{0}0.7}{0.1} & \cellcolor{tabcreamgray!1}\ms{\phantom{0}0.3}{0.0} & \cellcolor{tabcreamgray!4}\ms{1.3}{0.2} & \cellcolor{tabcreamgray!4}\ms{1.3}{0.3} & \cellcolor{tabcreamgray!3}\ms{\phantom{0}1.1}{0.3} & \cellcolor{tabcreamgray!2}\ms{\phantom{0}0.6}{0.1} & \cellcolor{tabcreamgray!4}\ms{\phantom{0}1.4}{0.3} & \cellcolor{tabcreamgray!1}\ms{\phantom{0}0.2}{0.0} & \cellcolor{tabcreamgray!0}\ms{\phantom{0}0.0}{0.0} & \cellcolor{tabcreamgray!0}\ms{0.0}{0.0} \\
Qwen3.5 & \cellcolor{tabgreenfig!1}\ms{\phantom{0}0.3}{0.0} & \cellcolor{tabcreamgray!0}\ms{\phantom{0}0.0}{0.0} & \cellcolor{tabcreamgray!1}\ms{0.5}{0.1} & \cellcolor{tabcreamgray!0}\ms{0.1}{0.0} & \cellcolor{tabcreamgray!3}\ms{\phantom{0}1.1}{0.3} & \cellcolor{tabcreamgray!2}\ms{\phantom{0}0.8}{0.3} & \cellcolor{tabcreamgray!1}\ms{\phantom{0}0.5}{0.1} & \cellcolor{tabcreamgray!0}\ms{\phantom{0}0.1}{0.1} & \cellcolor{tabcreamgray!0}\ms{\phantom{0}0.0}{0.0} & \cellcolor{tabcreamgray!0}\ms{0.0}{0.0} \\
Qwen3.6 & \cellcolor{tabgreenfig!0}\ms{\phantom{0}0.1}{0.0} & \cellcolor{tabcreamgray!0}\ms{\phantom{0}0.1}{0.0} & \cellcolor{tabcreamgray!0}\ms{0.1}{0.0} & \cellcolor{tabcreamgray!2}\ms{0.7}{0.1} & \cellcolor{tabcreamgray!0}\ms{\phantom{0}0.0}{0.0} & \cellcolor{tabcreamgray!0}\ms{\phantom{0}0.0}{0.0} & \cellcolor{tabcreamgray!0}\ms{\phantom{0}0.0}{0.0} & \cellcolor{tabcreamgray!0}\ms{\phantom{0}0.0}{0.0} & \cellcolor{tabcreamgray!0}\ms{\phantom{0}0.0}{0.0} & \cellcolor{tabcreamgray!0}\ms{0.0}{0.0} \\
Gemma 4 & \cellcolor{tabgreenfig!0}\ms{\phantom{0}0.1}{0.0} & \cellcolor{tabcreamgray!0}\ms{\phantom{0}0.0}{0.0} & \cellcolor{tabcreamgray!0}\ms{0.1}{0.0} & \cellcolor{tabcreamgray!0}\ms{0.0}{0.0} & \cellcolor{tabcreamgray!0}\ms{\phantom{0}0.0}{0.0} & \cellcolor{tabcreamgray!0}\ms{\phantom{0}0.0}{0.0} & \cellcolor{tabcreamgray!1}\ms{\phantom{0}0.3}{0.0} & \cellcolor{tabcreamgray!0}\ms{\phantom{0}0.0}{0.0} & \cellcolor{tabcreamgray!0}\ms{\phantom{0}0.0}{0.0} & \cellcolor{tabcreamgray!0}\ms{0.0}{0.0} \\
\dashedmidrule{11}
\multicolumn{11}{@{}l}{\textit{Other methods}} \\
Carbon2D + SAM~3 + DP2 & \cellcolor{tabgreenfig!36}\ms{\phantom{0}\underline{7.3}}{--} & \cellcolor{tabcreamgray!43}\ms{\textbf{15.0}}{--} & \cellcolor{tabcreamgray!27}\ms{\textbf{9.2}}{--} & \cellcolor{tabcreamgray!4}\ms{1.3}{--} & \cellcolor{tabcreamgray!10}\ms{\phantom{0}3.5}{--} & \cellcolor{tabcreamgray!76}\ms{\textbf{26.3}}{--} & \cellcolor{tabcreamgray!23}\ms{\phantom{0}\underline{8.1}}{--} & \cellcolor{tabcreamgray!3}\ms{\phantom{0}0.9}{--} & \cellcolor{tabcreamgray!0}\ms{\phantom{0}0.0}{--} & \cellcolor{tabcreamgray!4}\ms{\textbf{1.5}}{--} \\

WildDet3D + SAM~3 + DA3 & \cellcolor{tabgreenfig!8}\ms{\phantom{0}1.6}{0.0} & \cellcolor{tabcreamgray!7}\ms{\phantom{0}2.5}{0.0} & \cellcolor{tabcreamgray!12}\ms{\cmark{n}{u}{4.1}}{0.0} & \cellcolor{tabcreamgray!1}\ms{0.2}{0.0} & \cellcolor{tabcreamgray!0}\ms{\phantom{0}0.0}{0.0} & \cellcolor{tabcreamgray!18}\ms{\phantom{0}6.4}{0.0} & \cellcolor{tabcreamgray!2}\ms{\phantom{0}0.7}{0.0} & \cellcolor{tabcreamgray!0}\ms{\phantom{0}0.0}{0.0} & \cellcolor{tabcreamgray!0}\ms{\phantom{0}0.0}{0.0} & \cellcolor{tabcreamgray!1}\ms{\cmark{u}{b}{0.3}}{0.0} \\
Boxer + SAM~3 & \cellcolor{tabgreenfig!5}\ms{\phantom{0}1.1}{0.0} & \cellcolor{tabcreamgray!11}\ms{\phantom{0}3.9}{0.0} & \cellcolor{tabcreamgray!6}\ms{2.1}{0.0} & \cellcolor{tabcreamgray!3}\ms{1.2}{0.0} & \cellcolor{tabcreamgray!1}\ms{\phantom{0}0.2}{0.0} & \cellcolor{tabcreamgray!4}\ms{\phantom{0}1.3}{0.0} & \cellcolor{tabcreamgray!3}\ms{\phantom{0}1.1}{0.0} & \cellcolor{tabcreamgray!0}\ms{\phantom{0}0.0}{0.0} & \cellcolor{tabcreamgray!0}\ms{\phantom{0}0.1}{0.0} & \cellcolor{tabcreamgray!0}\ms{0.0}{0.0} \\
Boxer + SAM~3 + DA3 & \cellcolor{tabgreenfig!5}\ms{\phantom{0}1.0}{0.0} & \cellcolor{tabcreamgray!3}\ms{\phantom{0}1.0}{0.0} & \cellcolor{tabcreamgray!4}\ms{1.5}{0.0} & \cellcolor{tabcreamgray!4}\ms{1.5}{0.0} & \cellcolor{tabcreamgray!4}\ms{\phantom{0}1.5}{0.1} & \cellcolor{tabcreamgray!3}\ms{\phantom{0}1.2}{0.0} & \cellcolor{tabcreamgray!5}\ms{\phantom{0}1.9}{0.0} & \cellcolor{tabcreamgray!0}\ms{\phantom{0}0.0}{0.0} & \cellcolor{tabcreamgray!0}\ms{\phantom{0}0.1}{0.0} & \cellcolor{tabcreamgray!0}\ms{0.0}{0.0} \\
WildDet3D + SAM~3 & \cellcolor{tabgreenfig!2}\ms{\phantom{0}0.4}{0.0} & \cellcolor{tabcreamgray!0}\ms{\phantom{0}0.0}{0.0} & \cellcolor{tabcreamgray!7}\ms{2.6}{0.0} & \cellcolor{tabcreamgray!0}\ms{0.0}{0.0} & \cellcolor{tabcreamgray!1}\ms{\phantom{0}0.4}{0.0} & \cellcolor{tabcreamgray!2}\ms{\phantom{0}0.8}{0.0} & \cellcolor{tabcreamgray!0}\ms{\phantom{0}0.0}{0.0} & \cellcolor{tabcreamgray!0}\ms{\phantom{0}0.0}{0.0} & \cellcolor{tabcreamgray!0}\ms{\phantom{0}0.0}{0.0} & \cellcolor{tabcreamgray!0}\ms{0.1}{0.0} \\
SAM 3D + SAM~3 & \cellcolor{tabgreenfig!0}\ms{\phantom{0}0.0}{0.0} & \cellcolor{tabcreamgray!0}\ms{\phantom{0}0.0}{0.0} & \cellcolor{tabcreamgray!0}\ms{0.0}{0.0} & \cellcolor{tabcreamgray!0}\ms{0.0}{0.0} & \cellcolor{tabcreamgray!0}\ms{\phantom{0}0.0}{0.0} & \cellcolor{tabcreamgray!0}\ms{\phantom{0}0.0}{0.0} & \cellcolor{tabcreamgray!0}\ms{\phantom{0}0.1}{0.0} & \cellcolor{tabcreamgray!0}\ms{\phantom{0}0.0}{0.0} & \cellcolor{tabcreamgray!0}\ms{\phantom{0}0.0}{0.0} & \cellcolor{tabcreamgray!0}\ms{0.0}{0.0} \\
\bottomrule
\end{tabularx}
\setlength{\abovecaptionskip}{7pt}
\caption{\textbf{3D object grounding results (\APthreeD in \%).} The overall score is the average of the per-dataset scores.
}
\label{tab:results_3d}
\end{table*}

\bigskip
\begin{table*}[tp]
\centering
\scriptsize
\setlength{\tabcolsep}{0pt}
\renewcommand{\arraystretch}{1.15}  %
\begin{tabularx}{\textwidth}{@{}>{\raggedright\arraybackslash}p{\methodcol}@{\hspace{6pt}}C|*{9}{C}@{}}
\toprule
Method & \APncd~$\uparrow$ & HOT3D & HANDAL & HOPEv2 & YCB-V & LM & HB & T-LESS & ITODD & IPD \\
\midrule
Boxer + DA3 & \cellcolor{tabgreenfig!60}\ms{\textbf{31.4}}{0.1} & \cellcolor{tabcreamgray!40}\ms{36.0}{0.3} & \cellcolor{tabcreamgray!61}\ms{\textbf{54.5}}{0.1} & \cellcolor{tabcreamgray!44}\ms{39.1}{0.1} & \cellcolor{tabcreamgray!51}\ms{44.9}{0.4} & \cellcolor{tabcreamgray!68}\ms{60.3}{0.1} & \cellcolor{tabcreamgray!37}\ms{32.6}{0.2} & \cellcolor{tabcreamgray!1}\ms{\phantom{0}0.8}{0.0} & \cellcolor{tabcreamgray!2}\ms{\phantom{0}\underline{2.2}}{0.0} & \cellcolor{tabcreamgray!13}\ms{11.9}{0.0} \\
Boxer + UniDepthV2 & \cellcolor{tabgreenfig!58}\ms{\underline{30.2}}{0.2} & \cellcolor{tabcreamgray!50}\ms{44.3}{0.8} & \cellcolor{tabcreamgray!47}\ms{41.5}{0.2} & \cellcolor{tabcreamgray!31}\ms{27.9}{0.1} & \cellcolor{tabcreamgray!65}\ms{\underline{57.8}}{0.5} & \cellcolor{tabcreamgray!75}\ms{67.1}{0.4} & \cellcolor{tabcreamgray!29}\ms{26.1}{0.1} & \cellcolor{tabcreamgray!6}\ms{\phantom{0}5.7}{0.0} & \cellcolor{tabcreamgray!0}\ms{\phantom{0}0.1}{0.0} & \cellcolor{tabcreamgray!1}\ms{\phantom{0}0.8}{0.0} \\
Boxer & \cellcolor{tabgreenfig!54}\ms{28.1}{0.0} & \cellcolor{tabcreamgray!57}\ms{\underline{50.7}}{0.0} & \cellcolor{tabcreamgray!60}\ms{\underline{52.9}}{0.0} & \cellcolor{tabcreamgray!42}\ms{37.2}{0.0} & \cellcolor{tabcreamgray!23}\ms{20.4}{0.0} & \cellcolor{tabcreamgray!59}\ms{52.8}{0.0} & \cellcolor{tabcreamgray!27}\ms{23.6}{0.0} & \cellcolor{tabcreamgray!0}\ms{\phantom{0}0.3}{0.0} & \cellcolor{tabcreamgray!0}\ms{\phantom{0}0.2}{0.0} & \cellcolor{tabcreamgray!17}\ms{\underline{14.8}}{0.0} \\
WildDet3D + DA3 & \cellcolor{tabgreenfig!50}\ms{25.8}{0.0} & \cellcolor{tabcreamgray!55}\ms{49.2}{0.0} & \cellcolor{tabcreamgray!50}\ms{44.3}{0.0} & \cellcolor{tabcreamgray!28}\ms{24.8}{0.0} & \cellcolor{tabcreamgray!6}\ms{\phantom{0}5.1}{0.0} & \cellcolor{tabcreamgray!79}\ms{69.8}{0.0} & \cellcolor{tabcreamgray!21}\ms{18.3}{0.0} & \cellcolor{tabcreamgray!0}\ms{\phantom{0}0.1}{0.0} & \cellcolor{tabcreamgray!0}\ms{\phantom{0}0.0}{0.0} & \cellcolor{tabcreamgray!24}\ms{\textbf{21.2}}{0.0} \\
WildDet3D + UniDepthV2 & \cellcolor{tabgreenfig!45}\ms{23.6}{0.0} & \cellcolor{tabcreamgray!71}\ms{\textbf{62.9}}{0.0} & \cellcolor{tabcreamgray!38}\ms{34.2}{0.0} & \cellcolor{tabcreamgray!17}\ms{15.3}{0.0} & \cellcolor{tabcreamgray!21}\ms{18.4}{0.0} & \cellcolor{tabcreamgray!72}\ms{64.3}{0.0} & \cellcolor{tabcreamgray!16}\ms{13.8}{0.0} & \cellcolor{tabcreamgray!1}\ms{\phantom{0}0.5}{0.0} & \cellcolor{tabcreamgray!0}\ms{\phantom{0}0.0}{0.0} & \cellcolor{tabcreamgray!3}\ms{\phantom{0}2.9}{0.0} \\
WildDet3D & \cellcolor{tabgreenfig!28}\ms{14.3}{0.0} & \cellcolor{tabcreamgray!15}\ms{13.1}{0.0} & \cellcolor{tabcreamgray!47}\ms{42.1}{0.0} & \cellcolor{tabcreamgray!1}\ms{\phantom{0}1.3}{0.0} & \cellcolor{tabcreamgray!29}\ms{26.0}{0.0} & \cellcolor{tabcreamgray!46}\ms{40.5}{0.0} & \cellcolor{tabcreamgray!2}\ms{\phantom{0}1.4}{0.0} & \cellcolor{tabcreamgray!0}\ms{\phantom{0}0.2}{0.0} & \cellcolor{tabcreamgray!0}\ms{\phantom{0}0.0}{0.0} & \cellcolor{tabcreamgray!4}\ms{\phantom{0}3.6}{0.0} \\
SAM 3D & \cellcolor{tabgreenfig!5}\ms{\phantom{0}2.4}{0.1} & \cellcolor{tabcreamgray!3}\ms{\phantom{0}2.3}{0.1} & \cellcolor{tabcreamgray!7}\ms{\phantom{0}5.9}{0.4} & \cellcolor{tabcreamgray!1}\ms{\phantom{0}1.1}{0.0} & \cellcolor{tabcreamgray!4}\ms{\phantom{0}3.7}{0.5} & \cellcolor{tabcreamgray!2}\ms{\phantom{0}1.5}{0.0} & \cellcolor{tabcreamgray!8}\ms{\phantom{0}7.0}{0.2} & \cellcolor{tabcreamgray!0}\ms{\phantom{0}0.0}{0.0} & \cellcolor{tabcreamgray!0}\ms{\phantom{0}0.0}{0.0} & \cellcolor{tabcreamgray!0}\ms{\phantom{0}0.0}{0.0} \\
\dashedmidrule{11}
Boxer + sensor depth & -- & -- & -- & \cellcolor{tabcreamgray!92}\ms{\textbf{82.1}}{0.0} & \cellcolor{tabcreamgray!89}\ms{\textbf{79.3}}{0.3} & \cellcolor{tabcreamgray!100}\ms{\textbf{88.9}}{0.4} & \cellcolor{tabcreamgray!90}\ms{\textbf{79.6}}{0.2} & \cellcolor{tabcreamgray!87}\ms{\textbf{77.2}}{0.3} & \cellcolor{tabcreamgray!79}\ms{\textbf{70.3}}{0.1} & -- \\
WildDet3D + sensor depth & -- & -- & -- & \cellcolor{tabcreamgray!54}\ms{\underline{48.4}}{0.0} & \cellcolor{tabcreamgray!17}\ms{14.7}{0.0} & \cellcolor{tabcreamgray!84}\ms{\underline{74.3}}{0.0} & \cellcolor{tabcreamgray!61}\ms{\underline{54.0}}{0.0} & \cellcolor{tabcreamgray!9}\ms{\phantom{0}\underline{8.0}}{0.0} & \cellcolor{tabcreamgray!0}\ms{\phantom{0}0.0}{0.0} & -- \\
\bottomrule
\end{tabularx}
\setlength{\abovecaptionskip}{7pt}
\caption{\textbf{3D object grounding with ground-truth 2D boxes (\APncd in \%).} The specialist methods receive the ground-truth 2D box of each target, an oracle input used in the 2D-box and depth ablations (Sec.~\ref{sec:ablations_main}). The overall score is the average of the per-dataset scores. The rows with sensor depth cover only the six datasets that provide it and have no overall score. Tab.~\ref{tab:results_ncd} reports the scores without oracle inputs.
}
\label{tab:results_ncd_oracle}
\end{table*}

\bigskip
\begin{table*}[tp]
\centering
\scriptsize
\setlength{\tabcolsep}{0pt}
\renewcommand{\arraystretch}{1.2}  %
\begin{tabularx}{\textwidth}{@{}>{\raggedright\arraybackslash}p{\methodcol}@{\hspace{6pt}}C|*{9}{C}@{}}
\toprule
Method & \APthreeD~$\uparrow$ & HOT3D & HANDAL & HOPEv2 & YCB-V & LM & HB & T-LESS & ITODD & IPD \\
\midrule
Boxer + UniDepthV2 & \cellcolor{tabgreenfig!60}\ms{\textbf{3.2}}{0.0} & \cellcolor{tabcreamgray!3}\ms{2.0}{0.0} & \cellcolor{tabcreamgray!3}\ms{2.0}{0.1} & \cellcolor{tabcreamgray!2}\ms{\phantom{0}1.5}{0.0} & \cellcolor{tabcreamgray!16}\ms{\underline{10.2}}{0.0} & \cellcolor{tabcreamgray!19}\ms{12.1}{0.1} & \cellcolor{tabcreamgray!1}\ms{\phantom{0}0.7}{0.0} & \cellcolor{tabcreamgray!0}\ms{\phantom{0}\underline{0.3}}{0.0} & \cellcolor{tabcreamgray!0}\ms{0.0}{0.0} & \cellcolor{tabcreamgray!0}\ms{0.0}{0.0} \\
WildDet3D + DA3 & \cellcolor{tabgreenfig!51}\ms{\underline{2.7}}{0.0} & \cellcolor{tabcreamgray!6}\ms{3.9}{0.0} & \cellcolor{tabcreamgray!3}\ms{2.1}{0.0} & \cellcolor{tabcreamgray!1}\ms{\phantom{0}0.9}{0.0} & \cellcolor{tabcreamgray!0}\ms{\phantom{0}0.0}{0.0} & \cellcolor{tabcreamgray!26}\ms{\underline{16.6}}{0.0} & \cellcolor{tabcreamgray!0}\ms{\phantom{0}0.2}{0.0} & \cellcolor{tabcreamgray!0}\ms{\phantom{0}0.0}{0.0} & \cellcolor{tabcreamgray!0}\ms{0.0}{0.0} & \cellcolor{tabcreamgray!1}\ms{\textbf{0.8}}{0.0} \\
Boxer + DA3 & \cellcolor{tabgreenfig!49}\ms{2.6}{0.0} & \cellcolor{tabcreamgray!4}\ms{2.8}{0.0} & \cellcolor{tabcreamgray!6}\ms{\textbf{3.6}}{0.0} & \cellcolor{tabcreamgray!6}\ms{\phantom{0}4.0}{0.0} & \cellcolor{tabcreamgray!5}\ms{\phantom{0}3.0}{0.1} & \cellcolor{tabcreamgray!12}\ms{\phantom{0}7.6}{0.1} & \cellcolor{tabcreamgray!4}\ms{\phantom{0}2.5}{0.0} & \cellcolor{tabcreamgray!0}\ms{\phantom{0}0.0}{0.0} & \cellcolor{tabcreamgray!0}\ms{\underline{0.2}}{0.0} & \cellcolor{tabcreamgray!0}\ms{0.2}{0.0} \\
WildDet3D + UniDepthV2 & \cellcolor{tabgreenfig!49}\ms{2.6}{0.0} & \cellcolor{tabcreamgray!12}\ms{\textbf{7.7}}{0.0} & \cellcolor{tabcreamgray!2}\ms{1.1}{0.0} & \cellcolor{tabcreamgray!1}\ms{\phantom{0}0.8}{0.0} & \cellcolor{tabcreamgray!0}\ms{\phantom{0}0.1}{0.0} & \cellcolor{tabcreamgray!21}\ms{13.3}{0.0} & \cellcolor{tabcreamgray!0}\ms{\phantom{0}0.1}{0.0} & \cellcolor{tabcreamgray!0}\ms{\phantom{0}0.0}{0.0} & \cellcolor{tabcreamgray!0}\ms{0.0}{0.0} & \cellcolor{tabcreamgray!0}\ms{0.1}{0.0} \\
Boxer & \cellcolor{tabgreenfig!42}\ms{2.2}{0.0} & \cellcolor{tabcreamgray!10}\ms{\underline{6.6}}{0.0} & \cellcolor{tabcreamgray!5}\ms{\underline{3.0}}{0.0} & \cellcolor{tabcreamgray!5}\ms{\phantom{0}3.1}{0.0} & \cellcolor{tabcreamgray!0}\ms{\phantom{0}0.3}{0.0} & \cellcolor{tabcreamgray!8}\ms{\phantom{0}5.2}{0.0} & \cellcolor{tabcreamgray!2}\ms{\phantom{0}1.4}{0.0} & \cellcolor{tabcreamgray!0}\ms{\phantom{0}0.0}{0.0} & \cellcolor{tabcreamgray!0}\ms{0.0}{0.0} & \cellcolor{tabcreamgray!1}\ms{\underline{0.4}}{0.0} \\
WildDet3D & \cellcolor{tabgreenfig!8}\ms{0.4}{0.0} & \cellcolor{tabcreamgray!0}\ms{0.1}{0.0} & \cellcolor{tabcreamgray!3}\ms{1.6}{0.0} & \cellcolor{tabcreamgray!0}\ms{\phantom{0}0.0}{0.0} & \cellcolor{tabcreamgray!1}\ms{\phantom{0}0.5}{0.0} & \cellcolor{tabcreamgray!3}\ms{\phantom{0}1.6}{0.0} & \cellcolor{tabcreamgray!0}\ms{\phantom{0}0.0}{0.0} & \cellcolor{tabcreamgray!0}\ms{\phantom{0}0.0}{0.0} & \cellcolor{tabcreamgray!0}\ms{0.0}{0.0} & \cellcolor{tabcreamgray!0}\ms{0.2}{0.0} \\
SAM 3D & \cellcolor{tabgreenfig!0}\ms{0.0}{0.0} & \cellcolor{tabcreamgray!0}\ms{0.0}{0.0} & \cellcolor{tabcreamgray!0}\ms{0.0}{0.0} & \cellcolor{tabcreamgray!0}\ms{\phantom{0}0.1}{0.0} & \cellcolor{tabcreamgray!0}\ms{\phantom{0}0.0}{0.0} & \cellcolor{tabcreamgray!0}\ms{\phantom{0}0.0}{0.0} & \cellcolor{tabcreamgray!0}\ms{\phantom{0}0.1}{0.0} & \cellcolor{tabcreamgray!0}\ms{\phantom{0}0.0}{0.0} & \cellcolor{tabcreamgray!0}\ms{0.0}{0.0} & \cellcolor{tabcreamgray!0}\ms{0.0}{0.0} \\
\dashedmidrule{11}
Boxer + sensor depth & -- & -- & -- & \cellcolor{tabcreamgray!85}\ms{\textbf{54.3}}{0.2} & \cellcolor{tabcreamgray!76}\ms{\textbf{48.5}}{1.0} & \cellcolor{tabcreamgray!100}\ms{\textbf{63.6}}{0.4} & \cellcolor{tabcreamgray!70}\ms{\textbf{44.8}}{0.2} & \cellcolor{tabcreamgray!57}\ms{\textbf{36.3}}{0.1} & \cellcolor{tabcreamgray!15}\ms{\textbf{9.3}}{0.1} & -- \\
WildDet3D + sensor depth & -- & -- & -- & \cellcolor{tabcreamgray!7}\ms{\phantom{0}\underline{4.2}}{0.0} & \cellcolor{tabcreamgray!0}\ms{\phantom{0}0.0}{0.0} & \cellcolor{tabcreamgray!19}\ms{11.9}{0.0} & \cellcolor{tabcreamgray!6}\ms{\phantom{0}\underline{3.8}}{0.0} & \cellcolor{tabcreamgray!0}\ms{\phantom{0}0.0}{0.0} & \cellcolor{tabcreamgray!0}\ms{0.0}{0.0} & -- \\
\bottomrule
\end{tabularx}
\setlength{\abovecaptionskip}{7pt}
\caption{\textbf{3D object grounding with ground-truth 2D boxes (\APthreeD in \%).} The specialist methods receive the ground-truth 2D box of each target, an oracle input used in the 2D-box and depth ablations (Sec.~\ref{sec:ablations_main}). The overall score is the average of the per-dataset scores. The rows with sensor depth cover only the six datasets that provide it and have no overall score. Tab.~\ref{tab:results_3d} reports the scores without oracle inputs.
}
\label{tab:results_3d_oracle}
\end{table*}

\endgroup
\end{table*}
\begin{table*}[tp]  %
\begingroup
\renewenvironment{table*}[1][]{}{}%
\begin{table*}[tp]
\centering
\scriptsize
\setlength{\tabcolsep}{0pt}
\renewcommand{\arraystretch}{1.0}  %
\begin{tabularx}{\textwidth}{@{}>{\raggedright\arraybackslash}p{\methodcol}@{\hspace{6pt}}C|*{10}{C}@{}}
\toprule
& & \multicolumn{10}{c}{\NCD threshold $\delta$} \\
Method & \APncd~$\uparrow$ & $0.2$ & $0.4$ & $0.6$ & $0.8$ & $1.0$ & $1.2$ & $1.4$ & $1.6$ & $1.8$ & $2.0$ \\
\midrule
\multicolumn{12}{@{}l}{\textit{Proprietary VLMs}} \\
Gemini 4 Carbon & \cellcolor{tabgreenfig!60}\ms{\textbf{60.3}}{0.4} & \cellcolor{tabcreamgray!5}\ms{\phantom{0}\textbf{3.8}}{0.1} & \cellcolor{tabcreamgray!28}\ms{\textbf{23.1}}{1.7} & \cellcolor{tabcreamgray!58}\ms{\textbf{47.9}}{0.7} & \cellcolor{tabcreamgray!75}\ms{\textbf{61.8}}{1.8} & \cellcolor{tabcreamgray!85}\ms{\textbf{69.5}}{0.5} & \cellcolor{tabcreamgray!91}\ms{\textbf{74.7}}{0.7} & \cellcolor{tabcreamgray!95}\ms{\textbf{78.2}}{0.6} & \cellcolor{tabcreamgray!97}\ms{\textbf{80.1}}{0.5} & \cellcolor{tabcreamgray!99}\ms{\textbf{81.5}}{0.3} & \cellcolor{tabcreamgray!100}\ms{\textbf{82.2}}{0.2} \\
Claude Opus 5.5 & \cellcolor{tabgreenfig!47}\ms{\underline{46.5}}{0.2} & \cellcolor{tabcreamgray!0}\ms{\phantom{0}0.2}{0.0} & \cellcolor{tabcreamgray!12}\ms{\phantom{0}\cmark{n}{u}{9.7}}{0.4} & \cellcolor{tabcreamgray!32}\ms{\cmark{n}{u}{26.7}}{0.4} & \cellcolor{tabcreamgray!47}\ms{\cmark{n}{u}{38.6}}{0.6} & \cellcolor{tabcreamgray!60}\ms{49.2}{1.0} & \cellcolor{tabcreamgray!70}\ms{57.6}{0.5} & \cellcolor{tabcreamgray!79}\ms{64.6}{0.5} & \cellcolor{tabcreamgray!84}\ms{\underline{69.3}}{0.1} & \cellcolor{tabcreamgray!89}\ms{\underline{73.2}}{0.6} & \cellcolor{tabcreamgray!92}\ms{76.0}{0.6} \\
Gemini 4 Argon & \cellcolor{tabgreenfig!46}\ms{45.7}{1.0} & \cellcolor{tabcreamgray!0}\ms{\phantom{0}0.2}{0.1} & \cellcolor{tabcreamgray!6}\ms{\phantom{0}5.1}{1.2} & \cellcolor{tabcreamgray!28}\ms{22.7}{2.3} & \cellcolor{tabcreamgray!45}\ms{37.1}{1.2} & \cellcolor{tabcreamgray!61}\ms{\underline{50.3}}{1.9} & \cellcolor{tabcreamgray!71}\ms{\underline{58.5}}{1.3} & \cellcolor{tabcreamgray!79}\ms{\underline{64.7}}{0.7} & \cellcolor{tabcreamgray!84}\ms{\underline{69.3}}{0.7} & \cellcolor{tabcreamgray!89}\ms{72.9}{0.9} & \cellcolor{tabcreamgray!93}\ms{\underline{76.1}}{0.9} \\
GPT-6 Astra & \cellcolor{tabgreenfig!40}\ms{39.4}{0.7} & \cellcolor{tabcreamgray!1}\ms{\phantom{0}\cmark{n}{u}{0.6}}{0.1} & \cellcolor{tabcreamgray!9}\ms{\phantom{0}7.0}{0.3} & \cellcolor{tabcreamgray!22}\ms{18.4}{1.2} & \cellcolor{tabcreamgray!35}\ms{29.0}{0.5} & \cellcolor{tabcreamgray!49}\ms{40.0}{1.1} & \cellcolor{tabcreamgray!59}\ms{48.7}{1.4} & \cellcolor{tabcreamgray!67}\ms{54.8}{1.1} & \cellcolor{tabcreamgray!73}\ms{60.2}{1.0} & \cellcolor{tabcreamgray!79}\ms{65.2}{0.5} & \cellcolor{tabcreamgray!85}\ms{70.0}{0.2} \\
GPT-6.1 Sol & \cellcolor{tabgreenfig!39}\ms{38.4}{1.2} & \cellcolor{tabcreamgray!1}\ms{\phantom{0}0.5}{0.1} & \cellcolor{tabcreamgray!8}\ms{\phantom{0}6.7}{0.5} & \cellcolor{tabcreamgray!21}\ms{17.5}{1.9} & \cellcolor{tabcreamgray!34}\ms{28.2}{1.9} & \cellcolor{tabcreamgray!46}\ms{38.1}{2.7} & \cellcolor{tabcreamgray!56}\ms{46.4}{2.4} & \cellcolor{tabcreamgray!65}\ms{53.8}{2.1} & \cellcolor{tabcreamgray!73}\ms{59.9}{1.6} & \cellcolor{tabcreamgray!79}\ms{64.6}{0.8} & \cellcolor{tabcreamgray!83}\ms{68.2}{1.2} \\
GPT-5.5 & \cellcolor{tabgreenfig!33}\ms{32.5}{0.4} & \cellcolor{tabcreamgray!0}\ms{\phantom{0}0.1}{0.0} & \cellcolor{tabcreamgray!4}\ms{\phantom{0}3.3}{0.5} & \cellcolor{tabcreamgray!14}\ms{11.9}{0.9} & \cellcolor{tabcreamgray!26}\ms{21.3}{0.5} & \cellcolor{tabcreamgray!38}\ms{31.0}{0.9} & \cellcolor{tabcreamgray!47}\ms{38.5}{1.0} & \cellcolor{tabcreamgray!56}\ms{46.3}{0.6} & \cellcolor{tabcreamgray!64}\ms{52.5}{0.6} & \cellcolor{tabcreamgray!70}\ms{57.8}{0.5} & \cellcolor{tabcreamgray!76}\ms{62.2}{1.0} \\
Gemini 3.8 Flash & \cellcolor{tabgreenfig!30}\ms{29.6}{0.7} & \cellcolor{tabcreamgray!0}\ms{\phantom{0}0.0}{0.0} & \cellcolor{tabcreamgray!2}\ms{\phantom{0}1.4}{0.2} & \cellcolor{tabcreamgray!9}\ms{\phantom{0}7.5}{0.4} & \cellcolor{tabcreamgray!20}\ms{16.6}{1.6} & \cellcolor{tabcreamgray!32}\ms{26.7}{1.9} & \cellcolor{tabcreamgray!44}\ms{35.8}{1.4} & \cellcolor{tabcreamgray!53}\ms{43.7}{0.9} & \cellcolor{tabcreamgray!60}\ms{49.6}{0.1} & \cellcolor{tabcreamgray!67}\ms{54.7}{0.4} & \cellcolor{tabcreamgray!72}\ms{59.5}{0.7} \\
Gemini Robotics ER 2 & \cellcolor{tabgreenfig!29}\ms{28.8}{0.4} & \cellcolor{tabcreamgray!0}\ms{\phantom{0}0.0}{0.0} & \cellcolor{tabcreamgray!1}\ms{\phantom{0}0.9}{0.1} & \cellcolor{tabcreamgray!9}\ms{\phantom{0}7.0}{0.3} & \cellcolor{tabcreamgray!19}\ms{15.6}{0.4} & \cellcolor{tabcreamgray!29}\ms{23.9}{0.5} & \cellcolor{tabcreamgray!40}\ms{32.9}{0.4} & \cellcolor{tabcreamgray!51}\ms{41.6}{0.6} & \cellcolor{tabcreamgray!59}\ms{48.7}{1.1} & \cellcolor{tabcreamgray!68}\ms{56.3}{1.6} & \cellcolor{tabcreamgray!74}\ms{61.1}{1.0} \\
Gemini 3.1 Pro & \cellcolor{tabgreenfig!25}\ms{24.6}{1.0} & \cellcolor{tabcreamgray!0}\ms{\phantom{0}0.1}{0.1} & \cellcolor{tabcreamgray!2}\ms{\phantom{0}2.0}{0.4} & \cellcolor{tabcreamgray!10}\ms{\phantom{0}8.5}{0.6} & \cellcolor{tabcreamgray!18}\ms{15.2}{1.4} & \cellcolor{tabcreamgray!26}\ms{21.6}{1.7} & \cellcolor{tabcreamgray!34}\ms{27.9}{1.9} & \cellcolor{tabcreamgray!41}\ms{34.0}{1.6} & \cellcolor{tabcreamgray!49}\ms{39.9}{1.0} & \cellcolor{tabcreamgray!56}\ms{46.0}{1.3} & \cellcolor{tabcreamgray!62}\ms{51.1}{0.7} \\
GPT-5.4 & \cellcolor{tabgreenfig!25}\ms{24.5}{0.4} & \cellcolor{tabcreamgray!0}\ms{\phantom{0}0.0}{0.0} & \cellcolor{tabcreamgray!0}\ms{\phantom{0}0.2}{0.1} & \cellcolor{tabcreamgray!4}\ms{\phantom{0}3.0}{0.2} & \cellcolor{tabcreamgray!12}\ms{\phantom{0}9.7}{0.4} & \cellcolor{tabcreamgray!22}\ms{18.1}{1.1} & \cellcolor{tabcreamgray!33}\ms{27.0}{1.4} & \cellcolor{tabcreamgray!43}\ms{35.7}{0.3} & \cellcolor{tabcreamgray!53}\ms{43.4}{0.5} & \cellcolor{tabcreamgray!61}\ms{50.5}{0.1} & \cellcolor{tabcreamgray!69}\ms{57.0}{0.3} \\
Claude Opus 4.6 & \cellcolor{tabgreenfig!18}\ms{18.4}{0.1} & \cellcolor{tabcreamgray!0}\ms{\phantom{0}0.0}{0.0} & \cellcolor{tabcreamgray!0}\ms{\phantom{0}0.1}{0.0} & \cellcolor{tabcreamgray!2}\ms{\phantom{0}1.3}{0.1} & \cellcolor{tabcreamgray!6}\ms{\phantom{0}5.1}{0.7} & \cellcolor{tabcreamgray!15}\ms{12.1}{0.7} & \cellcolor{tabcreamgray!24}\ms{19.4}{0.9} & \cellcolor{tabcreamgray!33}\ms{26.8}{0.4} & \cellcolor{tabcreamgray!42}\ms{34.7}{0.6} & \cellcolor{tabcreamgray!48}\ms{39.8}{0.9} & \cellcolor{tabcreamgray!55}\ms{45.0}{0.1} \\
Claude Opus 4.7 & \cellcolor{tabgreenfig!18}\ms{18.3}{0.0} & \cellcolor{tabcreamgray!0}\ms{\phantom{0}0.0}{0.0} & \cellcolor{tabcreamgray!0}\ms{\phantom{0}0.1}{0.0} & \cellcolor{tabcreamgray!2}\ms{\phantom{0}1.4}{0.1} & \cellcolor{tabcreamgray!7}\ms{\phantom{0}5.8}{0.4} & \cellcolor{tabcreamgray!15}\ms{12.0}{0.7} & \cellcolor{tabcreamgray!24}\ms{19.6}{0.6} & \cellcolor{tabcreamgray!32}\ms{26.1}{0.4} & \cellcolor{tabcreamgray!41}\ms{33.6}{0.5} & \cellcolor{tabcreamgray!48}\ms{39.6}{0.5} & \cellcolor{tabcreamgray!55}\ms{44.8}{0.2} \\
\dashedmidrule{12}
\multicolumn{12}{@{}l}{\textit{Open-weight VLMs}} \\
Qwen3.8 Flash Next & \cellcolor{tabgreenfig!26}\ms{26.1}{0.5} & \cellcolor{tabcreamgray!0}\ms{\phantom{0}0.0}{0.0} & \cellcolor{tabcreamgray!1}\ms{\phantom{0}0.8}{0.1} & \cellcolor{tabcreamgray!8}\ms{\phantom{0}6.5}{0.4} & \cellcolor{tabcreamgray!19}\ms{15.4}{1.3} & \cellcolor{tabcreamgray!28}\ms{23.0}{1.6} & \cellcolor{tabcreamgray!37}\ms{30.4}{1.7} & \cellcolor{tabcreamgray!46}\ms{37.9}{1.3} & \cellcolor{tabcreamgray!53}\ms{43.8}{0.9} & \cellcolor{tabcreamgray!59}\ms{48.8}{1.0} & \cellcolor{tabcreamgray!66}\ms{54.1}{1.1} \\
Qwen3-VL & \cellcolor{tabgreenfig!23}\ms{23.4}{0.3} & \cellcolor{tabcreamgray!0}\ms{\phantom{0}0.0}{0.0} & \cellcolor{tabcreamgray!1}\ms{\phantom{0}1.0}{0.0} & \cellcolor{tabcreamgray!9}\ms{\phantom{0}7.4}{0.5} & \cellcolor{tabcreamgray!19}\ms{15.7}{0.2} & \cellcolor{tabcreamgray!27}\ms{22.3}{0.3} & \cellcolor{tabcreamgray!34}\ms{27.9}{0.4} & \cellcolor{tabcreamgray!41}\ms{33.9}{0.5} & \cellcolor{tabcreamgray!46}\ms{37.5}{0.3} & \cellcolor{tabcreamgray!51}\ms{42.1}{0.4} & \cellcolor{tabcreamgray!56}\ms{45.9}{0.8} \\
Qwen3.5 & \cellcolor{tabgreenfig!23}\ms{22.8}{0.1} & \cellcolor{tabcreamgray!0}\ms{\phantom{0}0.0}{0.0} & \cellcolor{tabcreamgray!0}\ms{\phantom{0}0.4}{0.1} & \cellcolor{tabcreamgray!5}\ms{\phantom{0}4.3}{0.4} & \cellcolor{tabcreamgray!14}\ms{11.4}{0.6} & \cellcolor{tabcreamgray!23}\ms{19.3}{0.3} & \cellcolor{tabcreamgray!32}\ms{26.5}{0.7} & \cellcolor{tabcreamgray!41}\ms{33.7}{0.8} & \cellcolor{tabcreamgray!48}\ms{39.6}{0.5} & \cellcolor{tabcreamgray!54}\ms{44.2}{0.5} & \cellcolor{tabcreamgray!60}\ms{49.0}{0.5} \\
Gemma 4 & \cellcolor{tabgreenfig!14}\ms{14.2}{0.7} & \cellcolor{tabcreamgray!0}\ms{\phantom{0}0.0}{0.0} & \cellcolor{tabcreamgray!0}\ms{\phantom{0}0.1}{0.0} & \cellcolor{tabcreamgray!1}\ms{\phantom{0}0.8}{0.1} & \cellcolor{tabcreamgray!4}\ms{\phantom{0}3.2}{0.2} & \cellcolor{tabcreamgray!9}\ms{\phantom{0}7.5}{0.4} & \cellcolor{tabcreamgray!16}\ms{12.9}{1.2} & \cellcolor{tabcreamgray!24}\ms{19.5}{1.1} & \cellcolor{tabcreamgray!32}\ms{26.2}{1.1} & \cellcolor{tabcreamgray!40}\ms{32.6}{1.2} & \cellcolor{tabcreamgray!48}\ms{39.5}{1.5} \\
Qwen3.6 & \cellcolor{tabgreenfig!14}\ms{13.9}{0.2} & \cellcolor{tabcreamgray!0}\ms{\phantom{0}0.0}{0.0} & \cellcolor{tabcreamgray!0}\ms{\phantom{0}0.1}{0.0} & \cellcolor{tabcreamgray!2}\ms{\phantom{0}1.4}{0.1} & \cellcolor{tabcreamgray!7}\ms{\phantom{0}5.4}{0.2} & \cellcolor{tabcreamgray!12}\ms{\phantom{0}9.7}{0.1} & \cellcolor{tabcreamgray!17}\ms{14.3}{0.5} & \cellcolor{tabcreamgray!23}\ms{19.2}{0.3} & \cellcolor{tabcreamgray!30}\ms{24.5}{0.7} & \cellcolor{tabcreamgray!36}\ms{29.6}{1.0} & \cellcolor{tabcreamgray!42}\ms{34.5}{0.6} \\
\dashedmidrule{12}
\multicolumn{12}{@{}l}{\textit{Other methods}} \\
Carbon2D + SAM~3 + DP2 & \cellcolor{tabgreenfig!46}\ms{45.7}{--} & \cellcolor{tabcreamgray!3}\ms{\phantom{0}\underline{2.6}}{--} & \cellcolor{tabcreamgray!22}\ms{\underline{18.3}}{--} & \cellcolor{tabcreamgray!39}\ms{\underline{32.0}}{--} & \cellcolor{tabcreamgray!53}\ms{\underline{43.8}}{--} & \cellcolor{tabcreamgray!60}\ms{49.7}{--} & \cellcolor{tabcreamgray!66}\ms{54.4}{--} & \cellcolor{tabcreamgray!71}\ms{58.4}{--} & \cellcolor{tabcreamgray!77}\ms{63.0}{--} & \cellcolor{tabcreamgray!81}\ms{66.7}{--} & \cellcolor{tabcreamgray!83}\ms{68.5}{--} \\

WildDet3D + SAM~3 + DA3 & \cellcolor{tabgreenfig!20}\ms{19.6}{0.0} & \cellcolor{tabcreamgray!1}\ms{\phantom{0}0.5}{0.0} & \cellcolor{tabcreamgray!5}\ms{\phantom{0}3.9}{0.0} & \cellcolor{tabcreamgray!11}\ms{\phantom{0}8.7}{0.0} & \cellcolor{tabcreamgray!17}\ms{13.8}{0.0} & \cellcolor{tabcreamgray!21}\ms{17.6}{0.0} & \cellcolor{tabcreamgray!28}\ms{23.0}{0.0} & \cellcolor{tabcreamgray!32}\ms{26.7}{0.0} & \cellcolor{tabcreamgray!38}\ms{30.9}{0.0} & \cellcolor{tabcreamgray!41}\ms{33.4}{0.0} & \cellcolor{tabcreamgray!46}\ms{37.9}{0.0} \\
Boxer + SAM~3 + DA3 & \cellcolor{tabgreenfig!13}\ms{12.8}{0.0} & \cellcolor{tabcreamgray!0}\ms{\phantom{0}0.3}{0.0} & \cellcolor{tabcreamgray!2}\ms{\phantom{0}2.0}{0.0} & \cellcolor{tabcreamgray!7}\ms{\phantom{0}5.4}{0.0} & \cellcolor{tabcreamgray!11}\ms{\phantom{0}8.7}{0.1} & \cellcolor{tabcreamgray!15}\ms{12.5}{0.0} & \cellcolor{tabcreamgray!19}\ms{15.9}{0.0} & \cellcolor{tabcreamgray!23}\ms{18.5}{0.1} & \cellcolor{tabcreamgray!25}\ms{20.2}{0.2} & \cellcolor{tabcreamgray!26}\ms{21.5}{0.1} & \cellcolor{tabcreamgray!28}\ms{23.3}{0.2} \\
WildDet3D + SAM~3 & \cellcolor{tabgreenfig!13}\ms{12.8}{0.0} & \cellcolor{tabcreamgray!0}\ms{\phantom{0}0.2}{0.0} & \cellcolor{tabcreamgray!1}\ms{\phantom{0}1.2}{0.0} & \cellcolor{tabcreamgray!4}\ms{\phantom{0}3.4}{0.0} & \cellcolor{tabcreamgray!8}\ms{\phantom{0}6.2}{0.0} & \cellcolor{tabcreamgray!13}\ms{10.8}{0.0} & \cellcolor{tabcreamgray!18}\ms{14.8}{0.0} & \cellcolor{tabcreamgray!22}\ms{17.7}{0.0} & \cellcolor{tabcreamgray!27}\ms{21.8}{0.0} & \cellcolor{tabcreamgray!30}\ms{24.4}{0.0} & \cellcolor{tabcreamgray!33}\ms{27.0}{0.0} \\
Boxer + SAM~3 & \cellcolor{tabgreenfig!12}\ms{11.5}{0.0} & \cellcolor{tabcreamgray!1}\ms{\phantom{0}0.5}{0.0} & \cellcolor{tabcreamgray!3}\ms{\phantom{0}2.3}{0.0} & \cellcolor{tabcreamgray!5}\ms{\phantom{0}4.5}{0.0} & \cellcolor{tabcreamgray!10}\ms{\phantom{0}8.2}{0.0} & \cellcolor{tabcreamgray!14}\ms{11.4}{0.0} & \cellcolor{tabcreamgray!17}\ms{13.9}{0.0} & \cellcolor{tabcreamgray!20}\ms{16.1}{0.0} & \cellcolor{tabcreamgray!21}\ms{17.3}{0.0} & \cellcolor{tabcreamgray!23}\ms{19.3}{0.0} & \cellcolor{tabcreamgray!26}\ms{21.5}{0.0} \\
SAM 3D + SAM~3 & \cellcolor{tabgreenfig!2}\ms{\phantom{0}1.9}{0.1} & \cellcolor{tabcreamgray!0}\ms{\phantom{0}0.0}{0.0} & \cellcolor{tabcreamgray!0}\ms{\phantom{0}0.0}{0.0} & \cellcolor{tabcreamgray!0}\ms{\phantom{0}0.1}{0.0} & \cellcolor{tabcreamgray!0}\ms{\phantom{0}0.3}{0.0} & \cellcolor{tabcreamgray!1}\ms{\phantom{0}0.8}{0.1} & \cellcolor{tabcreamgray!2}\ms{\phantom{0}1.3}{0.0} & \cellcolor{tabcreamgray!3}\ms{\phantom{0}2.2}{0.2} & \cellcolor{tabcreamgray!4}\ms{\phantom{0}3.3}{0.1} & \cellcolor{tabcreamgray!6}\ms{\phantom{0}4.7}{0.2} & \cellcolor{tabcreamgray!7}\ms{\phantom{0}5.9}{0.2} \\
\midrule
\multicolumn{12}{@{}l}{\textit{VLMs given the ground-truth object distance or 2D box (oracle input, Sec.~\ref{sec:ablations_main})}} \\
GPT-6 Astra + distance & \cellcolor{tabgreenfig!60}\ms{\textbf{90.6}}{0.1} & \cellcolor{tabcreamgray!45}\ms{\textbf{36.6}}{1.2} & \cellcolor{tabcreamgray!100}\ms{\textbf{86.7}}{0.2} & \cellcolor{tabcreamgray!100}\ms{\textbf{94.8}}{0.6} & \cellcolor{tabcreamgray!100}\ms{\textbf{96.8}}{0.3} & \cellcolor{tabcreamgray!100}\ms{\textbf{97.5}}{0.3} & \cellcolor{tabcreamgray!100}\ms{\textbf{98.0}}{0.2} & \cellcolor{tabcreamgray!100}\ms{\underline{98.7}}{0.1} & \cellcolor{tabcreamgray!100}\ms{\underline{98.8}}{0.1} & \cellcolor{tabcreamgray!100}\ms{\textbf{98.9}}{0.1} & \cellcolor{tabcreamgray!100}\ms{\underline{98.9}}{0.1} \\
Gemini 4 Carbon + distance & \cellcolor{tabgreenfig!60}\ms{\underline{87.5}}{0.8} & \cellcolor{tabcreamgray!33}\ms{\underline{27.0}}{1.5} & \cellcolor{tabcreamgray!93}\ms{76.8}{0.7} & \cellcolor{tabcreamgray!100}\ms{\underline{94.6}}{0.7} & \cellcolor{tabcreamgray!100}\ms{96.0}{0.5} & \cellcolor{tabcreamgray!100}\ms{96.3}{0.6} & \cellcolor{tabcreamgray!100}\ms{\underline{96.4}}{0.5} & \cellcolor{tabcreamgray!100}\ms{96.7}{0.7} & \cellcolor{tabcreamgray!100}\ms{97.0}{0.9} & \cellcolor{tabcreamgray!100}\ms{\underline{97.0}}{0.9} & \cellcolor{tabcreamgray!100}\ms{97.0}{0.9} \\
Claude Opus 5.5 + distance & \cellcolor{tabgreenfig!60}\ms{\underline{87.5}}{0.4} & \cellcolor{tabcreamgray!20}\ms{16.3}{0.6} & \cellcolor{tabcreamgray!94}\ms{\underline{76.9}}{1.4} & \cellcolor{tabcreamgray!100}\ms{94.1}{0.4} & \cellcolor{tabcreamgray!100}\ms{\underline{96.7}}{0.3} & \cellcolor{tabcreamgray!100}\ms{\underline{97.3}}{0.3} & \cellcolor{tabcreamgray!100}\ms{\textbf{98.0}}{0.4} & \cellcolor{tabcreamgray!100}\ms{\textbf{98.8}}{0.2} & \cellcolor{tabcreamgray!100}\ms{\textbf{98.9}}{0.3} & \cellcolor{tabcreamgray!100}\ms{\textbf{98.9}}{0.2} & \cellcolor{tabcreamgray!100}\ms{\textbf{99.0}}{0.3} \\
Gemini 4 Carbon + 2D box & \cellcolor{tabgreenfig!60}\ms{63.4}{0.5} & \cellcolor{tabcreamgray!6}\ms{\phantom{0}5.3}{0.8} & \cellcolor{tabcreamgray!32}\ms{25.9}{1.3} & \cellcolor{tabcreamgray!64}\ms{52.8}{1.2} & \cellcolor{tabcreamgray!81}\ms{66.2}{0.6} & \cellcolor{tabcreamgray!90}\ms{73.9}{0.8} & \cellcolor{tabcreamgray!95}\ms{78.1}{0.6} & \cellcolor{tabcreamgray!98}\ms{80.8}{0.5} & \cellcolor{tabcreamgray!100}\ms{82.7}{0.7} & \cellcolor{tabcreamgray!100}\ms{83.6}{0.7} & \cellcolor{tabcreamgray!100}\ms{84.5}{0.5} \\
Claude Opus 5.5 + 2D box & \cellcolor{tabgreenfig!49}\ms{48.6}{0.2} & \cellcolor{tabcreamgray!1}\ms{\phantom{0}0.7}{0.2} & \cellcolor{tabcreamgray!14}\ms{11.7}{0.5} & \cellcolor{tabcreamgray!35}\ms{28.8}{0.4} & \cellcolor{tabcreamgray!51}\ms{41.9}{1.3} & \cellcolor{tabcreamgray!65}\ms{53.5}{1.0} & \cellcolor{tabcreamgray!74}\ms{61.0}{1.1} & \cellcolor{tabcreamgray!81}\ms{66.3}{0.7} & \cellcolor{tabcreamgray!86}\ms{70.4}{0.5} & \cellcolor{tabcreamgray!90}\ms{73.7}{0.7} & \cellcolor{tabcreamgray!94}\ms{77.4}{1.0} \\
GPT-6 Astra + 2D box & \cellcolor{tabgreenfig!43}\ms{43.1}{0.4} & \cellcolor{tabcreamgray!1}\ms{\phantom{0}1.1}{0.2} & \cellcolor{tabcreamgray!11}\ms{\phantom{0}9.4}{0.5} & \cellcolor{tabcreamgray!27}\ms{21.8}{1.0} & \cellcolor{tabcreamgray!41}\ms{33.3}{1.0} & \cellcolor{tabcreamgray!55}\ms{44.9}{1.3} & \cellcolor{tabcreamgray!65}\ms{53.1}{0.9} & \cellcolor{tabcreamgray!73}\ms{60.1}{1.8} & \cellcolor{tabcreamgray!78}\ms{64.5}{1.1} & \cellcolor{tabcreamgray!85}\ms{69.6}{0.1} & \cellcolor{tabcreamgray!90}\ms{73.7}{0.7} \\
\bottomrule
\end{tabularx}
\setlength{\abovecaptionskip}{7pt}
\caption{\textbf{$\text{AP}_{\text{NCD}}\text{@}\delta$ per NCD threshold $\delta$} (average over the 9 datasets). $\text{AP}_{\text{NCD}}$ is the mean over $\delta$, and lower $\delta$ is stricter.}
\label{tab:ap_thresh_ncd}
\end{table*}

\medskip
\begin{table*}[tp]
\centering
\scriptsize
\setlength{\tabcolsep}{0pt}
\renewcommand{\arraystretch}{1.0}  %
\begin{tabularx}{\textwidth}{@{}>{\raggedright\arraybackslash}p{\methodcol}@{\hspace{6pt}}C|*{10}{C}@{}}
\toprule
& & \multicolumn{10}{c}{\IoUthreeD threshold $\tau$} \\
Method & \APthreeD~$\uparrow$ & $0.05$ & $0.10$ & $0.15$ & $0.20$ & $0.25$ & $0.30$ & $0.35$ & $0.40$ & $0.45$ & $0.50$ \\
\midrule
\multicolumn{12}{@{}l}{\textit{Proprietary VLMs}} \\
Gemini 4 Carbon & \cellcolor{tabgreenfig!60}\ms{\textbf{12.4}}{0.5} & \cellcolor{tabcreamgray!100}\ms{\textbf{33.3}}{0.9} & \cellcolor{tabcreamgray!76}\ms{\textbf{25.2}}{0.9} & \cellcolor{tabcreamgray!57}\ms{\textbf{18.9}}{1.3} & \cellcolor{tabcreamgray!44}\ms{\textbf{14.6}}{0.5} & \cellcolor{tabcreamgray!31}\ms{\textbf{10.4}}{0.8} & \cellcolor{tabcreamgray!23}\ms{\phantom{0}\textbf{7.6}}{0.6} & \cellcolor{tabcreamgray!17}\ms{\phantom{0}\textbf{5.5}}{0.5} & \cellcolor{tabcreamgray!12}\ms{\phantom{0}\textbf{4.0}}{0.3} & \cellcolor{tabcreamgray!8}\ms{\phantom{0}\textbf{2.7}}{0.2} & \cellcolor{tabcreamgray!5}\ms{\phantom{0}\textbf{1.6}}{0.2} \\
Claude Opus 5.5 & \cellcolor{tabgreenfig!26}\ms{\phantom{0}\cmark{n}{u}{5.3}}{0.3} & \cellcolor{tabcreamgray!53}\ms{\cmark{n}{u}{17.6}}{0.2} & \cellcolor{tabcreamgray!38}\ms{\cmark{n}{u}{12.6}}{0.3} & \cellcolor{tabcreamgray!26}\ms{\phantom{0}\cmark{n}{u}{8.8}}{0.7} & \cellcolor{tabcreamgray!17}\ms{\phantom{0}\cmark{n}{u}{5.8}}{0.8} & \cellcolor{tabcreamgray!11}\ms{\phantom{0}\cmark{n}{u}{3.7}}{0.4} & \cellcolor{tabcreamgray!6}\ms{\phantom{0}\cmark{n}{u}{2.0}}{0.5} & \cellcolor{tabcreamgray!4}\ms{\phantom{0}\cmark{n}{u}{1.2}}{0.2} & \cellcolor{tabcreamgray!2}\ms{\phantom{0}0.6}{0.2} & \cellcolor{tabcreamgray!1}\ms{\phantom{0}0.3}{0.0} & \cellcolor{tabcreamgray!0}\ms{\phantom{0}0.1}{0.0} \\
Gemini 4 Argon & \cellcolor{tabgreenfig!20}\ms{\phantom{0}4.2}{0.7} & \cellcolor{tabcreamgray!48}\ms{15.9}{1.6} & \cellcolor{tabcreamgray!32}\ms{10.6}{1.7} & \cellcolor{tabcreamgray!21}\ms{\phantom{0}6.9}{1.3} & \cellcolor{tabcreamgray!13}\ms{\phantom{0}4.2}{1.2} & \cellcolor{tabcreamgray!7}\ms{\phantom{0}2.2}{0.6} & \cellcolor{tabcreamgray!3}\ms{\phantom{0}1.1}{0.3} & \cellcolor{tabcreamgray!2}\ms{\phantom{0}0.6}{0.2} & \cellcolor{tabcreamgray!1}\ms{\phantom{0}0.3}{0.1} & \cellcolor{tabcreamgray!1}\ms{\phantom{0}0.2}{0.0} & \cellcolor{tabcreamgray!0}\ms{\phantom{0}0.1}{0.0} \\
GPT-6 Astra & \cellcolor{tabgreenfig!15}\ms{\phantom{0}3.0}{0.1} & \cellcolor{tabcreamgray!29}\ms{\phantom{0}9.7}{0.8} & \cellcolor{tabcreamgray!20}\ms{\phantom{0}6.5}{0.6} & \cellcolor{tabcreamgray!14}\ms{\phantom{0}4.5}{0.3} & \cellcolor{tabcreamgray!9}\ms{\phantom{0}3.1}{0.4} & \cellcolor{tabcreamgray!7}\ms{\phantom{0}2.2}{0.3} & \cellcolor{tabcreamgray!5}\ms{\phantom{0}1.5}{0.2} & \cellcolor{tabcreamgray!3}\ms{\phantom{0}1.1}{0.2} & \cellcolor{tabcreamgray!2}\ms{\phantom{0}\cmark{n}{u}{0.7}}{0.1} & \cellcolor{tabcreamgray!1}\ms{\phantom{0}\cmark{n}{u}{0.4}}{0.1} & \cellcolor{tabcreamgray!1}\ms{\phantom{0}\cmark{n}{u}{0.2}}{0.1} \\
GPT-6.1 Sol & \cellcolor{tabgreenfig!14}\ms{\phantom{0}2.9}{0.4} & \cellcolor{tabcreamgray!29}\ms{\phantom{0}9.6}{1.0} & \cellcolor{tabcreamgray!19}\ms{\phantom{0}6.4}{0.6} & \cellcolor{tabcreamgray!14}\ms{\phantom{0}4.5}{0.7} & \cellcolor{tabcreamgray!10}\ms{\phantom{0}3.2}{0.6} & \cellcolor{tabcreamgray!6}\ms{\phantom{0}2.1}{0.4} & \cellcolor{tabcreamgray!4}\ms{\phantom{0}1.3}{0.3} & \cellcolor{tabcreamgray!3}\ms{\phantom{0}0.9}{0.2} & \cellcolor{tabcreamgray!2}\ms{\phantom{0}0.5}{0.1} & \cellcolor{tabcreamgray!1}\ms{\phantom{0}0.3}{0.0} & \cellcolor{tabcreamgray!0}\ms{\phantom{0}0.1}{0.0} \\
GPT-5.5 & \cellcolor{tabgreenfig!8}\ms{\phantom{0}1.7}{0.1} & \cellcolor{tabcreamgray!20}\ms{\phantom{0}6.5}{0.1} & \cellcolor{tabcreamgray!12}\ms{\phantom{0}4.1}{0.1} & \cellcolor{tabcreamgray!8}\ms{\phantom{0}2.8}{0.2} & \cellcolor{tabcreamgray!5}\ms{\phantom{0}1.7}{0.3} & \cellcolor{tabcreamgray!3}\ms{\phantom{0}0.9}{0.2} & \cellcolor{tabcreamgray!2}\ms{\phantom{0}0.5}{0.1} & \cellcolor{tabcreamgray!1}\ms{\phantom{0}0.2}{0.1} & \cellcolor{tabcreamgray!0}\ms{\phantom{0}0.1}{0.0} & \cellcolor{tabcreamgray!0}\ms{\phantom{0}0.0}{0.0} & \cellcolor{tabcreamgray!0}\ms{\phantom{0}0.0}{0.0} \\
Gemini 3.8 Flash & \cellcolor{tabgreenfig!6}\ms{\phantom{0}1.3}{0.0} & \cellcolor{tabcreamgray!16}\ms{\phantom{0}5.3}{0.1} & \cellcolor{tabcreamgray!10}\ms{\phantom{0}3.2}{0.1} & \cellcolor{tabcreamgray!6}\ms{\phantom{0}2.0}{0.3} & \cellcolor{tabcreamgray!3}\ms{\phantom{0}1.1}{0.1} & \cellcolor{tabcreamgray!2}\ms{\phantom{0}0.6}{0.1} & \cellcolor{tabcreamgray!1}\ms{\phantom{0}0.4}{0.1} & \cellcolor{tabcreamgray!0}\ms{\phantom{0}0.1}{0.0} & \cellcolor{tabcreamgray!0}\ms{\phantom{0}0.1}{0.0} & \cellcolor{tabcreamgray!0}\ms{\phantom{0}0.0}{0.0} & \cellcolor{tabcreamgray!0}\ms{\phantom{0}0.0}{0.0} \\
Gemini 3.1 Pro & \cellcolor{tabgreenfig!5}\ms{\phantom{0}1.1}{0.3} & \cellcolor{tabcreamgray!14}\ms{\phantom{0}4.6}{0.5} & \cellcolor{tabcreamgray!8}\ms{\phantom{0}2.7}{0.4} & \cellcolor{tabcreamgray!5}\ms{\phantom{0}1.6}{0.4} & \cellcolor{tabcreamgray!3}\ms{\phantom{0}1.0}{0.4} & \cellcolor{tabcreamgray!2}\ms{\phantom{0}0.6}{0.4} & \cellcolor{tabcreamgray!1}\ms{\phantom{0}0.4}{0.4} & \cellcolor{tabcreamgray!1}\ms{\phantom{0}0.3}{0.2} & \cellcolor{tabcreamgray!1}\ms{\phantom{0}0.2}{0.2} & \cellcolor{tabcreamgray!0}\ms{\phantom{0}0.1}{0.0} & \cellcolor{tabcreamgray!0}\ms{\phantom{0}0.0}{0.0} \\
Gemini Robotics ER 2 & \cellcolor{tabgreenfig!4}\ms{\phantom{0}0.9}{0.1} & \cellcolor{tabcreamgray!13}\ms{\phantom{0}4.4}{0.2} & \cellcolor{tabcreamgray!6}\ms{\phantom{0}2.1}{0.3} & \cellcolor{tabcreamgray!3}\ms{\phantom{0}1.1}{0.2} & \cellcolor{tabcreamgray!2}\ms{\phantom{0}0.6}{0.2} & \cellcolor{tabcreamgray!1}\ms{\phantom{0}0.2}{0.0} & \cellcolor{tabcreamgray!0}\ms{\phantom{0}0.1}{0.0} & \cellcolor{tabcreamgray!0}\ms{\phantom{0}0.1}{0.0} & \cellcolor{tabcreamgray!0}\ms{\phantom{0}0.0}{0.0} & \cellcolor{tabcreamgray!0}\ms{\phantom{0}0.0}{0.0} & \cellcolor{tabcreamgray!0}\ms{\phantom{0}0.0}{0.0} \\
GPT-5.4 & \cellcolor{tabgreenfig!1}\ms{\phantom{0}0.3}{0.0} & \cellcolor{tabcreamgray!5}\ms{\phantom{0}1.5}{0.1} & \cellcolor{tabcreamgray!2}\ms{\phantom{0}0.8}{0.1} & \cellcolor{tabcreamgray!2}\ms{\phantom{0}0.5}{0.1} & \cellcolor{tabcreamgray!1}\ms{\phantom{0}0.2}{0.1} & \cellcolor{tabcreamgray!0}\ms{\phantom{0}0.1}{0.0} & \cellcolor{tabcreamgray!0}\ms{\phantom{0}0.0}{0.0} & \cellcolor{tabcreamgray!0}\ms{\phantom{0}0.0}{0.0} & \cellcolor{tabcreamgray!0}\ms{\phantom{0}0.0}{0.0} & \cellcolor{tabcreamgray!0}\ms{\phantom{0}0.0}{0.0} & \cellcolor{tabcreamgray!0}\ms{\phantom{0}0.0}{0.0} \\
Claude Opus 4.7 & \cellcolor{tabgreenfig!0}\ms{\phantom{0}0.1}{0.0} & \cellcolor{tabcreamgray!2}\ms{\phantom{0}0.6}{0.0} & \cellcolor{tabcreamgray!1}\ms{\phantom{0}0.2}{0.1} & \cellcolor{tabcreamgray!0}\ms{\phantom{0}0.1}{0.0} & \cellcolor{tabcreamgray!0}\ms{\phantom{0}0.0}{0.0} & \cellcolor{tabcreamgray!0}\ms{\phantom{0}0.0}{0.0} & \cellcolor{tabcreamgray!0}\ms{\phantom{0}0.0}{0.0} & \cellcolor{tabcreamgray!0}\ms{\phantom{0}0.0}{0.0} & \cellcolor{tabcreamgray!0}\ms{\phantom{0}0.0}{0.0} & \cellcolor{tabcreamgray!0}\ms{\phantom{0}0.0}{0.0} & \cellcolor{tabcreamgray!0}\ms{\phantom{0}0.0}{0.0} \\
Claude Opus 4.6 & \cellcolor{tabgreenfig!0}\ms{\phantom{0}0.1}{0.0} & \cellcolor{tabcreamgray!1}\ms{\phantom{0}0.4}{0.0} & \cellcolor{tabcreamgray!1}\ms{\phantom{0}0.2}{0.0} & \cellcolor{tabcreamgray!0}\ms{\phantom{0}0.1}{0.0} & \cellcolor{tabcreamgray!0}\ms{\phantom{0}0.0}{0.0} & \cellcolor{tabcreamgray!0}\ms{\phantom{0}0.0}{0.0} & \cellcolor{tabcreamgray!0}\ms{\phantom{0}0.0}{0.0} & \cellcolor{tabcreamgray!0}\ms{\phantom{0}0.0}{0.0} & \cellcolor{tabcreamgray!0}\ms{\phantom{0}0.0}{0.0} & \cellcolor{tabcreamgray!0}\ms{\phantom{0}0.0}{0.0} & \cellcolor{tabcreamgray!0}\ms{\phantom{0}0.0}{0.0} \\
\dashedmidrule{12}
\multicolumn{12}{@{}l}{\textit{Open-weight VLMs}} \\
Qwen3-VL & \cellcolor{tabgreenfig!6}\ms{\phantom{0}1.2}{0.1} & \cellcolor{tabcreamgray!16}\ms{\phantom{0}5.3}{0.3} & \cellcolor{tabcreamgray!9}\ms{\phantom{0}3.0}{0.2} & \cellcolor{tabcreamgray!6}\ms{\phantom{0}1.9}{0.1} & \cellcolor{tabcreamgray!3}\ms{\phantom{0}1.0}{0.1} & \cellcolor{tabcreamgray!2}\ms{\phantom{0}0.5}{0.1} & \cellcolor{tabcreamgray!1}\ms{\phantom{0}0.2}{0.0} & \cellcolor{tabcreamgray!0}\ms{\phantom{0}0.1}{0.0} & \cellcolor{tabcreamgray!0}\ms{\phantom{0}0.0}{0.0} & \cellcolor{tabcreamgray!0}\ms{\phantom{0}0.0}{0.0} & \cellcolor{tabcreamgray!0}\ms{\phantom{0}0.0}{0.0} \\
Qwen3.8 Flash Next & \cellcolor{tabgreenfig!3}\ms{\phantom{0}0.7}{0.1} & \cellcolor{tabcreamgray!10}\ms{\phantom{0}3.3}{0.2} & \cellcolor{tabcreamgray!5}\ms{\phantom{0}1.8}{0.2} & \cellcolor{tabcreamgray!3}\ms{\phantom{0}0.9}{0.1} & \cellcolor{tabcreamgray!2}\ms{\phantom{0}0.5}{0.0} & \cellcolor{tabcreamgray!1}\ms{\phantom{0}0.2}{0.0} & \cellcolor{tabcreamgray!0}\ms{\phantom{0}0.1}{0.0} & \cellcolor{tabcreamgray!0}\ms{\phantom{0}0.1}{0.0} & \cellcolor{tabcreamgray!0}\ms{\phantom{0}0.0}{0.0} & \cellcolor{tabcreamgray!0}\ms{\phantom{0}0.0}{0.0} & \cellcolor{tabcreamgray!0}\ms{\phantom{0}0.0}{0.0} \\
Qwen3.5 & \cellcolor{tabgreenfig!1}\ms{\phantom{0}0.3}{0.0} & \cellcolor{tabcreamgray!6}\ms{\phantom{0}2.0}{0.2} & \cellcolor{tabcreamgray!3}\ms{\phantom{0}0.9}{0.1} & \cellcolor{tabcreamgray!1}\ms{\phantom{0}0.3}{0.1} & \cellcolor{tabcreamgray!1}\ms{\phantom{0}0.2}{0.1} & \cellcolor{tabcreamgray!0}\ms{\phantom{0}0.1}{0.0} & \cellcolor{tabcreamgray!0}\ms{\phantom{0}0.0}{0.0} & \cellcolor{tabcreamgray!0}\ms{\phantom{0}0.0}{0.0} & \cellcolor{tabcreamgray!0}\ms{\phantom{0}0.0}{0.0} & \cellcolor{tabcreamgray!0}\ms{\phantom{0}0.0}{0.0} & \cellcolor{tabcreamgray!0}\ms{\phantom{0}0.0}{0.0} \\
Qwen3.6 & \cellcolor{tabgreenfig!0}\ms{\phantom{0}0.1}{0.0} & \cellcolor{tabcreamgray!2}\ms{\phantom{0}0.7}{0.0} & \cellcolor{tabcreamgray!1}\ms{\phantom{0}0.2}{0.0} & \cellcolor{tabcreamgray!0}\ms{\phantom{0}0.1}{0.0} & \cellcolor{tabcreamgray!0}\ms{\phantom{0}0.0}{0.0} & \cellcolor{tabcreamgray!0}\ms{\phantom{0}0.0}{0.0} & \cellcolor{tabcreamgray!0}\ms{\phantom{0}0.0}{0.0} & \cellcolor{tabcreamgray!0}\ms{\phantom{0}0.0}{0.0} & \cellcolor{tabcreamgray!0}\ms{\phantom{0}0.0}{0.0} & \cellcolor{tabcreamgray!0}\ms{\phantom{0}0.0}{0.0} & \cellcolor{tabcreamgray!0}\ms{\phantom{0}0.0}{0.0} \\
Gemma 4 & \cellcolor{tabgreenfig!0}\ms{\phantom{0}0.1}{0.0} & \cellcolor{tabcreamgray!1}\ms{\phantom{0}0.3}{0.0} & \cellcolor{tabcreamgray!0}\ms{\phantom{0}0.1}{0.0} & \cellcolor{tabcreamgray!0}\ms{\phantom{0}0.0}{0.0} & \cellcolor{tabcreamgray!0}\ms{\phantom{0}0.0}{0.0} & \cellcolor{tabcreamgray!0}\ms{\phantom{0}0.0}{0.0} & \cellcolor{tabcreamgray!0}\ms{\phantom{0}0.0}{0.0} & \cellcolor{tabcreamgray!0}\ms{\phantom{0}0.0}{0.0} & \cellcolor{tabcreamgray!0}\ms{\phantom{0}0.0}{0.0} & \cellcolor{tabcreamgray!0}\ms{\phantom{0}0.0}{0.0} & \cellcolor{tabcreamgray!0}\ms{\phantom{0}0.0}{0.0} \\
\dashedmidrule{12}
\multicolumn{12}{@{}l}{\textit{Other methods}} \\
Carbon2D + SAM~3 + DP2 & \cellcolor{tabgreenfig!36}\ms{\phantom{0}\underline{7.3}}{--} & \cellcolor{tabcreamgray!57}\ms{\underline{19.1}}{--} & \cellcolor{tabcreamgray!46}\ms{\underline{15.3}}{--} & \cellcolor{tabcreamgray!36}\ms{\underline{12.0}}{--} & \cellcolor{tabcreamgray!26}\ms{\phantom{0}\underline{8.7}}{--} & \cellcolor{tabcreamgray!20}\ms{\phantom{0}\underline{6.7}}{--} & \cellcolor{tabcreamgray!14}\ms{\phantom{0}\underline{4.5}}{--} & \cellcolor{tabcreamgray!10}\ms{\phantom{0}\underline{3.3}}{--} & \cellcolor{tabcreamgray!6}\ms{\phantom{0}\underline{2.0}}{--} & \cellcolor{tabcreamgray!3}\ms{\phantom{0}\underline{1.1}}{--} & \cellcolor{tabcreamgray!2}\ms{\phantom{0}\underline{0.5}}{--} \\

WildDet3D + SAM~3 + DA3 & \cellcolor{tabgreenfig!8}\ms{\phantom{0}1.6}{0.0} & \cellcolor{tabcreamgray!16}\ms{\phantom{0}5.3}{0.0} & \cellcolor{tabcreamgray!11}\ms{\phantom{0}3.6}{0.0} & \cellcolor{tabcreamgray!7}\ms{\phantom{0}2.3}{0.0} & \cellcolor{tabcreamgray!5}\ms{\phantom{0}1.6}{0.0} & \cellcolor{tabcreamgray!3}\ms{\phantom{0}1.1}{0.0} & \cellcolor{tabcreamgray!2}\ms{\phantom{0}0.7}{0.0} & \cellcolor{tabcreamgray!1}\ms{\phantom{0}0.4}{0.0} & \cellcolor{tabcreamgray!1}\ms{\phantom{0}0.3}{0.0} & \cellcolor{tabcreamgray!1}\ms{\phantom{0}0.2}{0.0} & \cellcolor{tabcreamgray!1}\ms{\phantom{0}\cmark{n}{u}{0.2}}{0.0} \\
Boxer + SAM~3 & \cellcolor{tabgreenfig!5}\ms{\phantom{0}1.1}{0.0} & \cellcolor{tabcreamgray!10}\ms{\phantom{0}3.3}{0.0} & \cellcolor{tabcreamgray!7}\ms{\phantom{0}2.2}{0.0} & \cellcolor{tabcreamgray!5}\ms{\phantom{0}1.5}{0.0} & \cellcolor{tabcreamgray!3}\ms{\phantom{0}1.1}{0.0} & \cellcolor{tabcreamgray!2}\ms{\phantom{0}0.8}{0.0} & \cellcolor{tabcreamgray!2}\ms{\phantom{0}0.7}{0.0} & \cellcolor{tabcreamgray!2}\ms{\phantom{0}0.6}{0.0} & \cellcolor{tabcreamgray!1}\ms{\phantom{0}0.4}{0.0} & \cellcolor{tabcreamgray!1}\ms{\phantom{0}0.2}{0.0} & \cellcolor{tabcreamgray!1}\ms{\phantom{0}\cmark{n}{u}{0.2}}{0.0} \\
Boxer + SAM~3 + DA3 & \cellcolor{tabgreenfig!5}\ms{\phantom{0}1.0}{0.0} & \cellcolor{tabcreamgray!10}\ms{\phantom{0}3.4}{0.1} & \cellcolor{tabcreamgray!6}\ms{\phantom{0}1.9}{0.0} & \cellcolor{tabcreamgray!4}\ms{\phantom{0}1.3}{0.1} & \cellcolor{tabcreamgray!3}\ms{\phantom{0}0.9}{0.0} & \cellcolor{tabcreamgray!2}\ms{\phantom{0}0.7}{0.0} & \cellcolor{tabcreamgray!2}\ms{\phantom{0}0.5}{0.0} & \cellcolor{tabcreamgray!1}\ms{\phantom{0}0.4}{0.0} & \cellcolor{tabcreamgray!1}\ms{\phantom{0}0.3}{0.0} & \cellcolor{tabcreamgray!1}\ms{\phantom{0}0.2}{0.0} & \cellcolor{tabcreamgray!0}\ms{\phantom{0}0.1}{0.0} \\
WildDet3D + SAM~3 & \cellcolor{tabgreenfig!2}\ms{\phantom{0}0.4}{0.0} & \cellcolor{tabcreamgray!5}\ms{\phantom{0}1.6}{0.0} & \cellcolor{tabcreamgray!3}\ms{\phantom{0}1.0}{0.0} & \cellcolor{tabcreamgray!2}\ms{\phantom{0}0.7}{0.0} & \cellcolor{tabcreamgray!1}\ms{\phantom{0}0.4}{0.0} & \cellcolor{tabcreamgray!1}\ms{\phantom{0}0.2}{0.0} & \cellcolor{tabcreamgray!1}\ms{\phantom{0}0.2}{0.0} & \cellcolor{tabcreamgray!0}\ms{\phantom{0}0.1}{0.0} & \cellcolor{tabcreamgray!0}\ms{\phantom{0}0.1}{0.0} & \cellcolor{tabcreamgray!0}\ms{\phantom{0}0.1}{0.0} & \cellcolor{tabcreamgray!0}\ms{\phantom{0}0.1}{0.0} \\
SAM 3D + SAM~3 & \cellcolor{tabgreenfig!0}\ms{\phantom{0}0.0}{0.0} & \cellcolor{tabcreamgray!0}\ms{\phantom{0}0.0}{0.0} & \cellcolor{tabcreamgray!0}\ms{\phantom{0}0.0}{0.0} & \cellcolor{tabcreamgray!0}\ms{\phantom{0}0.0}{0.0} & \cellcolor{tabcreamgray!0}\ms{\phantom{0}0.0}{0.0} & \cellcolor{tabcreamgray!0}\ms{\phantom{0}0.0}{0.0} & \cellcolor{tabcreamgray!0}\ms{\phantom{0}0.0}{0.0} & \cellcolor{tabcreamgray!0}\ms{\phantom{0}0.0}{0.0} & \cellcolor{tabcreamgray!0}\ms{\phantom{0}0.0}{0.0} & \cellcolor{tabcreamgray!0}\ms{\phantom{0}0.0}{0.0} & \cellcolor{tabcreamgray!0}\ms{\phantom{0}0.0}{0.0} \\
\midrule
\multicolumn{12}{@{}l}{\textit{VLMs given the ground-truth object distance or 2D box (oracle input, Sec.~\ref{sec:ablations_main})}} \\
GPT-6 Astra + distance & \cellcolor{tabgreenfig!60}\ms{\textbf{60.1}}{0.2} & \cellcolor{tabcreamgray!100}\ms{\underline{88.0}}{0.8} & \cellcolor{tabcreamgray!100}\ms{\underline{82.2}}{0.8} & \cellcolor{tabcreamgray!100}\ms{\textbf{77.8}}{0.9} & \cellcolor{tabcreamgray!100}\ms{\textbf{73.5}}{0.3} & \cellcolor{tabcreamgray!100}\ms{\textbf{67.6}}{0.6} & \cellcolor{tabcreamgray!100}\ms{\textbf{60.4}}{0.4} & \cellcolor{tabcreamgray!100}\ms{\textbf{52.5}}{0.7} & \cellcolor{tabcreamgray!100}\ms{\textbf{42.6}}{1.0} & \cellcolor{tabcreamgray!98}\ms{\underline{32.5}}{1.1} & \cellcolor{tabcreamgray!71}\ms{\underline{23.6}}{0.5} \\
Gemini 4 Carbon + distance & \cellcolor{tabgreenfig!60}\ms{\underline{58.7}}{1.5} & \cellcolor{tabcreamgray!100}\ms{\textbf{91.2}}{1.1} & \cellcolor{tabcreamgray!100}\ms{\textbf{84.3}}{1.5} & \cellcolor{tabcreamgray!100}\ms{\underline{77.2}}{0.7} & \cellcolor{tabcreamgray!100}\ms{\underline{71.1}}{0.9} & \cellcolor{tabcreamgray!100}\ms{\underline{62.6}}{1.5} & \cellcolor{tabcreamgray!100}\ms{\underline{54.9}}{2.1} & \cellcolor{tabcreamgray!100}\ms{\underline{47.8}}{2.6} & \cellcolor{tabcreamgray!100}\ms{\underline{40.2}}{2.5} & \cellcolor{tabcreamgray!99}\ms{\textbf{33.0}}{1.8} & \cellcolor{tabcreamgray!74}\ms{\textbf{24.5}}{1.4} \\
Claude Opus 5.5 + distance & \cellcolor{tabgreenfig!60}\ms{49.5}{0.6} & \cellcolor{tabcreamgray!100}\ms{86.2}{0.8} & \cellcolor{tabcreamgray!100}\ms{78.1}{1.5} & \cellcolor{tabcreamgray!100}\ms{71.4}{1.2} & \cellcolor{tabcreamgray!100}\ms{63.8}{0.5} & \cellcolor{tabcreamgray!100}\ms{54.7}{0.3} & \cellcolor{tabcreamgray!100}\ms{45.2}{0.2} & \cellcolor{tabcreamgray!100}\ms{36.2}{0.7} & \cellcolor{tabcreamgray!83}\ms{27.5}{0.8} & \cellcolor{tabcreamgray!58}\ms{19.2}{0.9} & \cellcolor{tabcreamgray!37}\ms{12.2}{0.3} \\
Gemini 4 Carbon + 2D box & \cellcolor{tabgreenfig!60}\ms{14.2}{1.2} & \cellcolor{tabcreamgray!100}\ms{37.1}{2.7} & \cellcolor{tabcreamgray!86}\ms{28.7}{2.2} & \cellcolor{tabcreamgray!67}\ms{22.2}{1.8} & \cellcolor{tabcreamgray!51}\ms{16.9}{1.1} & \cellcolor{tabcreamgray!36}\ms{11.9}{1.5} & \cellcolor{tabcreamgray!27}\ms{\phantom{0}9.1}{1.2} & \cellcolor{tabcreamgray!19}\ms{\phantom{0}6.3}{1.0} & \cellcolor{tabcreamgray!13}\ms{\phantom{0}4.4}{0.5} & \cellcolor{tabcreamgray!9}\ms{\phantom{0}3.1}{0.4} & \cellcolor{tabcreamgray!6}\ms{\phantom{0}2.1}{0.3} \\
Claude Opus 5.5 + 2D box & \cellcolor{tabgreenfig!28}\ms{\phantom{0}5.8}{0.2} & \cellcolor{tabcreamgray!54}\ms{17.9}{1.1} & \cellcolor{tabcreamgray!38}\ms{12.8}{0.4} & \cellcolor{tabcreamgray!29}\ms{\phantom{0}9.6}{0.4} & \cellcolor{tabcreamgray!21}\ms{\phantom{0}6.9}{0.5} & \cellcolor{tabcreamgray!14}\ms{\phantom{0}4.5}{0.2} & \cellcolor{tabcreamgray!8}\ms{\phantom{0}2.8}{0.2} & \cellcolor{tabcreamgray!5}\ms{\phantom{0}1.7}{0.4} & \cellcolor{tabcreamgray!3}\ms{\phantom{0}0.9}{0.2} & \cellcolor{tabcreamgray!2}\ms{\phantom{0}0.5}{0.1} & \cellcolor{tabcreamgray!1}\ms{\phantom{0}0.3}{0.1} \\
GPT-6 Astra + 2D box & \cellcolor{tabgreenfig!18}\ms{\phantom{0}3.6}{0.2} & \cellcolor{tabcreamgray!32}\ms{10.7}{1.0} & \cellcolor{tabcreamgray!23}\ms{\phantom{0}7.5}{0.5} & \cellcolor{tabcreamgray!16}\ms{\phantom{0}5.4}{0.5} & \cellcolor{tabcreamgray!12}\ms{\phantom{0}3.9}{0.2} & \cellcolor{tabcreamgray!9}\ms{\phantom{0}2.9}{0.4} & \cellcolor{tabcreamgray!6}\ms{\phantom{0}2.1}{0.4} & \cellcolor{tabcreamgray!4}\ms{\phantom{0}1.3}{0.2} & \cellcolor{tabcreamgray!2}\ms{\phantom{0}0.8}{0.1} & \cellcolor{tabcreamgray!2}\ms{\phantom{0}0.6}{0.1} & \cellcolor{tabcreamgray!1}\ms{\phantom{0}0.3}{0.1} \\
\bottomrule
\end{tabularx}
\setlength{\abovecaptionskip}{7pt}
\caption{\textbf{$\text{AP}_{\text{IoU3D}}\text{@}\tau$ per IoU3D threshold $\tau$} (average over the 9 datasets). $\text{AP}_{\text{IoU3D}}$ is the mean over $\tau$, and higher $\tau$ is stricter.}
\label{tab:ap_thresh_iou3d}
\end{table*}

\endgroup
\end{table*}

\section{Results on the validation split}
\label{app:val_results}
Tabs.~\ref{tab:val_results_ncd}, \ref{tab:val_results_3d}, and~\ref{tab:val_results_2d} report \APncd, \APthreeD, and \APtwoD on the validation split for the three ablated VLMs, Gemini~4 Carbon, Claude Opus~5.5, and GPT-6 Astra, evaluated as on the test split. On all three metrics, the three VLMs rank in the same order as on the test split, with overall scores within 2.7\,pp of their test scores; the largest differences are those of GPT-6 Astra ($+2.7$\,pp \APncd and $-2.7$\,pp \APtwoD). 

\begin{table*}[tp]  %
\begingroup
\renewenvironment{table*}[1][]{}{}%
\begin{table*}[tp]
\centering
\scriptsize
\setlength{\tabcolsep}{0pt}
\renewcommand{\arraystretch}{1.0}  %
\begin{tabularx}{\textwidth}{@{}>{\raggedright\arraybackslash}p{\methodcol}@{\hspace{6pt}}C|*{10}{C}@{}}
\toprule
& & \multicolumn{10}{c}{\IoUtwoD threshold $\tau$} \\
Method & \APtwoD~$\uparrow$ & $0.50$ & $0.55$ & $0.60$ & $0.65$ & $0.70$ & $0.75$ & $0.80$ & $0.85$ & $0.90$ & $0.95$ \\
\midrule
\multicolumn{12}{@{}l}{\textit{Proprietary VLMs}} \\
Gemini 4 Carbon & \cellcolor{tabgreenfig!60}\ms{\textbf{72.6}}{0.1} & \cellcolor{tabcreamgray!100}\ms{\textbf{91.8}}{0.2} & \cellcolor{tabcreamgray!99}\ms{\textbf{90.9}}{0.2} & \cellcolor{tabcreamgray!98}\ms{\textbf{89.9}}{0.3} & \cellcolor{tabcreamgray!95}\ms{\textbf{87.5}}{0.1} & \cellcolor{tabcreamgray!93}\ms{\textbf{85.0}}{0.1} & \cellcolor{tabcreamgray!89}\ms{\textbf{81.9}}{0.2} & \cellcolor{tabcreamgray!83}\ms{\textbf{76.0}}{0.5} & \cellcolor{tabcreamgray!73}\ms{\textbf{66.6}}{0.3} & \cellcolor{tabcreamgray!50}\ms{\textbf{46.2}}{0.2} & \cellcolor{tabcreamgray!11}\ms{\textbf{9.9}}{0.4} \\
Gemini 4 Argon & \cellcolor{tabgreenfig!58}\ms{\underline{69.1}}{0.5} & \cellcolor{tabcreamgray!96}\ms{88.2}{0.8} & \cellcolor{tabcreamgray!95}\ms{87.3}{0.8} & \cellcolor{tabcreamgray!94}\ms{86.2}{0.9} & \cellcolor{tabcreamgray!92}\ms{84.2}{0.9} & \cellcolor{tabcreamgray!89}\ms{81.9}{0.7} & \cellcolor{tabcreamgray!85}\ms{\underline{78.3}}{0.9} & \cellcolor{tabcreamgray!79}\ms{\underline{72.8}}{0.6} & \cellcolor{tabcreamgray!68}\ms{\underline{62.0}}{1.1} & \cellcolor{tabcreamgray!46}\ms{\underline{42.1}}{1.4} & \cellcolor{tabcreamgray!9}\ms{8.4}{0.2} \\
Claude Opus 5.5 & \cellcolor{tabgreenfig!58}\ms{69.0}{0.5} & \cellcolor{tabcreamgray!99}\ms{\underline{91.1}}{0.8} & \cellcolor{tabcreamgray!98}\ms{\underline{90.1}}{0.6} & \cellcolor{tabcreamgray!97}\ms{\underline{88.8}}{0.8} & \cellcolor{tabcreamgray!95}\ms{\underline{86.9}}{0.5} & \cellcolor{tabcreamgray!91}\ms{\underline{83.3}}{0.9} & \cellcolor{tabcreamgray!85}\ms{78.1}{0.2} & \cellcolor{tabcreamgray!77}\ms{70.6}{0.7} & \cellcolor{tabcreamgray!62}\ms{57.2}{0.8} & \cellcolor{tabcreamgray!40}\ms{36.7}{0.3} & \cellcolor{tabcreamgray!8}\ms{7.7}{0.4} \\
GPT-6.1 Sol & \cellcolor{tabgreenfig!55}\ms{65.8}{0.6} & \cellcolor{tabcreamgray!93}\ms{85.4}{0.5} & \cellcolor{tabcreamgray!92}\ms{84.6}{0.4} & \cellcolor{tabcreamgray!91}\ms{83.6}{0.7} & \cellcolor{tabcreamgray!90}\ms{82.6}{0.9} & \cellcolor{tabcreamgray!88}\ms{80.5}{1.2} & \cellcolor{tabcreamgray!83}\ms{76.5}{1.2} & \cellcolor{tabcreamgray!76}\ms{69.7}{1.3} & \cellcolor{tabcreamgray!62}\ms{56.8}{1.0} & \cellcolor{tabcreamgray!37}\ms{33.6}{1.1} & \cellcolor{tabcreamgray!6}\ms{5.1}{0.2} \\
GPT-6 Astra & \cellcolor{tabgreenfig!55}\ms{65.5}{0.2} & \cellcolor{tabcreamgray!94}\ms{86.5}{0.1} & \cellcolor{tabcreamgray!93}\ms{85.3}{0.0} & \cellcolor{tabcreamgray!92}\ms{84.5}{0.3} & \cellcolor{tabcreamgray!90}\ms{82.6}{0.4} & \cellcolor{tabcreamgray!88}\ms{80.6}{0.4} & \cellcolor{tabcreamgray!83}\ms{76.4}{0.5} & \cellcolor{tabcreamgray!76}\ms{69.7}{0.6} & \cellcolor{tabcreamgray!60}\ms{54.9}{1.0} & \cellcolor{tabcreamgray!34}\ms{30.9}{0.8} & \cellcolor{tabcreamgray!4}\ms{3.8}{0.2} \\
Gemini Robotics ER 2 & \cellcolor{tabgreenfig!53}\ms{63.1}{0.3} & \cellcolor{tabcreamgray!91}\ms{83.8}{0.3} & \cellcolor{tabcreamgray!90}\ms{82.2}{0.1} & \cellcolor{tabcreamgray!88}\ms{80.4}{0.5} & \cellcolor{tabcreamgray!85}\ms{78.4}{0.6} & \cellcolor{tabcreamgray!83}\ms{75.8}{0.9} & \cellcolor{tabcreamgray!78}\ms{71.2}{0.2} & \cellcolor{tabcreamgray!70}\ms{64.4}{0.6} & \cellcolor{tabcreamgray!59}\ms{54.1}{0.6} & \cellcolor{tabcreamgray!38}\ms{35.2}{0.6} & \cellcolor{tabcreamgray!7}\ms{6.0}{0.4} \\
Gemini 3.8 Flash & \cellcolor{tabgreenfig!52}\ms{61.9}{0.1} & \cellcolor{tabcreamgray!90}\ms{82.9}{0.1} & \cellcolor{tabcreamgray!89}\ms{81.3}{0.1} & \cellcolor{tabcreamgray!87}\ms{80.1}{0.2} & \cellcolor{tabcreamgray!85}\ms{77.6}{0.4} & \cellcolor{tabcreamgray!81}\ms{74.7}{0.4} & \cellcolor{tabcreamgray!77}\ms{70.3}{0.4} & \cellcolor{tabcreamgray!70}\ms{64.1}{0.8} & \cellcolor{tabcreamgray!56}\ms{51.8}{1.3} & \cellcolor{tabcreamgray!34}\ms{31.4}{0.6} & \cellcolor{tabcreamgray!5}\ms{4.8}{0.1} \\
Gemini 3.1 Pro & \cellcolor{tabgreenfig!51}\ms{61.0}{0.4} & \cellcolor{tabcreamgray!87}\ms{79.9}{0.6} & \cellcolor{tabcreamgray!85}\ms{78.4}{0.5} & \cellcolor{tabcreamgray!84}\ms{77.4}{0.7} & \cellcolor{tabcreamgray!82}\ms{75.2}{0.6} & \cellcolor{tabcreamgray!78}\ms{71.6}{0.8} & \cellcolor{tabcreamgray!73}\ms{67.4}{0.6} & \cellcolor{tabcreamgray!67}\ms{61.9}{0.3} & \cellcolor{tabcreamgray!57}\ms{52.6}{0.2} & \cellcolor{tabcreamgray!40}\ms{36.7}{0.4} & \cellcolor{tabcreamgray!10}\ms{\underline{9.3}}{0.4} \\
GPT-5.5 & \cellcolor{tabgreenfig!21}\ms{25.2}{0.5} & \cellcolor{tabcreamgray!60}\ms{54.9}{1.0} & \cellcolor{tabcreamgray!54}\ms{49.8}{1.0} & \cellcolor{tabcreamgray!47}\ms{42.7}{0.4} & \cellcolor{tabcreamgray!38}\ms{35.2}{0.8} & \cellcolor{tabcreamgray!30}\ms{27.7}{0.8} & \cellcolor{tabcreamgray!21}\ms{19.7}{0.5} & \cellcolor{tabcreamgray!14}\ms{13.1}{0.5} & \cellcolor{tabcreamgray!7}\ms{\phantom{0}6.4}{0.6} & \cellcolor{tabcreamgray!3}\ms{\phantom{0}2.3}{0.1} & \cellcolor{tabcreamgray!0}\ms{0.2}{0.0} \\
Claude Opus 4.7 & \cellcolor{tabgreenfig!16}\ms{18.8}{0.2} & \cellcolor{tabcreamgray!53}\ms{49.0}{0.3} & \cellcolor{tabcreamgray!45}\ms{41.4}{0.5} & \cellcolor{tabcreamgray!37}\ms{33.9}{0.1} & \cellcolor{tabcreamgray!29}\ms{26.2}{0.5} & \cellcolor{tabcreamgray!20}\ms{18.3}{0.5} & \cellcolor{tabcreamgray!13}\ms{11.6}{0.4} & \cellcolor{tabcreamgray!6}\ms{\phantom{0}5.8}{0.4} & \cellcolor{tabcreamgray!2}\ms{\phantom{0}2.0}{0.2} & \cellcolor{tabcreamgray!0}\ms{\phantom{0}0.2}{0.0} & \cellcolor{tabcreamgray!0}\ms{0.0}{0.0} \\
GPT-5.4 & \cellcolor{tabgreenfig!4}\ms{\phantom{0}4.6}{0.1} & \cellcolor{tabcreamgray!17}\ms{15.7}{0.2} & \cellcolor{tabcreamgray!12}\ms{11.2}{0.6} & \cellcolor{tabcreamgray!8}\ms{\phantom{0}7.3}{0.2} & \cellcolor{tabcreamgray!5}\ms{\phantom{0}4.8}{0.2} & \cellcolor{tabcreamgray!3}\ms{\phantom{0}3.1}{0.2} & \cellcolor{tabcreamgray!2}\ms{\phantom{0}1.9}{0.0} & \cellcolor{tabcreamgray!1}\ms{\phantom{0}1.1}{0.2} & \cellcolor{tabcreamgray!0}\ms{\phantom{0}0.4}{0.0} & \cellcolor{tabcreamgray!0}\ms{\phantom{0}0.1}{0.0} & \cellcolor{tabcreamgray!0}\ms{0.0}{0.0} \\
Claude Opus 4.6 & \cellcolor{tabgreenfig!1}\ms{\phantom{0}1.3}{0.0} & \cellcolor{tabcreamgray!7}\ms{\phantom{0}6.2}{0.3} & \cellcolor{tabcreamgray!4}\ms{\phantom{0}3.5}{0.2} & \cellcolor{tabcreamgray!2}\ms{\phantom{0}1.8}{0.0} & \cellcolor{tabcreamgray!1}\ms{\phantom{0}0.8}{0.1} & \cellcolor{tabcreamgray!0}\ms{\phantom{0}0.4}{0.0} & \cellcolor{tabcreamgray!0}\ms{\phantom{0}0.1}{0.0} & \cellcolor{tabcreamgray!0}\ms{\phantom{0}0.0}{0.0} & \cellcolor{tabcreamgray!0}\ms{\phantom{0}0.0}{0.0} & \cellcolor{tabcreamgray!0}\ms{\phantom{0}0.0}{0.0} & \cellcolor{tabcreamgray!0}\ms{0.0}{0.0} \\
\dashedmidrule{12}
\multicolumn{12}{@{}l}{\textit{Open-weight VLMs}} \\
Qwen3.8 Flash Next & \cellcolor{tabgreenfig!47}\ms{56.9}{0.4} & \cellcolor{tabcreamgray!83}\ms{76.6}{0.7} & \cellcolor{tabcreamgray!82}\ms{75.0}{0.6} & \cellcolor{tabcreamgray!80}\ms{73.8}{0.4} & \cellcolor{tabcreamgray!78}\ms{71.2}{0.7} & \cellcolor{tabcreamgray!74}\ms{67.8}{0.5} & \cellcolor{tabcreamgray!69}\ms{63.3}{0.7} & \cellcolor{tabcreamgray!62}\ms{57.1}{0.2} & \cellcolor{tabcreamgray!51}\ms{46.7}{0.2} & \cellcolor{tabcreamgray!34}\ms{31.6}{0.5} & \cellcolor{tabcreamgray!7}\ms{6.0}{0.2} \\
Qwen3.6 & \cellcolor{tabgreenfig!38}\ms{45.6}{0.6} & \cellcolor{tabcreamgray!79}\ms{72.2}{1.1} & \cellcolor{tabcreamgray!75}\ms{68.5}{0.8} & \cellcolor{tabcreamgray!71}\ms{64.8}{0.7} & \cellcolor{tabcreamgray!67}\ms{61.5}{1.1} & \cellcolor{tabcreamgray!62}\ms{56.9}{0.3} & \cellcolor{tabcreamgray!54}\ms{49.3}{0.3} & \cellcolor{tabcreamgray!44}\ms{40.6}{0.7} & \cellcolor{tabcreamgray!31}\ms{28.1}{1.4} & \cellcolor{tabcreamgray!14}\ms{13.3}{0.3} & \cellcolor{tabcreamgray!1}\ms{1.3}{0.1} \\
Qwen3-VL & \cellcolor{tabgreenfig!37}\ms{44.5}{0.4} & \cellcolor{tabcreamgray!71}\ms{65.4}{0.9} & \cellcolor{tabcreamgray!69}\ms{63.5}{0.6} & \cellcolor{tabcreamgray!67}\ms{61.5}{0.9} & \cellcolor{tabcreamgray!64}\ms{58.7}{0.6} & \cellcolor{tabcreamgray!60}\ms{55.2}{0.4} & \cellcolor{tabcreamgray!55}\ms{50.3}{0.9} & \cellcolor{tabcreamgray!47}\ms{43.4}{0.7} & \cellcolor{tabcreamgray!34}\ms{31.3}{0.5} & \cellcolor{tabcreamgray!16}\ms{14.7}{0.3} & \cellcolor{tabcreamgray!2}\ms{1.4}{0.1} \\
Gemma 4 & \cellcolor{tabgreenfig!36}\ms{43.7}{0.5} & \cellcolor{tabcreamgray!72}\ms{65.8}{0.6} & \cellcolor{tabcreamgray!69}\ms{63.7}{0.4} & \cellcolor{tabcreamgray!66}\ms{60.8}{0.6} & \cellcolor{tabcreamgray!63}\ms{58.1}{0.5} & \cellcolor{tabcreamgray!58}\ms{53.4}{0.8} & \cellcolor{tabcreamgray!52}\ms{47.9}{0.6} & \cellcolor{tabcreamgray!45}\ms{41.1}{0.8} & \cellcolor{tabcreamgray!33}\ms{30.4}{0.4} & \cellcolor{tabcreamgray!15}\ms{14.1}{0.6} & \cellcolor{tabcreamgray!2}\ms{1.5}{0.1} \\
Qwen3.5 & \cellcolor{tabgreenfig!3}\ms{\phantom{0}3.8}{0.2} & \cellcolor{tabcreamgray!16}\ms{14.6}{0.3} & \cellcolor{tabcreamgray!11}\ms{10.1}{0.6} & \cellcolor{tabcreamgray!7}\ms{\phantom{0}6.6}{0.3} & \cellcolor{tabcreamgray!4}\ms{\phantom{0}3.8}{0.3} & \cellcolor{tabcreamgray!2}\ms{\phantom{0}2.0}{0.2} & \cellcolor{tabcreamgray!1}\ms{\phantom{0}1.0}{0.1} & \cellcolor{tabcreamgray!0}\ms{\phantom{0}0.4}{0.0} & \cellcolor{tabcreamgray!0}\ms{\phantom{0}0.1}{0.0} & \cellcolor{tabcreamgray!0}\ms{\phantom{0}0.0}{0.0} & \cellcolor{tabcreamgray!0}\ms{0.0}{0.0} \\
\bottomrule
\end{tabularx}
\setlength{\abovecaptionskip}{7pt}
\caption{\textbf{$\text{AP}_{\text{IoU2D}}\text{@}\tau$ per IoU2D threshold $\tau$} (average over the 9 datasets). $\text{AP}_{\text{IoU2D}}$ is the mean over $\tau$, and higher $\tau$ is stricter.}
\label{tab:ap_thresh_iou2d}
\end{table*}

\medskip
\begin{table*}[tp]
\centering
\scriptsize
\setlength{\tabcolsep}{0pt}
\renewcommand{\arraystretch}{1.15}  %
\begin{tabularx}{\textwidth}{@{}>{\raggedright\arraybackslash}p{\methodcol}@{\hspace{6pt}}C|*{9}{C}@{}}
\toprule
Method & \APncd~$\uparrow$ & HOT3D & HANDAL & HOPEv2 & YCB-V & LM & HB & T-LESS & ITODD & IPD \\
\midrule
Gemini 4 Carbon & \cellcolor{tabgreenfig!60}\ms{\textbf{61.8}}{0.4} & \cellcolor{tabcreamgray!68}\ms{\textbf{58.3}}{0.6} & \cellcolor{tabcreamgray!84}\ms{\textbf{72.0}}{0.9} & \cellcolor{tabcreamgray!57}\ms{\textbf{48.4}}{2.7} & \cellcolor{tabcreamgray!100}\ms{\textbf{85.6}}{1.3} & \cellcolor{tabcreamgray!83}\ms{\textbf{70.8}}{0.9} & \cellcolor{tabcreamgray!90}\ms{\textbf{76.9}}{2.3} & \cellcolor{tabcreamgray!74}\ms{\textbf{63.3}}{0.3} & \cellcolor{tabcreamgray!91}\ms{\textbf{78.3}}{1.9} & \cellcolor{tabcreamgray!3}\ms{2.7}{1.6} \\
Claude Opus 5.5 & \cellcolor{tabgreenfig!45}\ms{\underline{46.0}}{0.5} & \cellcolor{tabcreamgray!68}\ms{\underline{58.0}}{1.3} & \cellcolor{tabcreamgray!74}\ms{\underline{63.6}}{1.1} & \cellcolor{tabcreamgray!44}\ms{\underline{37.8}}{1.3} & \cellcolor{tabcreamgray!87}\ms{\underline{74.4}}{1.4} & \cellcolor{tabcreamgray!68}\ms{\underline{57.9}}{1.0} & \cellcolor{tabcreamgray!72}\ms{\underline{61.3}}{1.9} & \cellcolor{tabcreamgray!44}\ms{37.3}{0.5} & \cellcolor{tabcreamgray!23}\ms{\underline{19.9}}{2.1} & \cellcolor{tabcreamgray!5}\ms{\underline{4.0}}{0.9} \\
GPT-6 Astra & \cellcolor{tabgreenfig!41}\ms{42.1}{0.4} & \cellcolor{tabcreamgray!62}\ms{53.0}{1.0} & \cellcolor{tabcreamgray!71}\ms{60.7}{0.7} & \cellcolor{tabcreamgray!17}\ms{14.7}{1.4} & \cellcolor{tabcreamgray!78}\ms{66.7}{2.1} & \cellcolor{tabcreamgray!62}\ms{53.2}{0.4} & \cellcolor{tabcreamgray!71}\ms{61.0}{3.0} & \cellcolor{tabcreamgray!53}\ms{\underline{45.2}}{0.9} & \cellcolor{tabcreamgray!23}\ms{19.6}{2.3} & \cellcolor{tabcreamgray!6}\ms{\textbf{5.0}}{0.7} \\
\bottomrule
\end{tabularx}
\setlength{\abovecaptionskip}{7pt}
\caption{\textbf{3D object grounding results on the validation split (\APncd in \%).} Gemini~4 Carbon, Claude Opus~5.5, and GPT-6 Astra, evaluated as on the test split (Tab.~\ref{tab:results_ncd}). The overall score is the average of the per-dataset scores.}
\label{tab:val_results_ncd}
\end{table*}

\medskip
\begin{table*}[tp]
\centering
\scriptsize
\setlength{\tabcolsep}{0pt}
\renewcommand{\arraystretch}{1.15}  %
\begin{tabularx}{\textwidth}{@{}>{\raggedright\arraybackslash}p{\methodcol}@{\hspace{6pt}}C|*{9}{C}@{}}
\toprule
Method & \APthreeD~$\uparrow$ & HOT3D & HANDAL & HOPEv2 & YCB-V & LM & HB & T-LESS & ITODD & IPD \\
\midrule
Gemini 4 Carbon & \cellcolor{tabgreenfig!60}\ms{\textbf{14.1}}{0.4} & \cellcolor{tabcreamgray!25}\ms{\textbf{7.7}}{0.9} & \cellcolor{tabcreamgray!22}\ms{\textbf{6.7}}{0.3} & \cellcolor{tabcreamgray!18}\ms{\textbf{5.5}}{1.3} & \cellcolor{tabcreamgray!100}\ms{\textbf{31.1}}{4.8} & \cellcolor{tabcreamgray!59}\ms{\textbf{18.5}}{1.5} & \cellcolor{tabcreamgray!75}\ms{\textbf{23.3}}{1.7} & \cellcolor{tabcreamgray!42}\ms{\textbf{13.0}}{0.5} & \cellcolor{tabcreamgray!66}\ms{\textbf{20.5}}{1.1} & \cellcolor{tabcreamgray!1}\ms{\textbf{0.3}}{0.5} \\
Claude Opus 5.5 & \cellcolor{tabgreenfig!20}\ms{\phantom{0}\underline{4.7}}{0.1} & \cellcolor{tabcreamgray!18}\ms{\underline{5.5}}{0.2} & \cellcolor{tabcreamgray!9}\ms{\underline{2.8}}{0.3} & \cellcolor{tabcreamgray!5}\ms{\underline{1.6}}{0.3} & \cellcolor{tabcreamgray!46}\ms{\underline{14.3}}{0.4} & \cellcolor{tabcreamgray!24}\ms{\phantom{0}\underline{7.4}}{1.1} & \cellcolor{tabcreamgray!23}\ms{\phantom{0}\underline{7.0}}{1.0} & \cellcolor{tabcreamgray!9}\ms{\phantom{0}2.9}{0.4} & \cellcolor{tabcreamgray!2}\ms{\phantom{0}\underline{0.7}}{0.5} & \cellcolor{tabcreamgray!0}\ms{0.0}{0.0} \\
GPT-6 Astra & \cellcolor{tabgreenfig!12}\ms{\phantom{0}2.8}{0.2} & \cellcolor{tabcreamgray!11}\ms{3.4}{0.1} & \cellcolor{tabcreamgray!7}\ms{2.2}{0.2} & \cellcolor{tabcreamgray!0}\ms{0.1}{0.0} & \cellcolor{tabcreamgray!11}\ms{\phantom{0}3.5}{0.2} & \cellcolor{tabcreamgray!21}\ms{\phantom{0}6.5}{0.5} & \cellcolor{tabcreamgray!19}\ms{\phantom{0}5.8}{0.9} & \cellcolor{tabcreamgray!12}\ms{\phantom{0}\underline{3.6}}{0.8} & \cellcolor{tabcreamgray!2}\ms{\phantom{0}0.6}{0.1} & \cellcolor{tabcreamgray!0}\ms{0.0}{0.0} \\
\bottomrule
\end{tabularx}
\setlength{\abovecaptionskip}{7pt}
\caption{\textbf{3D object grounding results on the validation split (\APthreeD in \%).} Gemini~4 Carbon, Claude Opus~5.5, and GPT-6 Astra, evaluated as on the test split (Tab.~\ref{tab:results_3d}). The overall score is the average of the per-dataset scores.}
\label{tab:val_results_3d}
\end{table*}

\medskip
\begin{table*}[tp]
\centering
\scriptsize
\setlength{\tabcolsep}{0pt}
\renewcommand{\arraystretch}{1.15}  %
\begin{tabularx}{\textwidth}{@{}>{\raggedright\arraybackslash}p{\methodcol}@{\hspace{6pt}}C|*{9}{C}@{}}
\toprule
Method & \APtwoD~$\uparrow$ & HOT3D & HANDAL & HOPEv2 & YCB-V & LM & HB & T-LESS & ITODD & IPD \\
\midrule
Gemini 4 Carbon & \cellcolor{tabgreenfig!60}\ms{\textbf{73.8}}{0.2} & \cellcolor{tabcreamgray!94}\ms{\textbf{75.1}}{0.2} & \cellcolor{tabcreamgray!97}\ms{\textbf{77.3}}{0.4} & \cellcolor{tabcreamgray!76}\ms{60.6}{0.4} & \cellcolor{tabcreamgray!94}\ms{74.9}{2.0} & \cellcolor{tabcreamgray!84}\ms{\textbf{67.1}}{1.8} & \cellcolor{tabcreamgray!100}\ms{\textbf{80.1}}{0.8} & \cellcolor{tabcreamgray!89}\ms{\textbf{71.3}}{1.6} & \cellcolor{tabcreamgray!97}\ms{\textbf{77.6}}{1.2} & \cellcolor{tabcreamgray!100}\ms{\textbf{80.1}}{2.7} \\
Claude Opus 5.5 & \cellcolor{tabgreenfig!56}\ms{\underline{68.9}}{0.5} & \cellcolor{tabcreamgray!92}\ms{73.9}{0.5} & \cellcolor{tabcreamgray!93}\ms{74.1}{0.5} & \cellcolor{tabcreamgray!79}\ms{\underline{63.5}}{1.6} & \cellcolor{tabcreamgray!95}\ms{\underline{76.3}}{0.5} & \cellcolor{tabcreamgray!67}\ms{\underline{53.9}}{2.1} & \cellcolor{tabcreamgray!93}\ms{\underline{74.5}}{1.4} & \cellcolor{tabcreamgray!76}\ms{\underline{60.7}}{1.3} & \cellcolor{tabcreamgray!89}\ms{\underline{71.5}}{1.5} & \cellcolor{tabcreamgray!90}\ms{\underline{71.9}}{0.8} \\
GPT-6 Astra & \cellcolor{tabgreenfig!51}\ms{62.8}{0.4} & \cellcolor{tabcreamgray!93}\ms{\underline{74.6}}{0.4} & \cellcolor{tabcreamgray!93}\ms{\underline{74.3}}{0.9} & \cellcolor{tabcreamgray!84}\ms{\textbf{67.4}}{1.2} & \cellcolor{tabcreamgray!100}\ms{\textbf{79.9}}{1.7} & \cellcolor{tabcreamgray!55}\ms{44.3}{0.7} & \cellcolor{tabcreamgray!91}\ms{72.5}{1.9} & \cellcolor{tabcreamgray!66}\ms{52.5}{1.8} & \cellcolor{tabcreamgray!68}\ms{54.4}{1.6} & \cellcolor{tabcreamgray!56}\ms{44.8}{4.0} \\
\bottomrule
\end{tabularx}
\setlength{\abovecaptionskip}{7pt}
\caption{\textbf{2D object grounding results on the validation split (\APtwoD in \%).} Gemini~4 Carbon, Claude Opus~5.5, and GPT-6 Astra, evaluated as on the test split (Tab.~\ref{tab:results_2d}). The overall score is the average of the per-dataset scores.}
\label{tab:val_results_2d}
\end{table*}

\endgroup
\end{table*}

\section{Qualitative results}
\label{app:query_examples}

\label{app:qualitative_predictions}
Figs.~\ref{fig:qual1}--\ref{fig:qual6} (at the end) show validation samples from all 9 source datasets with their queries and ground-truth 3D boxes, together with the 3D boxes predicted by Gemini~4 Carbon, Claude Opus~5.5, and GPT-6 Astra, and Figs.~\ref{fig:qualtwod1}--\ref{fig:qualtwod3} show ground-truth and predicted 2D boxes on other samples, one row per dataset. The samples were picked to cover a range of outcomes rather than the typical one. In 3D, they include queries on which all three models succeed, on which only one does, on which all three find the target in the image but miss it in 3D, on which a model boxes a different object or nothing, and multi-target queries. In 2D, they include two per dataset on which all three models are accurate.

The most common 3D failure is the one the projections hide: the predicted box overlaps the target in the image but is off in distance and size, which moves and scales the box along the viewing rays and barely changes its projection. On the test split, the 2D boxes enclosing the projected predicted and ground-truth 3D boxes have an IoU of at least 0.5 for 88\%, 83\%, and 94\% of the queries (first run of Gemini~4 Carbon, Claude Opus~5.5, and GPT-6 Astra), yet \IoUthreeD stays below 0.25 for 72 to 89\% of them. The \NCD and \IoUthreeD values in the figures expose such misses, \eg for the ITODD query ``the top cap'', the projected box of Claude Opus~5.5 nearly coincides with the ground truth, but its \NCD is 4.0 and \IoUthreeD 0. Boxing a clearly different object (an IoU of the enclosing 2D boxes below 0.2) is rarer: 2.5\%, 2.4\%, and 1.6\% of the test queries, \eg GPT-6 Astra for the T-LESS query ``the widest of the socket holders on top of the book''. For the HB query ``all the objects placed on the blue surface'', Opus~5.5 and Astra box all five targets, but even their closest boxes have \IoUthreeD 0.00 and 0.04.

In 2D, the models localize the referred objects well: the \IoUtwoD between each target and its best-matching box, averaged per query, is at least 0.5 for 96.6\%, 96.2\%, and 96.6\% of the test queries. GPT-6 Astra, however, returns several boxes for 10.2\% of the single-target test queries (Gemini~4 Carbon and Claude Opus~5.5 for 0.9\% each), \eg five boxes for the IPD query ``the rightmost T-shaped bracket''.

\section{Source datasets and licenses}
\label{app:datasets}

Tab.~\ref{tab:datasets} summarizes the 9 source BOP datasets. The six everyday-object datasets show household items, toys, and tools, and the three industrial ones mostly textureless or metallic parts. Most targets are hand-sized objects 50 to 100\,cm from the camera (Fig.~\ref{fig:dataset_comparison}).

\begin{table}[!htbp]
\centering
\footnotesize
\setlength{\tabcolsep}{4.5pt}
\begin{tabular}{@{}llcccl@{}}
\toprule
 & & & \multicolumn{2}{c}{\textbf{Images}} & \\
\cmidrule(lr){4-5}
\textbf{Dataset} & \textbf{Domain} & \textbf{Objects} & \textbf{Val} & \textbf{Test} & \textbf{Modality} \\
\midrule
HOT3D~\citep{banerjee2024hot3d}       & Everyday & 33 & 300 & 300 & RGB/mono. \\
HANDAL~\citep{guo2023handal}          & Everyday & 40 & 300 & 300 & RGB \\
HOPEv2~\citep{tyree2022hope}          & Everyday & 28 & 200 & 200 & RGB \\
LM~\citep{hinterstoisser2012lm}       & Everyday & 14 & 100 & 100 & RGB \\
HB~\citep{kaskman2019homebreweddb}    & Everyday & 33 & 100 & 100 & RGB \\
YCB-V~\citep{xiang2018posecnn}        & Everyday & 21 & 50 & 50 & RGB \\
T-LESS~\citep{hodan2017tless}         & Industrial      & 30 & 150 & 150 & RGB \\
ITODD~\citep{drost2017itodd}          & Industrial      & 27 & 150 & 150 & Mono. \\
IPD~\citep{kalra2024ipd}              & Industrial      & 9 & 50 & 50 & Mono. \\
\midrule
Total & & 225 & 1{,}400 & 1{,}400 & \\
\bottomrule
\end{tabular}
\setlength{\abovecaptionskip}{7pt}
\caption{\textbf{Overview of the 9 BOP datasets used in BOP-Refer.} The columns give the type and number of objects, the number of validation and test images, and the image type. The object counts include only objects referred to by the queries (235 in total), and since some objects appear in more than one dataset, the total counts the 225 unique objects. For LM, we also use the annotations of LM-O~\citep{brachmann2014lmo}, which adds poses of more objects to the images of one of the LM scenes. Its queries are counted under LM.}
\label{tab:datasets}
\end{table}

\paragraph{Frame sampling.}
The images are drawn from the BOP test splits and, for HANDAL, HOPEv2, HB, ITODD, and IPD, also from the BOP validation splits. For each dataset, candidates are sampled at equal spacing over the images sorted by scene and image ID (only over the BOP test targets~\citep{hodan2020bop} for LM, T-LESS, and YCB-V, and only over images with at least two visible objects for HOT3D and ITODD), which yields at most one image per HOT3D clip. Validation and test images come from disjoint scenes, except for the single ITODD scene and the LM-O scene, which are split by image. The candidates were reduced to the counts of Tab.~\ref{tab:datasets} during query verification. Of all pairs of selected images from the same scene (excluding HOT3D), 91.5\% in the test split and 92.0\% in the validation split refer to different sets of objects.

\paragraph{Licenses.}
\label{app:licenses}
BOP-Refer reuses images, 3D object models, and 6DoF pose annotations from 9 publicly released BOP datasets, whose licenses are listed in Tab.~\ref{tab:dataset_licenses}. We respect each upstream license: BOP-Refer redistributes only the components permitted by the source license, and downstream users inherit the restrictions of each source.

\begin{table}[!htbp]
\centering
\footnotesize
\begin{tabular}{@{}lll@{}}
\toprule
\textbf{Dataset} & \textbf{License} & \textbf{Commercial use} \\
\midrule
HOT3D     & Custom agreement            & see agreement \\
HANDAL    & CC BY-SA 4.0                & yes \\
HOPEv2    & CC BY-SA 4.0                & yes \\
YCB-V     & MIT                         & yes \\
LM        & CC BY 4.0                   & yes \\
LM-O      & CC BY-SA 4.0                & yes \\
HB        & CC0 1.0                     & yes \\
T-LESS    & CC BY 4.0                   & yes \\
ITODD     & CC BY-NC-SA 4.0$^\dagger$   & no$^\dagger$ \\
IPD       & CC BY-SA 4.0$^\dagger$      & yes \\
\bottomrule
\end{tabular}
\setlength{\abovecaptionskip}{7pt}
\caption{\textbf{Licenses of the 9 source BOP datasets reused by BOP-Refer}, as declared on the BOP website. HOT3D is governed by the HOT3D Dataset License Agreement. LM-O provides the images and pose annotations of one LM scene, counted under LM. $^\dagger$A custom license is available (for IPD on request). Commercial use of ITODD is not permitted under its standard license.}
\label{tab:dataset_licenses}
\end{table}

\paragraph{BOP-Refer license.}
The new contributions of BOP-Refer (the queries, the derived ground truth, and the toolkit) are released under permissive licenses: the toolkit code under MIT, and the queries and derived annotations under CC BY 4.0, except where derived from a source dataset whose license imposes share-alike or non-commercial restrictions, in which case the upstream license applies to that subset.

\section{Ground-truth bounding boxes}
\label{sec:gt}
\label{sec:model_bbox}

\paragraph{3D bounding boxes.} Ground-truth 3D bounding boxes are constructed from the 3D object models and 6DoF pose annotations of the BOP datasets. For each object instance, the ground-truth pose $(\mathbf{R}_P, \mathbf{t}_P) \in SE(3)$ is the rigid transform from the model frame $M$ to the camera frame $C$.

We first fit a canonical oriented bounding box to each 3D model (shared by all its instances), parameterized as $(\mathbf{R}_0, \mathbf{t}_0, \mathbf{s})$, where $(\mathbf{R}_0, \mathbf{t}_0)$ is the rigid transform from the box frame $B$ to the model frame $M$ and $\mathbf{s} \in \mathbb{R}^3_+$ are the full extents along the box axes. Then, for each object instance, we compose this canonical box with the ground-truth pose to obtain the 3D bounding box in the camera frame, $b_{\mathrm{3D}} = (\mathbf{R}, \mathbf{t}, \mathbf{s})$, with $\mathbf{R} = \mathbf{R}_P \mathbf{R}_0$ and ${\mathbf{t} = \mathbf{R}_P \mathbf{t}_0 + \mathbf{t}_P}$.

\paragraph{Canonical 3D bounding boxes.} For each 3D object model $\mathcal{M}$, the canonical 3D bounding box is the tightest oriented 3D box whose axes follow the model's rotational and reflective symmetries (otherwise the dataset's up direction, with the remaining axes minimizing the cross-sectional area), so that predictions differing from the ground truth only by a symmetry are scored as equivalent under \IoUthreeD. Its extents $\mathbf{s}$ are tightened to the model (code in the BOP-Refer toolkit).

\paragraph{2D bounding boxes.}
\label{sec:gt_2d}

The 2D ground-truth bounding box of an object instance is the tightest axis-aligned box around the projections $\pi(\mathbf{K}, \mathbf{R}_P \mathbf{v} + \mathbf{t}_P)$ of all mesh vertices $\mathbf{v}$ of its 3D object model in the ground-truth pose, where $\pi$ denotes perspective projection with the camera intrinsic matrix $\mathbf{K}$. The box is amodal, covering occluded parts, and truncated at the image borders, as in Omni3D~\citep{brazil2023omni3d}.

\section{Query generation and verification}
\label{app:query_prompts}
This section documents the query generation of Sec.~\ref{sec:queries} and its automatic and manual verification.

\subsection{Query generation}

\paragraph{Scene graph.}
For each selected image we build a scene graph that summarizes everything the ground-truth annotations tell us about the annotated objects: their class, normalized 2D box, distance from the camera, visible fraction, apparent and physical size ranks, a coarse position description, and an explicit list of pairwise spatial relations with qualitative margins. The graph is serialized as YAML and given to the query generator as text alongside the image. Spatial relations are computed from the 6DoF poses rather than being inferred by a model, and the generator is instructed to treat the graph as the sole source of truth for any spatial claim. An abbreviated example follows:

\begin{promptverb}
scene_graph:
  image_size: [1280, 960]
  num_annotated_objects: 3
  bbox_format: [x_min, y_min, x_max, y_max] normalized
               to 0-1 relative to image_size
  note: there may be other visible objects in the scene
        that are not annotated below

objects:
  - obj_id: 1
    class: "ceramic coffee mug"
    bbox_norm: [0.1406, 0.2917, 0.3281, 0.6042]
    depth_m: 0.45
    visibility: 0.92
    apparent_size_rank: 2
    physical_size_rank: 3
    position_description: "left side"
    relations:
      - [left-of, 2, large_margin]
      - [in-front-of, 2, large_margin]
      - [left-of, 3, large_margin]
      - [in-front-of, 3, large_margin]
      - [nearest-to, 2]
      - [farthest-from, 3]

  - obj_id: 2
    ...
\end{promptverb}

\paragraph{Object appearance descriptions.}
The scene graph carries geometry but no surface detail, and a single image often hides the distinguishing features of an object. We therefore describe every 3D object model once, in advance, independently of any image: each model is rendered from 8 canonical viewpoints distributed around it, and the same two VLMs used for query generation (GPT-5.4 and Gemini~3.1 Pro) are asked for a short name and a 3 to 4 sentence description of its identity, colors and appearance, shape, and typical use. The generators receive these descriptions alongside the scene graph. The prompt, without its example answer, is:

\begin{promptverb}
You are looking at 8 rendered views of a single 3D
object from a robotics / household dataset.

Please provide the following as a JSON object (no
markdown, just raw JSON):
{
  "name": "<concise object name, 2-5 words>",
  "description": "<3-4 sentence description covering
   what the object is, its color(s) and visual
   appearance, its overall geometric shape, and what it
   is typically used for. Be specific and factual based
   on what you see.>"
}


If the object appears gray or untextured, describe the
geometry and likely identity based on shape alone. If
unsure, give your best guess.
\end{promptverb}

\paragraph{Query generation prompt.}
The generator receives the image, the camera intrinsics, the scene graph, and the object descriptions, and is asked for 5 candidate queries spanning different referring strategies. It is told explicitly that the evaluated model will see only the image, the intrinsics, and the query. Abbreviated template:

\begin{promptverb}
You are generating text queries for a vision benchmark
that evaluates language-grounded object localization.
Each query will be paired with the attached image and
camera intrinsics. An evaluated model will receive ONLY
the image, intrinsics, and your query -- not the scene
graph or object descriptions below.

[Task] Select one or more annotated objects and
generate exactly
5 text queries, varying the strategy:
  - APPEARANCE: color, shape, texture, material
  - SPATIAL-RELATIONAL: position relative to other
    objects (must be grounded in scene graph)
  - COMPARATIVE: distinguish from similar objects
  - FUNCTIONAL: refer by likely use or category
  - MULTI-OBJECT: shared property across multiple
    targets

[Key rules]
  - Exactly one correct grounding per query
  - Natural language (not commands)
  - For harder queries, do NOT name the target directly
  - Minimal but sufficient -- fewest attributes to
    disambiguate
  - No pixel coordinates or object IDs; at most 30
    words
  - Spatial claims must correspond to explicit scene
    graph entries
  - Prefer queries testing 3D understanding (depth,
    occlusion, size)

[Difficulty scoring] 1-100 scale anchored at:
  1-20 (trivial), 21-40 (simple attribute/relation),
  41-60 (combined reasoning), 61-80 (fine-grained 3D),
  81-100 (multi-step, near-ambiguity resolution)

[Output] JSON array of 5 entries, each with:
target_object_ids,
query, strategy, difficulty, reasoning.
\end{promptverb}

\subsection{Automatic verification}
\label{app:verification}
Every candidate query is passed to Claude Opus~4.7, a model that took no part in query generation, together with the image, the scene graph, the object descriptions, the intended target objects, and the generator's strategy, difficulty, and reasoning. The verifier judges the query against the 14 numbered criteria of the prompt below and rejects it if \emph{any} criterion fails. Of the 21{,}190 candidate queries generated for the test split, 62.3\% passed. Among the 7{,}984 rejections, the two most often cited criteria are unambiguity (27.7\%), where the expression matches several objects, and spatial-anchor quality (25.2\%), where the query anchors the target to a landmark that is itself ambiguous, distant, or hallucinated. Abbreviated prompt:

\begin{promptverb}
You are a strict quality-assurance reviewer for a VLM
benchmark. You receive: an image, the scene graph (all
annotated objects with spatial relations), per-object
descriptions, and for each query: the text, target IDs,
strategy, difficulty, and generator's reasoning.

[Verification criteria -- ALL must pass for "Correct"]
  1. Target match: query refers to exactly the listed
     targets
  2. Unambiguity: no other visible object could
     plausibly match
  3. Factual accuracy: all mentioned attributes correct
     in image
  4. Spatial accuracy: spatial claims must appear in
     target's scene graph relations list (scene graph
     is sole source of truth)
  5. Reasoning validity: generator's reasoning must not
     make false claims
  6. Naturalness: reads like natural spoken language
  7. No over-description: fewest attributes necessary
  8. Difficulty-appropriate indirection: no direct
     naming for difficulty > 50
  9-12. No coordinates, no object counts in
     multi-target, no stacked comparatives, at most 30
     words
  13. Strict duplicate-class disambiguation check
  14. Spatial queries must use natural, unambiguous,
      nearby anchors

[Output] JSON array: [{"index", "label":
"Correct"|"Incorrect",
"reason": "..."}, ...]
\end{promptverb}

Tab.~\ref{tab:verification_passrates} breaks the pass rate down by target size and by source dataset. Small targets pass only slightly less often than large ones, so the filter does not disproportionately remove small objects, and every dataset remains well represented, with the industrial datasets of small textureless parts in dense clutter at the low end. At the object level, the same 217 objects appear both in passed and in failed queries, so verification excludes no object systematically.

\begin{table}[!htbp]
\centering
\footnotesize
\setlength{\tabcolsep}{6pt}
\begin{tabular}{@{}lrrr@{}}
\toprule
 & Pass & Total & Rate \\
\midrule
\multicolumn{4}{@{}l}{\textit{By target size (relative image area)}} \\
Small ($<$1\%)         & 2{,}777 & 4{,}842  & 57.4\% \\
Medium (1--5\%)        & 7{,}281 & 11{,}608 & 62.7\% \\
Large ($>$5\%)         & 3{,}148 & 4{,}740  & 66.4\% \\
\dashedmidrule{4}
\multicolumn{4}{@{}l}{\textit{By dataset}} \\
YCB-V  & 427     & 600     & 71.2\% \\
HB     & 874     & 1{,}250 & 69.9\% \\
HANDAL & 2{,}343 & 3{,}500 & 66.9\% \\
HOT3D  & 4{,}596 & 6{,}980 & 65.8\% \\
LM     & 793     & 1{,}250 & 63.4\% \\
T-LESS & 1{,}201 & 2{,}000 & 60.1\% \\
HOPEv2 & 1{,}430 & 2{,}460 & 58.1\% \\
IPD    & 564     & 1{,}150 & 49.0\% \\
ITODD  & 978     & 2{,}000 & 48.9\% \\
\bottomrule
\end{tabular}
\setlength{\abovecaptionskip}{7pt}
\caption{\textbf{Automatic verification pass rates}, over all 21{,}190 candidate test queries, broken down by the relative image area of the target (top, where a multi-target query is binned by its largest target) and by dataset (bottom). The LM row covers the LM and LM-O queries together, as everywhere else in the benchmark. Pass rates vary only mildly with target size ($57$--$66\%$), so the filter does not systematically discard queries about small objects. Every dataset contributes, with pass rates from $49\%$ (ITODD and IPD, with small textureless parts in clutter) to $71\%$ (YCB-V).}
\label{tab:verification_passrates}
\end{table}

\subsection{Manual verification}
\label{app:manual_verification}
The surviving candidates were inspected by 25 human reviewers in a web tool that shows each candidate under its image, with the targets highlighted by red boxes, and lets the reviewer accept the query, rewrite it, or flag the sample as unusable. Reviewers saw 3{,}347 test samples, where a sample is one image with its candidate queries. For 2{,}714 samples (81\%), the reviewers accepted at least one query as generated, and for 356 (11\%), they rewrote a query and kept the rewrite. The remaining 277 samples (8\%) were flagged unusable and dropped. A sample was flagged unusable only when no rewrite could produce a sound query, typically because the target is almost fully occluded, only a few pixels across, or too dark or blurred to identify.

A diversity-trimming step then capped each query specification at $1.3\times$ the mean occurrence, removing 242 over-represented queries. After the IPD images were reduced to 50, 2{,}730 verified queries over 1{,}400 unique images remained. One query per image was then chosen, balancing object coverage and including multi-target queries, which gives the final 1{,}400 test queries with 1{,}803 ground-truth boxes in total. The final selection covers the objects of each dataset nearly uniformly (the normalized entropy of the number of queries per object is 0.88 to 0.99 in every dataset of both splits) and spreads the images evenly over the scenes, except in LM and HOPEv2, where one scene contributes a larger share of the images. The validation split went through the same generation, verification, and selection.

\section{Evaluation metric details}
\label{app:metrics}

All three scores, \APncd, \APthreeD, and \APtwoD, share the protocol below and differ only in the box comparison function that defines matches, and in the thresholds at which it is applied: \APncd uses the symmetry-aware \NCD of Eq.~\eqref{eq:ncd} with $\delta \in \{0.2, 0.4, \ldots, 2.0\}$, \APthreeD uses the symmetry-aware \IoUthreeD of Eq.~\eqref{eq:iou3d} with $\tau \in \{0.05, 0.10, \ldots, 0.50\}$, and \APtwoD uses \IoUtwoD of axis-aligned rectangles with $\tau \in \{0.50, 0.55, \ldots, 0.95\}$ following COCO~\citep{lin2014coco}. \NCD is a distance, so a larger threshold is looser, whereas \IoUthreeD and \IoUtwoD are similarities, so a larger threshold is stricter.

\subsection{AP protocol}

\paragraph{Per-dataset, per-threshold AP.}
Write $e$ for the comparison function of the score being computed and $\Theta$ for its set of thresholds (ten for each of the three scores). For each dataset $D$ and each $\theta$, $\text{AP}_{D, \theta}$ is computed as follows:
\begin{enumerate}
    \item For each query $q \in D$, the model's predictions are sorted by descending confidence and truncated to the top 100.
    \item Predictions are greedily assigned, in confidence order, to the still-unmatched ground-truth box that $e$ ranks best among those already passing the threshold test, \ie the smallest $\NCD \leq \theta$ for \APncd and the highest $\IoUthreeD \geq \theta$ or $\IoUtwoD \geq \theta$ for the overlap scores. Each ground-truth box is matched to at most one prediction, and the assignment is recomputed independently for every $\theta$, following COCO~\citep{lin2014coco}. Assigned predictions are true positives, and unassigned ones false positives.
    \item All labeled predictions from $D$ are pooled and ranked by descending confidence, yielding one precision--recall curve. Predictions with equal confidence form a single operating point: within a tie, false positives are ranked before true positives, so every recall level inside the tie gets the precision of the whole tie, and the precision--recall curve does not depend on how tied predictions are ranked (within a query, ties are matched in file order). For the VLMs, whose confidences come from the box order in each answer (Supp.\ Sec.~\ref{app:output_parsing}), all first boxes form one operating point, all second boxes the next, and so on.
    \item The (monotonically enveloped) precision is averaged over 101 equally-spaced recall points $r \in \{0, 0.01, \ldots, 1.0\}$ to obtain $\text{AP}_{D, \theta}$.
\end{enumerate}

\paragraph{Aggregation.}
The per-dataset score is the mean over its thresholds, $\text{AP}_D = \tfrac{1}{|\Theta|}\sum_{\theta \in \Theta} \text{AP}_{D, \theta}$, and the headline score is the mean across the 9 datasets, $\text{AP} = \tfrac{1}{9}\sum_{D} \text{AP}_D$. Queries on LM-O images count under LM (Tab.~\ref{tab:datasets}).

\paragraph{Subset metrics.}
For breakdowns by query or target category (\eg single- vs.\ multi-target queries, visibility or size bins), the same protocol is applied to the subset of queries, and the headline subset score is the mean across datasets that have at least one query in the subset.

\subsection{Symmetry-aware Normalized Corner Distance}
\label{app:ncd}

\paragraph{Corner labelings.}
A box $b = (\mathbf{R}, \mathbf{t}, \mathbf{s})$ with rotation $\mathbf{R}$, center $\mathbf{t}$, and extents $\mathbf{s}$ has the corners $\mathbf{c}_{\boldsymbol{\sigma}}(b) = \mathbf{R}\,(\boldsymbol{\sigma} \odot \mathbf{s})/2 + \mathbf{t}$, indexed by the sign vectors $\boldsymbol{\sigma} \in \{-1, 1\}^3$. This parameterization is not unique. Let $\mathcal{O}$ be the group of the 24 rotations that map a cube onto itself, \ie the signed permutation matrices with determinant $1$. For every $\mathbf{P} \in \mathcal{O}$, the box $(\mathbf{R}\mathbf{P}, \mathbf{t}, |\mathbf{P}|^{\top}\mathbf{s})$, whose extents are permuted along with its axes, is the same box with relabeled corners, $\mathbf{c}_{\boldsymbol{\sigma}}(\mathbf{R}\mathbf{P}, \mathbf{t}, |\mathbf{P}|^{\top}\mathbf{s}) = \mathbf{c}_{\mathbf{P}\boldsymbol{\sigma}}(\mathbf{R}, \mathbf{t}, \mathbf{s})$, where $|\mathbf{P}|$ takes absolute values elementwise. The corner permutations in Eq.~\eqref{eq:ncd} are therefore $\Pi = \{\boldsymbol{\sigma} \mapsto \mathbf{P}\boldsymbol{\sigma} : \mathbf{P} \in \mathcal{O}\}$.

These 24 relabelings include the self-symmetries of the box, \ie the $\mathbf{P}$ that leave its extents in place (4 for three distinct extents, 8 for a square prism, 24 for a cube), and those that interchange distinct extents, which occur when a method orders the box axes differently from the ground truth. All 24 describe the same box, so the minimum over $\Pi$ needs no extent tolerance and never credits a different box.

\paragraph{What $\NCD = 0$ certifies.}
Since every relabeling maps the corner set of a box onto itself, a prediction that describes the ground-truth box in a different parameterization yields $\NCD = 0$, whereas a fixed corner correspondence would penalize it, \eg by $0.82$ for a cube turned by $180^\circ$ about an axis. Conversely, $\NCD = 0$ implies that the predicted box coincides with $\mathbf{S} \circ b_{\mathrm{3D}}$ for some $\mathbf{S} \in \mathcal{S}$, and thus $\IoUthreeD = 1$.

\paragraph{Orientation-only errors.}
Normalizing by the ground-truth box diagonal bounds orientation-only errors by a shape-independent constant: a prediction with the correct center and extents that differs from the ground truth only by a rotation has $\NCD \leq 1/\sqrt{2} \approx 0.71$. Over the four relabelings in $\mathcal{O}$ that keep the extents in place, every coordinate of a corner is negated by exactly two, so the squared distance of each corner averages to $d(b_{\mathrm{3D}})^2/2$, and the best relabeling has a root-mean-square, and hence mean, corner distance of at most $d(b_{\mathrm{3D}})/\sqrt{2}$. An \NCD above $1/\sqrt{2}$ therefore implies an error in translation or extents.

\paragraph{Relation to prior corner-distance metrics.}
Distance-based localization metrics have precedent in nuScenes~\citep{caesar2020nuscenes}, which thresholds the center distance on the ground plane instead of \IoUthreeD but, as \citet{brazil2023omni3d} note, thereby ignores object size and rotation, whereas \NCD is sensitive to all three. Corner-distance metrics for 3D boxes have also been proposed, most closely the Normalized Hungarian Distance (NHD)~\citep{bian2024catfree3d}, which, like our \NCD, measures corner-to-corner error normalized by the ground-truth box diagonal (we average NHD over the 8 corners, like \NCD). The two metrics resolve the corner-labeling ambiguity differently. NHD re-matches the 8 predicted and ground-truth corners by optimal assignment (the Hungarian algorithm) and thus admits all $8! = 40{,}320$ corner permutations, whereas \NCD admits only the 24 in $\Pi$, which correspond to rigid relabelings of the box. Without object symmetries, NHD therefore never exceeds \NCD, but the two are close in practice: over $4{,}000$ random poses of each of four box shapes, they coincide in $32\%$ to $63\%$ of the cases, with a median difference below $0.001$ and a maximum of $0.12$.

The substantive difference is that \NCD also minimizes over the object symmetries $\mathcal{S}$, as \IoUthreeD does, whereas NHD compares the boxes regardless of the object. For a cylinder whose $1 \times 1 \times 3$ box is turned by $45^\circ$ about the cylinder axis (Fig.~\ref{fig:metrics}d), NHD is $0.16$ although the turned box is an equally valid ground truth, whereas \NCD is $0$ and \IoUthreeD is $1$. \NCD is also related to pose errors that measure distances between object model points instead of box corners, such as ADD(-S)~\citep{hinterstoisser2012lm,xiang2018posecnn} and the MSSD of the BOP challenges~\citep{hodan2020bop}, which handle symmetries in the same way (see \citet{hodan2016evaluation} for a discussion of ambiguity-invariant pose errors).

\section{VLM prompts}
\label{app:prompts}

This section lists the prompts used to evaluate the VLMs, the output parsing, and the decoding parameters. All models receive the same 3D prompt, our own, which asks for 9-DoF boxes, whereas each model family receives its own zero-shot 2D prompt in its documented box convention. Line breaks within paragraphs were added for readability.

\subsection{3D grounding prompt}
\label{app:prompts_3d}
All models receive the 3D prompt below, with 5 text-only in-context examples (Sec.~\ref{app:prompts_fewshot}). It gives the image size and the camera intrinsics and asks for 9-DoF boxes in the OpenCV camera frame, in meters and degrees.

\begin{promptverb}
System:
You are an expert 3D spatial reasoning model.

User:
<image>
Image size: {W}x{H} px. Camera intrinsics [fx, fy, cx,
cy]: [{fx}, {fy}, {cx}, {cy}].

Given the provided image and camera intrinsics, predict
3D bounding boxes of the following objects: "{query}".

Output conventions:
(1) Report the 3D bounding boxes in the OpenCV camera
frame (X right, Y down, Z forward) as a JSON list:
[{"box_3d": [x_center, y_center, z_center, x_size,
y_size, z_size, roll, pitch, yaw], "label": "{query}"},
...].
(2) Center and size are in meters; roll/pitch/yaw are
in degrees.
(3) x, y, z: 3D box center in the camera coordinate
frame, in meters.
(4) Camera frame follows OpenCV convention (+x right,
+y down, +z forward from the camera).
(5) w, h, d: full box extents in meters along the
object's own axes.
(6) roll, pitch, yaw: rotation in degrees (extrinsic
Tait-Bryan XYZ). The rotation maps the object's
canonical axes into the camera frame (OpenCV).
(7) If no object matches the query, return [].
\end{promptverb}

\subsection{Few-shot in-context examples (3D task)}
\label{app:prompts_fewshot}
In the 3D prompt, the system message is followed by 5 alternating (user, assistant) turn pairs and then by a final user turn with the test query. Each example user turn is the same user turn filled in for another query, without an image, and each assistant turn is the ground-truth list of 9-DoF boxes of that query, in meters and degrees, under the key \texttt{box\_3d}. The examples therefore convey only the expected output schema and the camera-frame and unit conventions, not visual information. The examples are 5 queries from the validation split. Gemini Robotics ER~2 receives them as text in the single user turn instead of as separate turns. Without the examples, \APncd of the three ablated VLMs (Gemini~4 Carbon, Claude Opus~5.5, and GPT-6 Astra) drops by 1.4 to 5.5\,pp, and with 10 examples it changes by $-0.4$ to $+2.7$\,pp relative to 5 examples (Sec.~\ref{app:ablation_incontext_examples}).

\subsection{2D grounding prompts}
\label{app:prompts_2d}
Each model family receives its own system and user prompt, listed verbatim below. The user turn contains the image followed by the text, without camera intrinsics. Except for the GPT family, for which we found no documented box format and request absolute pixel coordinates, the prompts ask for boxes in each family's documented coordinate convention. Only the prompts for GPT and Claude ask for amodal boxes.

\paragraph{Gemini.} Used by Gemini~3.1 Pro, 3.8 Flash, 4 Argon, and 4 Carbon. Boxes as \texttt{[ymin, xmin, ymax, xmax]} normalized to $[0, 1000]$. The user prompt follows the object-detection example of Google's Gemini spatial-understanding notebook, including its label instruction:
\begin{promptverb}
System:
You are Gemini's visual grounding system. Output
bounding boxes in your native [ymin, xmin, ymax, xmax]
format, normalized to 0-1000.

User:
<image>
Detect the 2d bounding boxes of the {query} (with
"label" as topping description").
\end{promptverb}

\paragraph{Gemini Robotics ER~2.} Same box format as Gemini, with the system message prepended to the user text:
\begin{promptverb}
System:
You are Gemini Robotics ER, a helpful visual grounding
assistant.

User:
<image>
Return bounding boxes as a JSON array with labels
highlighting {query}. The format should be as follows:
[{"box_2d": [ymin, xmin, ymax, xmax], "label": }]
normalized to 0-1000. The values in box_2d must only
be integers.
\end{promptverb}

\paragraph{Gemma~4.} Same box format as Gemini:
\begin{promptverb}
System:
You are Gemma's visual grounding system. Output
bounding boxes in your native [ymin, xmin, ymax, xmax]
format, normalized to 0-1000.

User:
<image>
Detect {query}. Output only ```json".
\end{promptverb}

\paragraph{GPT.} Used by GPT-5.4, 5.5, 6 Astra, and 6.1 Sol. Boxes as \texttt{[x\_min, y\_min, x\_max, y\_max]} in pixels:
\begin{promptverb}
System:
You are GPT, a helpful visual grounding assistant.

User:
<image>
Return a tight AMODAL 2D bounding box for every
distinct object instance that matches {query}. AMODAL
means the box must cover the whole object, including
any occluded parts. Output bounding boxes in image
pixel coordinates. Output format should be a JSON list
with each entry of the format: {"label": "<short
name>", "bbox_2d": [x_min, y_min, x_max, y_max]}. If
nothing matches, return [].
\end{promptverb}

\paragraph{Claude.} Used by Claude Opus~4.6, 4.7, and~5.5. Boxes as \texttt{[x\_min, y\_min, x\_max, y\_max]} in absolute pixel coordinates, with the image resolution stated in the user turn:
\begin{promptverb}
System:
You are Claude, an expert visual grounding assistant.
Output bounding boxes in your native [x_min, y_min,
x_max, y_max] format, in pixel coordinates.

User:
<image>
Image resolution: {W} x {H} pixels (width x height).
Referring expression: "{query}". Find EVERY distinct
object instance in the image that matches the
referring expression. For each match, return a tight
AMODAL 2D bounding box enclosing the WHOLE object
(including any parts that are occluded in this view).
Output a single JSON list with each entry as:
{"label": "<short name>", "bbox_2d": [x_min, y_min,
x_max, y_max]}. If nothing matches, return [].
\end{promptverb}

\paragraph{Qwen.} Used by Qwen3-VL, Qwen3.5, Qwen3.6, and Qwen3.8 Flash Next. Boxes as \texttt{[x\_min, y\_min, x\_max, y\_max]} normalized to $[0, 1000]$, matching Qwen's public grounding format:
\begin{promptverb}
System:
You are Qwen, a helpful visual grounding assistant.

User:
<image>
locate every instance that belongs to the following
categories: {query}. Report bbox coordinates, in JSON
format.
\end{promptverb}

Because Claude is prompted in its documented box convention, and the boxes of Claude Opus~4.6 are mapped back from the API-downscaled image frame (Sec.~\ref{app:output_parsing}), the low 2D score of Claude Opus~4.6 in Tab.~\ref{tab:results_2d} is a genuine measurement rather than an artifact of a foreign output convention.

\subsection{Output parsing}
\label{app:output_parsing}
\label{app:postprocessing}
All responses go through a shared parser. It keeps the text after the last ``Final Answer:'' tag if there is one, strips markdown code fences, and parses the rest as JSON, falling back to the first balanced JSON array or object if that fails. Predictions that still fail to parse count as zero detections. For Gemini~4 Carbon and Gemini Robotics ER~2, calls that return no answer, typically because thinking used up the token budget, are re-sampled up to 5 times (ER~2 also when cut off). This yields an answer for about 22\% of Carbon's 3D test queries, and the roughly 6\% still without one after 5 re-samples count as misses, so its 3D scores are slightly underestimated. ER~2 needed this for at most one query per run, and the other models, which are not re-sampled, rarely return empty answers (GPT-5.5: about 5\%).

\paragraph{2D boxes.} They are converted to pixels from the convention each model is prompted in (Sec.~\ref{app:prompts_2d}). Boxes requested in absolute pixel coordinates (Claude and GPT models) are rescaled only when all their coordinates are at most 1.5, \ie normalized to $[0, 1]$. Boxes are clipped to the image. The Claude API downscales images to roughly 1.15 megapixels and at most 1{,}568 pixels on the long side before Opus~4.6 sees them, and it returns its 2D boxes in the downscaled frame, so we map them back to the original image. Claude Opus~4.7 and~5.5 return boxes in the original frame.

\paragraph{3D boxes.} The parser accepts the \texttt{box\_3d}, \texttt{bbox\_3d}, \texttt{box}, and \texttt{bbox} keys, also nested under keys such as \texttt{objects}, flattens boxes given as three nested triples, sets missing angles to zero, and reads six-value boxes as axis-aligned min/max corners. To catch unit errors, a box whose center coordinates are all below 20 in magnitude is read as meters, and otherwise as millimeters. Angles are read as degrees. Following the official Qwen 3D grounding cookbook~\citep{qwen2025qwen3vl}, the Qwen 3D boxes are read as metric boxes in the camera frame, since applying the Omni3D virtual-camera depth denormalization~\citep{brazil2023omni3d} instead degraded their scores. No other model receives a model-specific adjustment.

\paragraph{Confidences.} None of the prompts asks for a confidence, so we derive one from the order of the boxes in each answer: the $i$-th box gets the confidence $1 - 10^{-4}(i-1)$. A model's first box thus ranks before its second box, while boxes at the same position in different answers are tied (Supp.\ Sec.~\ref{app:metrics}). Every model lists its most reliable box first: compared with tying all boxes, ranking them by their order raises \APncd by 0.3 to 1.5 percentage points for every VLM, \APthreeD by at most 0.2, and \APtwoD by 0.1 to 2.2, except for GPT-6 Astra and GPT-6.1 Sol ($+6.6$ and $+7.5$), which often add lower-ranked candidate boxes.

\subsection{Decoding parameters}
\label{app:decoding}
The proprietary models run at medium reasoning effort (or thinking level). Qwen3.5, Qwen3.6, and Qwen3.8 Flash Next, served via NVIDIA, are run with thinking on (Qwen3.6 off in 3D: with thinking, 62\% of its answers hit the token cap), and Qwen3-VL (thinking off) and Gemma~4 with the OpenRouter defaults. The sampling temperature is $0.6$ for Gemma~4 and Qwen3-VL in 2D, $1$ for GPT-5.5, GPT-6 Astra, GPT-6.1 Sol, Claude Opus~4.6, 4.7, and~5.5, and Gemini Robotics ER~2, and $0.1$ for the other models. Responses are capped at 4{,}096 output tokens (8{,}000 for GPT-6.1 Sol, 16{,}384 for Qwen3.5 and Qwen3.8 Flash Next in 3D, 32{,}768 for Carbon and the other Qwen runs except Qwen3-VL in 2D, 65{,}536 for ER~2). Unparsable responses are not re-queried.

\newcommand{\omnithreedfig}{%
\begin{figure*}[t]
    \centering
    \includegraphics[width=\textwidth]{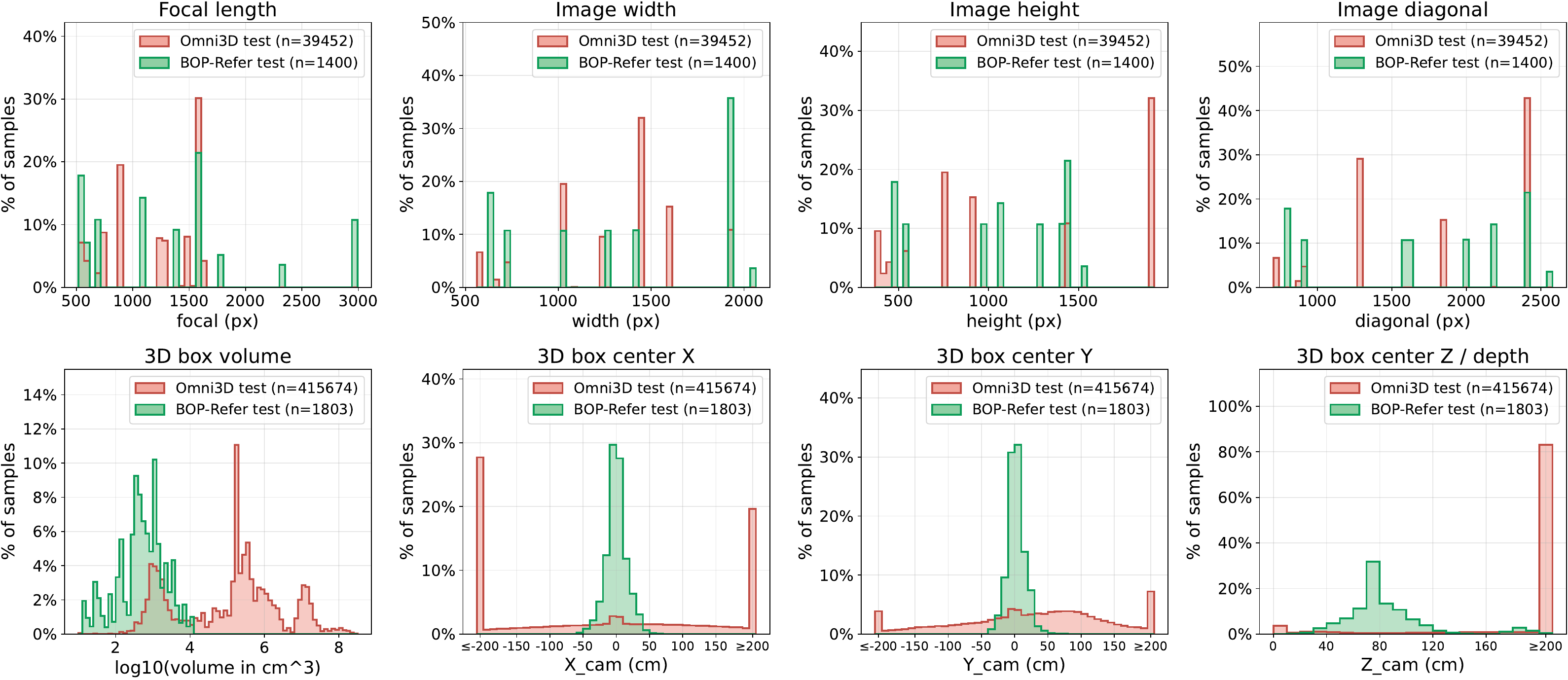}
    \setlength{\abovecaptionskip}{0pt}  %
\caption{%
        \textbf{BOP-Refer \vs Omni3D~\citep{brazil2023omni3d} data distribution.} BOP-Refer (1{,}803 object instances in the 1{,}400 test images of
        9 sub-datasets) targets the close-range, manipulable-object regime
        (mostly $10^2$--$10^{3.5}$\,cm$^3$ at 50--100\,cm depth) that is underrepresented
        in Omni3D (415{,}674 object instances in the 39{,}452 test images of 6 sub-datasets), which is dominated
        by furniture- and vehicle-scale objects at multi-meter depth.
        Top row: image-level pixel statistics. Bottom row: per-object 3D box
        statistics in camera-frame coordinates ($X$, $Y$, $Z$ in cm,
        volume in $\log_{10}\,\mathrm{cm}^3$); samples beyond
        $\pm 200$\,cm fall into the outermost columns. The BOP-Refer validation split (not shown) has nearly identical distributions (a Kolmogorov-Smirnov distance to the test split of at most $0.08$ for every statistic).
    }
    \label{fig:dataset_comparison}
\end{figure*}
}
\section{Other 3D grounding methods}
\label{sec:supp_other_methods}

This section describes how the other 3D grounding methods in Tabs.~\ref{tab:results_ncd}, \ref{tab:results_3d}, \ref{tab:results_ncd_oracle}, and~\ref{tab:results_3d_oracle} are implemented. They use public checkpoints without fine-tuning{} (mask back-projection also uses Gemini~4 Carbon) and run on every test image with its camera intrinsics. The rows fall into two groups. In Tabs.~\ref{tab:results_ncd} and~\ref{tab:results_3d}, the referred objects are localized from the text query alone: WildDet3D uses its own text-grounded 2D head, which is built on SAM~3~\citep{meta2025sam3}, while Boxer and SAM~3D use 2D boxes and masks predicted by SAM~3 (Sec.~\ref{sec:supp_predictors}), and mask back-projection uses the 2D boxes of Gemini~4 Carbon (Sec.~\ref{sec:supp_depthlift}). In Tabs.~\ref{tab:results_ncd_oracle} and~\ref{tab:results_3d_oracle}, each method instead receives one prompt per ground-truth 2D box of the query's target objects, which factors out 2D detection errors. WildDet3D and Boxer without depth are deterministic, whereas the Boxer rows with depth and SAM~3D vary slightly across runs because of the random sampling of points and seeds (Sec.~\ref{sec:supp_boxer}).

\subsection{WildDet3D}
\label{sec:supp_wilddet3d}

\paragraph{Model.} WildDet3D~\citep{huang2026wilddet3d} is a promptable monocular 3D detector. It takes an RGB image, the intrinsic matrix $\mathbf{K}$, an optional dense depth map, and a text, 2D-box, or 2D-point prompt, and returns a 2D box and a 9-DoF 3D box per detected object in the camera frame. Every image is fitted into a $1008 \times 1008$ canvas (the longer side resized to 1008 pixels, the shorter side zero-padded, and the intrinsics and box prompts transformed accordingly), and non-maximum suppression is applied at IoU $0.6$.

\paragraph{Prompting.} In Tabs.~\ref{tab:results_ncd} and~\ref{tab:results_3d}, WildDet3D runs in its native text-prompt mode: the benchmark query is passed as the input text, the internal text-grounded 2D head proposes the objects, and all 3D candidates with a score of at least $0.3$ (the upstream default) are kept as predictions. The four rows using ground-truth 2D detections run WildDet3D in box-prompt mode. Each ground-truth 2D box is passed as a separate prompt together with the generic text label \texttt{object}, which scored better than the upstream default label \texttt{geometric:\ object} and the full query. The score threshold is lowered to $0$ so that every prompt returns a candidate, and the best candidate of each prompt is kept.

\paragraph{Depth.} The rows differ in the depth map passed to WildDet3D: none (the monocular geometry head is used), depth predicted by Depth Anything~3 (DA3) or UniDepthV2~\citep{piccinelli2025unidepthv2} (Sec.~\ref{sec:supp_predictors}), or the real sensor depth provided with the images. Depth maps are resampled into the canvas with nearest-neighbour interpolation, which avoids blending foreground and background depths at object silhouettes. HOT3D, HANDAL, and IPD provide no dense depth, so the sensor-depth rows of WildDet3D and Boxer are reported as \texttt{--} on these datasets and without an overall score.

The depth map is passed in meters with depth input enabled, as the WildDet3D README specifies, and WildDet3D does use it: the median ratio of predicted to true object distance falls from 1.21--4.49 per dataset without depth to 1.09--1.89 with sensor depth. It still places objects too far, whereas Boxer, given the same depth maps, is within 3\% of the true distance on every dataset. This matches how WildDet3D is trained: 70\% of its training samples have no depth input, 20\% a patch-masked depth map, and only 10\% the full depth map~\citep[Sec.~2.1]{huang2026wilddet3d}.

\subsection{Boxer}
\label{sec:supp_boxer}

\paragraph{Model.} Boxer~\citep{kirillov2025boxer} detects objects with an open-vocabulary 2D detector and lifts each 2D detection to an oriented 3D box with a transformer called BoxerNet. We use the official release v1.0.0 (tag \texttt{ef6cad8}) with the public \texttt{boxernet\_hw960in2x6d768-c88128f8.ckpt} checkpoint and bypass the built-in 2D detector: BoxerNet receives either every SAM~3 box of the query with a score of at least $0.3$, the same boxes as SAM~3D (Tabs.~\ref{tab:results_ncd} and~\ref{tab:results_3d}, where queries for which SAM~3 returns no such box get no prediction) or each ground-truth 2D box as a separate prompt (Tabs.~\ref{tab:results_ncd_oracle} and~\ref{tab:results_3d_oracle}). The image is resized to $960 \times 960$ with correspondingly rescaled intrinsics, and the predictions are mapped back to the original image.

\paragraph{Single-image inference.} Boxer is designed for video with known per-frame camera poses and a semi-dense point cloud. Its README also covers single images given the camera intrinsics and the direction of gravity, and lists only images without this information as untested. BoxerNet regresses gravity-aligned boxes, so it needs the direction of gravity in the camera frame. Since many BOP-Refer views are tilted, we estimate the gravity direction of each image with GeoCalib~\citep{veicht2024geocalib}, as the Boxer authors suggest for inputs without known gravity, and set the camera pose passed to Boxer accordingly. The rows without depth pass an empty point cloud. In the rows with depth (\emph{+ DA3}, \emph{+ UniDepthV2}, and \emph{+ sensor depth}), the depth map, the same as for WildDet3D, is back-projected into a 3D point cloud with the camera intrinsics, and 10{,}000 points sampled from it at random, with a new seed per run, are passed as Boxer's point input.

\subsection{SAM 3D}
\label{sec:supp_sam3d}
\omnithreedfig  %
\paragraph{Model.} SAM~3D~\citep{meta2025sam3d} is a generative single-image 3D object reconstructor. From an RGB image and a binary object mask, it produces a textured mesh of the object in a canonical frame, with a 6DoF pose and an isotropic scale that place the mesh into the camera. We use the released \texttt{checkpoints/hf/pipeline.yaml} configuration. It uses neither the camera intrinsics nor a depth map, so it infers the distance and scale of the object from the image alone, without any metric reference.

\paragraph{Masks.} In Tabs.~\ref{tab:results_ncd} and~\ref{tab:results_3d}, every mask that SAM~3 returns for the query with a score of at least $0.3$ is lifted to a separate prediction, and queries without such a mask get no prediction. In Tabs.~\ref{tab:results_ncd_oracle} and~\ref{tab:results_3d_oracle}, each ground-truth 2D box is rasterized into a rectangular mask at the original image resolution.

\paragraph{From mesh to 3D box.} SAM~3D outputs no 3D box, so we take the axis-aligned box of the reconstructed mesh in its canonical frame: its extents are the per-axis ranges of the mesh vertices multiplied by the predicted scale, and it is placed into the camera by the predicted pose, converted from the PyTorch3D to the OpenCV camera convention by a $180^{\circ}$ rotation about the optical axis. Its center is the predicted translation, \ie the center of SAM~3D's canonical cube, which need not coincide with the center of the mesh. The canonical frame is gravity-aligned and unrelated to the oriented-box frame that BOP-Refer fits to each object model (Supp.\ Sec.~\ref{sec:gt}), so its axes and center can differ from the ground truth even for a perfect reconstruction. The SAM~3D scores therefore measure this canonical-box approximation rather than the quality of the reconstruction.

\paragraph{Box size.} With ground-truth 2D boxes, the boxes are 2.6 times too large (median ratio of the cube roots of the volumes), mostly because SAM~3D places them 1.9 times too far away, where an object must be larger to cover the same pixels. With each box's distance error divided out, they are still 1.5 times too large (1.1--1.2 on HB, LM, and HOT3D, and 1.15 for Boxer), so part of the excess likely comes from the reconstructed shape or from our axis-aligned box, which is loose for objects not aligned with the canonical axes.

\subsection{Mask back-projection}
\label{sec:supp_depthlift}

\paragraph{Pipeline.} This baseline combines VLM grounding, object masks, and monocular metric depth: Gemini~4 Carbon localizes the referred objects in 2D, and the mask of the object in each 2D box is back-projected to 3D with a predicted depth map. The 2D boxes come from one of Carbon's three default 2D test runs (1{,}800 boxes for the 1{,}400 queries), and each box yields one 3D box with the box's confidence, derived from its position in Carbon's answer (Sec.~\ref{app:postprocessing}). It uses no ground truth at test time and is run once.

\paragraph{Masks and depth.} SAM~3~\citep{meta2025sam3} computes the masks, prompted with each Carbon box rather than with the text query as for Boxer and SAM~3D (Sec.~\ref{sec:supp_predictors}), and the mask is eroded by 3 pixels. SAM~2.1~\citep{ravi2025sam2} masks give nearly the same scores (46.0\% \APncd and 7.3\% \APthreeD, \vs 45.7\% and 7.3\%). The depth map is predicted at full resolution by the H+ model of Depth Pro~2~\citep{hu2026depthpro2}, given the true focal length. We selected this estimator on the validation split among seven monocular depth estimators, including Depth Anything~3 and UniDepthV2, as the one whose median depth in the object mask is most often within 10\% of the depth of the ground-truth box center (45\% of the objects, \vs 41\% for DA3 and 29\% for UniDepthV2).

\paragraph{From points to 3D box.} The masked pixels are back-projected with the true intrinsics, points with a depth outside the 5th to 95th percentile of the mask are dropped, and the 3D box is the bounding box of the remaining points along their principal axes. The points cover only the visible surface of the object, so the box leaves out its occluded parts. Its \APncd exceeds that of Carbon's 3D boxes on HOT3D, LM, and IPD but is only 12.5\% and 0.0\% on T-LESS and ITODD, although its 2D boxes are accurate there (Tab.~\ref{tab:results_2d}).

\subsection{Off-the-shelf input predictors}
\label{sec:supp_predictors}
\paragraph{SAM~3 (2D boxes and masks for Boxer and SAM~3D).} We run SAM~3~\citep{meta2025sam3}, a text-promptable open-vocabulary segmenter, on each image with the benchmark query, and cache the returned masks, boxes, and confidence scores, keeping detections with a score of at least $0.3$, the threshold WildDet3D applies to its own text head. SAM~3 returns such a detection for 744 of the 1{,}400 test queries, and the other 656 queries count as misses for Boxer and SAM~3D in Tabs.~\ref{tab:results_ncd} and~\ref{tab:results_3d}. SAM~3 misses them because it expects short noun phrases rather than full referring expressions.

\paragraph{Depth Anything~3 and UniDepthV2.} For the + DA3 and + UniDepthV2 rows, we use the \texttt{depth-anything/\allowbreak DA3NESTED-GIANT-LARGE} checkpoint of Depth Anything~3~\citep{depthanything3} at a processing resolution of 504 pixels and the ViT-L/14 checkpoint \texttt{lpiccinelli/\allowbreak unidepth-v2-vitl14} of UniDepthV2~\citep{piccinelli2025unidepthv2}, both conditioned on the camera intrinsics of each image. Both predict metric depth, which we use directly, without scale or shift alignment, and the DA3 depth map is resized bilinearly to the original image resolution.

\section{BOP-Refer \vs Omni3D data distribution}
\label{app:omni3d_comparison}

Fig.~\ref{fig:dataset_comparison} compares the data distributions of BOP-Refer and Omni3D. The median BOP-Refer object has a volume of $528\,\mathrm{cm}^{3}$ (an 8\,cm cube), whereas the median Omni3D object has about $1.9\times10^{5}\,\mathrm{cm}^{3}$ (furniture and vehicles), roughly $350\times$ larger. Likewise, $83.1\%$ of the Omni3D test annotations lie more than $200$\,cm from the camera but only $0.3\%$ of the BOP-Refer ones, which concentrate at arm's length (median depth $76$\,cm). On the image side, Omni3D is dominated by a single focal-length and resolution mode (from Hypersim), whereas BOP-Refer spans the distinct sensors of its nine sources. Its close-up objects come mainly from Objectron~\citep{ahmadyan2021objectron}, which covers nine categories and has no referring expressions. The gap may also explain why VLMs whose 3D training data follows the conventions and source datasets of Omni3D, most notably Qwen3-VL~\citep{qwen2025qwen3vl}, score high on Omni3D yet transfer poorly to BOP-Refer (Sec.~\ref{sec:results}).

\paragraph{Scores of trained models on existing benchmarks.}
Models trained for an existing monocular 3D grounding benchmark score high on it. On Mono3DRefer~\citep{zhan2024mono3dvg}, a KITTI-derived benchmark with the same input-output structure as BOP-Refer, Mono3DVG-EnSD~\citep{li2025mono3dvgensd} reaches $69.5\%$ Acc@0.25. On Omni3D, LocateAnything3D~\citep{nvidia2025locateanything3d} reaches $38.9\%$ \APthreeD, and Qwen3-VL reaches $39.4\%$ mAP@0.15 on SUN-RGBD, which we reproduced. On BOP-Refer, Qwen3-VL reaches $1.9\%$ \APthreeDat{0.15} (Tab.~\ref{tab:ap_thresh_iou3d}).

\clearpage
\ifdefined\qualcard\else\newlength{\qualw}\setlength{\qualw}{0.14\textwidth}     %
\newlength{\qualih}\setlength{\qualih}{0.75\qualw}       %
\newlength{\qualgap}\setlength{\qualgap}{1.5pt}          %
\newlength{\qualpad}\setlength{\qualpad}{3pt}            %
\newlength{\qualrowgap}\setlength{\qualrowgap}{5pt}      %
\newlength{\qualqh}\newlength{\qualch}
\newsavebox{\qualqbox}
\newif\ifqualshowscores\qualshowscorestrue               %
\newif\ifqualshowds\qualshowdstrue                       %
\newcommand{\qualimgdir}{figures/qual_val_3d}
\newcommand{\qualq}[1]{\scriptsize\itshape\parbox[t]{\dimexpr\qualw-2\qualpad\relax}{\microtypesetup{expansion=false}\centering\strut #1\strut}}
\newcommand{\qualmeasure}[5]{\sbox{\qualqbox}{\qualq{#5}}%
  \ifdim\dimexpr\ht\qualqbox+\dp\qualqbox\relax>\qualqh\global\qualqh=\dimexpr\ht\qualqbox+\dp\qualqbox\relax\fi}
\newcommand{\qualtag}[2]{\ifqualshowscores%
  \node[anchor=south east,inner sep=0.9pt,fill=black,fill opacity=0.6,text opacity=1,text=white,font=\fontsize{4.5}{4.5}\selectfont]
    at (\qualw,{-(#1+1)*\qualih-#1*\qualgap}) {#2};\fi}
\newcommand{\qualcard}[5]{%
  \setlength{\qualch}{\dimexpr3\qualih+2\qualgap+\qualqh+2\qualpad\relax}%
  \begin{tikzpicture}[baseline={(0,-3\qualih-2\qualgap)}]
    \begin{scope}
      \clip[rounded corners=2.2pt] (0,0) rectangle (\qualw,-\qualch);
      \node[inner sep=0,anchor=north west] at (0,0) {\includegraphics[width=\qualw]{\qualimgdir/#1_carbon.jpg}};
      \node[inner sep=0,anchor=north west] at (0,{-\qualih-\qualgap}) {\includegraphics[width=\qualw]{\qualimgdir/#1_opus.jpg}};
      \node[inner sep=0,anchor=north west] at (0,{-2\qualih-2\qualgap}) {\includegraphics[width=\qualw]{\qualimgdir/#1_astra.jpg}};
      \qualtag{0}{#2}\qualtag{1}{#3}\qualtag{2}{#4}
    \end{scope}
    \draw[black,line width=0.4pt,rounded corners=2.2pt] (0.2pt,-0.2pt) rectangle ({\qualw-0.2pt},{-\qualch+0.2pt});
    \node[inner sep=0,anchor=north west] at (\qualpad,{-3\qualih-2\qualgap-\qualpad}) {\qualq{#5}};
  \end{tikzpicture}}
\newcommand{\qualmodel}[2]{\node[rotate=90,inner sep=0,font=\scriptsize] at (0,{-(#1+0.5)*\qualih-#1*\qualgap}) {#2};}
\newcommand{\quallabels}[1]{%
  \begin{tikzpicture}[baseline={(0,-3\qualih-2\qualgap)}]
    \path (0,0) (0,{-3\qualih-2\qualgap});
    \ifqualshowds\node[rotate=90,inner sep=0,font=\scriptsize\bfseries] at (-9pt,{-1.5\qualih-\qualgap}) {#1};\fi
    \qualmodel{0}{Gemini 4 Carbon}\qualmodel{1}{Claude Opus 5.5}\qualmodel{2}{GPT-6 Astra}
  \end{tikzpicture}}
\newcommand{\qualrow}[7]{\global\qualqh=0pt
  \qualmeasure#2\qualmeasure#3\qualmeasure#4\qualmeasure#5\qualmeasure#6\qualmeasure#7%
  \quallabels{#1} & \qualcard#2 & \qualcard#3 & \qualcard#4 & \qualcard#5 & \qualcard#6 & \qualcard#7 \\}
\fi

\begin{figure*}[tp!]
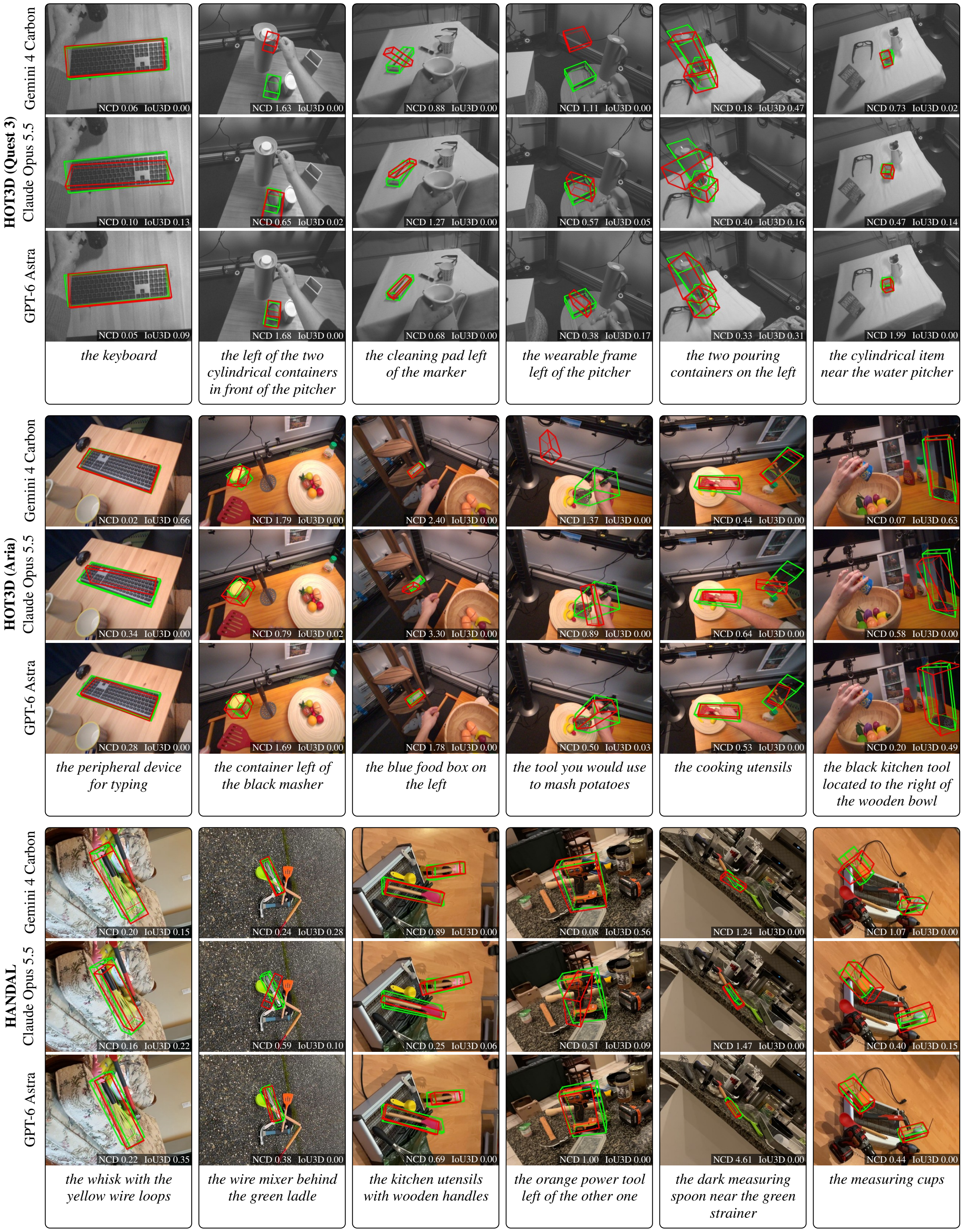

\centering
\renewcommand{\qualimgdir}{figures/qual_val_3d}
\setlength{\tabcolsep}{1.5pt}
\begin{tabular}{@{}r@{\hspace{4pt}}cccccc@{}}
\qualrow{HOT3D (Quest 3)}
  {{hot3d_quest_q00725}{\NCD 0.06\hspace{3pt}\IoUthreeD 0.00}{\NCD 0.10\hspace{3pt}\IoUthreeD 0.13}{\NCD 0.05\hspace{3pt}\IoUthreeD 0.09}{the keyboard}}  %
  {{hot3d_quest_q00611}{\NCD 1.63\hspace{3pt}\IoUthreeD 0.00}{\NCD 0.65\hspace{3pt}\IoUthreeD 0.02}{\NCD 1.68\hspace{3pt}\IoUthreeD 0.00}{the left of the two cylindrical containers in front of the pitcher}}  %
  {{hot3d_quest_q00673}{\NCD 0.88\hspace{3pt}\IoUthreeD 0.00}{\NCD 1.27\hspace{3pt}\IoUthreeD 0.00}{\NCD 0.68\hspace{3pt}\IoUthreeD 0.00}{the cleaning pad left of the marker}}  %
  {{hot3d_quest_q00657}{\NCD 1.11\hspace{3pt}\IoUthreeD 0.00}{\NCD 0.57\hspace{3pt}\IoUthreeD 0.05}{\NCD 0.38\hspace{3pt}\IoUthreeD 0.17}{the wearable frame left of the pitcher}}  %
  {{hot3d_quest_q00674}{\NCD 0.18\hspace{3pt}\IoUthreeD 0.47}{\NCD 0.40\hspace{3pt}\IoUthreeD 0.16}{\NCD 0.33\hspace{3pt}\IoUthreeD 0.31}{the two pouring containers on the left}}  %
  {{hot3d_quest_q00682}{\NCD 0.73\hspace{3pt}\IoUthreeD 0.02}{\NCD 0.47\hspace{3pt}\IoUthreeD 0.14}{\NCD 1.99\hspace{3pt}\IoUthreeD 0.00}{the cylindrical item near the water pitcher}}  %
\noalign{\vskip\qualrowgap}
\qualrow{HOT3D (Aria)}
  {{hot3d_aria_q00890}{\NCD 0.02\hspace{3pt}\IoUthreeD 0.66}{\NCD 0.34\hspace{3pt}\IoUthreeD 0.00}{\NCD 0.28\hspace{3pt}\IoUthreeD 0.00}{the peripheral device for typing}}  %
  {{hot3d_aria_q00821}{\NCD 1.79\hspace{3pt}\IoUthreeD 0.00}{\NCD 0.79\hspace{3pt}\IoUthreeD 0.02}{\NCD 1.69\hspace{3pt}\IoUthreeD 0.00}{the container left of the black masher}}  %
  {{hot3d_aria_q00780}{\NCD 2.40\hspace{3pt}\IoUthreeD 0.00}{\NCD 3.30\hspace{3pt}\IoUthreeD 0.00}{\NCD 1.78\hspace{3pt}\IoUthreeD 0.00}{the blue food box on the left}}  %
  {{hot3d_aria_q00841}{\NCD 1.37\hspace{3pt}\IoUthreeD 0.00}{\NCD 0.89\hspace{3pt}\IoUthreeD 0.00}{\NCD 0.50\hspace{3pt}\IoUthreeD 0.03}{the tool you would use to mash potatoes}}  %
  {{hot3d_aria_q00840}{\NCD 0.44\hspace{3pt}\IoUthreeD 0.00}{\NCD 0.64\hspace{3pt}\IoUthreeD 0.00}{\NCD 0.53\hspace{3pt}\IoUthreeD 0.00}{the cooking utensils}}  %
  {{hot3d_aria_q00806}{\NCD 0.07\hspace{3pt}\IoUthreeD 0.63}{\NCD 0.58\hspace{3pt}\IoUthreeD 0.00}{\NCD 0.20\hspace{3pt}\IoUthreeD 0.49}{the black kitchen tool located to the right of the wooden bowl}}  %
\noalign{\vskip\qualrowgap}
\qualrow{HANDAL}
  {{handal_q00189}{\NCD 0.20\hspace{3pt}\IoUthreeD 0.15}{\NCD 0.16\hspace{3pt}\IoUthreeD 0.22}{\NCD 0.22\hspace{3pt}\IoUthreeD 0.35}{the whisk with the yellow wire loops}}  %
  {{handal_q00089}{\NCD 0.24\hspace{3pt}\IoUthreeD 0.28}{\NCD 0.59\hspace{3pt}\IoUthreeD 0.10}{\NCD 0.38\hspace{3pt}\IoUthreeD 0.00}{the wire mixer behind the green ladle}}  %
  {{handal_q00270}{\NCD 0.89\hspace{3pt}\IoUthreeD 0.00}{\NCD 0.25\hspace{3pt}\IoUthreeD 0.06}{\NCD 0.69\hspace{3pt}\IoUthreeD 0.00}{the kitchen utensils with wooden handles}}  %
  {{handal_q00145}{\NCD 0.08\hspace{3pt}\IoUthreeD 0.56}{\NCD 0.51\hspace{3pt}\IoUthreeD 0.09}{\NCD 1.00\hspace{3pt}\IoUthreeD 0.00}{the orange power tool left of the other one}}  %
  {{handal_q00167}{\NCD 1.24\hspace{3pt}\IoUthreeD 0.00}{\NCD 1.47\hspace{3pt}\IoUthreeD 0.00}{\NCD 4.61\hspace{3pt}\IoUthreeD 0.00}{the dark measuring spoon near the green strainer}}  %
  {{handal_q00261}{\NCD 1.07\hspace{3pt}\IoUthreeD 0.00}{\NCD 0.40\hspace{3pt}\IoUthreeD 0.15}{\NCD 0.44\hspace{3pt}\IoUthreeD 0.00}{the measuring cups}}  %
\end{tabular}
\vspace{-3pt}
\caption{\textbf{3D predictions of three strong VLMs on BOP-Refer.} Each card shows a sample from the validation split (query at the bottom) and the 3D boxes predicted by Gemini~4 Carbon, Claude Opus~5.5, and GPT-6 Astra (top to bottom, red), together with the ground-truth boxes (green). The corner of each image gives the \NCD (lower is better; 1 means the box corners are off by one box diagonal on average) and the \IoUthreeD (higher is better) of the prediction. For multi-target queries, only the scores of the best prediction are shown, i.e., of the pair of predicted and ground-truth boxes with the lowest \NCD. Images are cropped around the boxes. Continued in Figs.~\ref{fig:qual2}--\ref{fig:qual6}.}
\label{fig:qual1}
\end{figure*}

\begin{figure*}[tp!]
\centering
\renewcommand{\qualimgdir}{figures/qual_val_3d}
\setlength{\tabcolsep}{1.5pt}
\begin{tabular}{@{}r@{\hspace{4pt}}cccccc@{}}
\qualrow{HANDAL}
  {{handal_q00043}{\NCD 0.11\hspace{3pt}\IoUthreeD 0.16}{\NCD 0.30\hspace{3pt}\IoUthreeD 0.09}{\NCD 0.11\hspace{3pt}\IoUthreeD 0.16}{the slotted utensils}}  %
  {{handal_q00080}{\NCD 0.23\hspace{3pt}\IoUthreeD 0.07}{\NCD 0.11\hspace{3pt}\IoUthreeD 0.14}{\NCD 0.25\hspace{3pt}\IoUthreeD 0.01}{the utensil for scraping batter from a bowl}}  %
  {{handal_q00289}{\NCD 0.10\hspace{3pt}\IoUthreeD 0.62}{\NCD 0.54\hspace{3pt}\IoUthreeD 0.00}{\NCD 0.99\hspace{3pt}\IoUthreeD 0.00}{the leftmost power drill}}  %
  {{handal_q00029}{\NCD 0.82\hspace{3pt}\IoUthreeD 0.00}{\NCD 1.56\hspace{3pt}\IoUthreeD 0.00}{\NCD 1.30\hspace{3pt}\IoUthreeD 0.00}{the deeper green spoon with a wooden handle}}  %
  {{handal_q00241}{\NCD 0.23\hspace{3pt}\IoUthreeD 0.32}{\NCD 0.43\hspace{3pt}\IoUthreeD 0.13}{\NCD 0.61\hspace{3pt}\IoUthreeD 0.00}{the kitchen whisks}}  %
  {{handal_q00229}{\NCD 0.13\hspace{3pt}\IoUthreeD 0.61}{\NCD 0.66\hspace{3pt}\IoUthreeD 0.03}{\NCD 1.00\hspace{3pt}\IoUthreeD 0.00}{the power tool located in front of the hammer}}  %
\noalign{\vskip\qualrowgap}
\qualrow{HOPEv2}
  {{hopev2_q00449}{\NCD 0.08\hspace{3pt}\IoUthreeD 0.73}{\NCD 0.92\hspace{3pt}\IoUthreeD 0.00}{\NCD 1.43\hspace{3pt}\IoUthreeD 0.00}{the barbecue sauce bottle behind the corn can}}  %
  {{hopev2_q00454}{\NCD 0.38\hspace{3pt}\IoUthreeD 0.00}{\NCD 1.33\hspace{3pt}\IoUthreeD 0.00}{\NCD 0.93\hspace{3pt}\IoUthreeD 0.00}{the partly hidden dessert-mix box behind the other one}}  %
  {{hopev2_q00506}{\NCD 1.07\hspace{3pt}\IoUthreeD 0.00}{\NCD 0.31\hspace{3pt}\IoUthreeD 0.25}{\NCD 0.63\hspace{3pt}\IoUthreeD 0.03}{the canned fruit to the right of the milk carton}}  %
  {{hopev2_q00596}{\NCD 0.75\hspace{3pt}\IoUthreeD 0.00}{\NCD 2.78\hspace{3pt}\IoUthreeD 0.00}{\NCD 0.84\hspace{3pt}\IoUthreeD 0.00}{the squeeze bottle furthest from the camera}}  %
  {{hopev2_q00519}{\NCD 2.24\hspace{3pt}\IoUthreeD 0.00}{\NCD 2.26\hspace{3pt}\IoUthreeD 0.00}{\NCD 1.12\hspace{3pt}\IoUthreeD 0.00}{the cylindrical food container to the right of the chocolate pudding box}}  %
  {{hopev2_q00413}{\NCD 0.12\hspace{3pt}\IoUthreeD 0.46}{\NCD 0.72\hspace{3pt}\IoUthreeD 0.05}{\NCD 3.07\hspace{3pt}\IoUthreeD 0.00}{the mayonnaise squeeze bottle that is adjacent to the canned peaches}}  %
\noalign{\vskip\qualrowgap}
\qualrow{HOPEv2}
  {{hopev2_q00440}{\NCD 0.31\hspace{3pt}\IoUthreeD 0.21}{\NCD 0.22\hspace{3pt}\IoUthreeD 0.44}{\NCD 0.64\hspace{3pt}\IoUthreeD 0.01}{the blue soup cans}}  %
  {{hopev2_q00443}{\NCD 0.15\hspace{3pt}\IoUthreeD 0.37}{\NCD 1.43\hspace{3pt}\IoUthreeD 0.00}{\NCD 1.80\hspace{3pt}\IoUthreeD 0.00}{the red squeeze bottle next to the small pudding box}}  %
  {{hopev2_q00417}{\NCD 1.71\hspace{3pt}\IoUthreeD 0.00}{\NCD 0.78\hspace{3pt}\IoUthreeD 0.01}{\NCD 1.04\hspace{3pt}\IoUthreeD 0.00}{the can with big yellow CORN text}}  %
  {{hopev2_q00432}{\NCD 0.43\hspace{3pt}\IoUthreeD 0.17}{\NCD 0.45\hspace{3pt}\IoUthreeD 0.11}{\NCD 1.35\hspace{3pt}\IoUthreeD 0.00}{the can of sliced fruit next to the yellow mustard bottle}}  %
  {{hopev2_q00412}{\NCD 0.26\hspace{3pt}\IoUthreeD 0.18}{\NCD 0.54\hspace{3pt}\IoUthreeD 0.09}{\NCD 1.03\hspace{3pt}\IoUthreeD 0.00}{the milk cartons}}  %
  {{hopev2_q00407}{\NCD 0.36\hspace{3pt}\IoUthreeD 0.07}{\NCD 0.22\hspace{3pt}\IoUthreeD 0.30}{\NCD 1.09\hspace{3pt}\IoUthreeD 0.00}{the popcorn box lying flat at the front left}}  %
\end{tabular}
\vspace{-3pt}
\caption{\textbf{3D predictions of three strong VLMs on BOP-Refer} (continued from Fig.~\ref{fig:qual1}; 2/6).}
\label{fig:qual2}
\end{figure*}

\begin{figure*}[tp!]
\centering
\renewcommand{\qualimgdir}{figures/qual_val_3d}
\setlength{\tabcolsep}{1.5pt}
\begin{tabular}{@{}r@{\hspace{4pt}}cccccc@{}}
\qualrow{YCB-V}
  {{ycbv_q01395}{\NCD 0.13\hspace{3pt}\IoUthreeD 0.50}{\NCD 0.15\hspace{3pt}\IoUthreeD 0.48}{\NCD 0.23\hspace{3pt}\IoUthreeD 0.22}{the small round can in front of the rectangular meat can}}  %
  {{ycbv_q01362}{\NCD 0.19\hspace{3pt}\IoUthreeD 0.58}{\NCD 0.42\hspace{3pt}\IoUthreeD 0.24}{\NCD 0.57\hspace{3pt}\IoUthreeD 0.08}{the blue cylindrical coffee tin}}  %
  {{ycbv_q01382}{\NCD 0.06\hspace{3pt}\IoUthreeD 0.61}{\NCD 0.61\hspace{3pt}\IoUthreeD 0.00}{\NCD 0.59\hspace{3pt}\IoUthreeD 0.00}{the red box in front of the brown box}}  %
  {{ycbv_q01374}{\NCD 0.13\hspace{3pt}\IoUthreeD 0.49}{\NCD 0.82\hspace{3pt}\IoUthreeD 0.00}{\NCD 0.52\hspace{3pt}\IoUthreeD 0.04}{the rectangular block with three holes on its top surface}}  %
  {{ycbv_q01353}{\NCD 0.51\hspace{3pt}\IoUthreeD 0.15}{\NCD 0.35\hspace{3pt}\IoUthreeD 0.30}{\NCD 1.29\hspace{3pt}\IoUthreeD 0.00}{the adjacent tools on the right}}  %
  {{ycbv_q01363}{\NCD 0.30\hspace{3pt}\IoUthreeD 0.08}{\NCD 0.68\hspace{3pt}\IoUthreeD 0.00}{\NCD 0.36\hspace{3pt}\IoUthreeD 0.00}{the bottle behind the wooden piece}}  %
\noalign{\vskip\qualrowgap}
\qualrow{YCB-V}
  {{ycbv_q01370}{\NCD 0.16\hspace{3pt}\IoUthreeD 0.47}{\NCD 0.34\hspace{3pt}\IoUthreeD 0.58}{\NCD 0.54\hspace{3pt}\IoUthreeD 0.01}{the blue containers}}  %
  {{ycbv_q01381}{\NCD 0.18\hspace{3pt}\IoUthreeD 0.42}{\NCD 0.21\hspace{3pt}\IoUthreeD 0.49}{\NCD 0.07\hspace{3pt}\IoUthreeD 0.78}{the soup can}}  %
  {{ycbv_q01356}{\NCD 0.41\hspace{3pt}\IoUthreeD 0.40}{\NCD 0.30\hspace{3pt}\IoUthreeD 0.27}{\NCD 0.37\hspace{3pt}\IoUthreeD 0.20}{the black gripper behind the drill}}  %
  {{ycbv_q01388}{\NCD 0.09\hspace{3pt}\IoUthreeD 0.51}{\NCD 0.32\hspace{3pt}\IoUthreeD 0.25}{\NCD 0.81\hspace{3pt}\IoUthreeD 0.00}{the sugar box}}  %
  {{ycbv_q01367}{\NCD 0.12\hspace{3pt}\IoUthreeD 0.74}{\NCD 0.31\hspace{3pt}\IoUthreeD 0.21}{\NCD 0.39\hspace{3pt}\IoUthreeD 0.13}{the cylindrical cans}}  %
  {{ycbv_q01394}{\NCD 0.07\hspace{3pt}\IoUthreeD 0.52}{\NCD 0.62\hspace{3pt}\IoUthreeD 0.00}{\NCD 0.63\hspace{3pt}\IoUthreeD 0.00}{the only battery-powered tool in the scene}}  %
\noalign{\vskip\qualrowgap}
\qualrow{LM}
  {{lm_q01188}{\NCD 0.14\hspace{3pt}\IoUthreeD 0.41}{\NCD 0.43\hspace{3pt}\IoUthreeD 0.29}{\NCD 0.48\hspace{3pt}\IoUthreeD 0.04}{the power tool just behind the bench vise}}  %
  {{lm_q01130}{\NCD 0.20\hspace{3pt}\IoUthreeD 0.37}{\NCD 0.77\hspace{3pt}\IoUthreeD 0.00}{\NCD 1.63\hspace{3pt}\IoUthreeD 0.00}{the tool in between the drill and duck}}  %
  {{lm_q01147}{\NCD 0.70\hspace{3pt}\IoUthreeD 0.03}{\NCD 0.53\hspace{3pt}\IoUthreeD 0.09}{\NCD 0.90\hspace{3pt}\IoUthreeD 0.00}{the office tool partially hidden by the red figurine}}  %
  {{lm_q01107}{\NCD 0.08\hspace{3pt}\IoUthreeD 0.78}{\NCD 0.34\hspace{3pt}\IoUthreeD 0.47}{\NCD 0.46\hspace{3pt}\IoUthreeD 0.25}{the object to the left of the yellow plastic cup}}  %
  {{lm_q01156}{\NCD 1.60\hspace{3pt}\IoUthreeD 0.00}{\NCD 1.96\hspace{3pt}\IoUthreeD 0.00}{\NCD 2.49\hspace{3pt}\IoUthreeD 0.00}{the container for holding eggs}}  %
  {{lm_q01154}{\NCD 0.79\hspace{3pt}\IoUthreeD 0.00}{\NCD 0.36\hspace{3pt}\IoUthreeD 0.17}{\NCD 0.96\hspace{3pt}\IoUthreeD 0.00}{the animal figurine behind the power drill}}  %
\end{tabular}
\vspace{-3pt}
\caption{\textbf{3D predictions of three strong VLMs on BOP-Refer} (continued from Fig.~\ref{fig:qual1}; 3/6).}
\label{fig:qual3}
\end{figure*}

\begin{figure*}[tp!]
\centering
\renewcommand{\qualimgdir}{figures/qual_val_3d}
\setlength{\tabcolsep}{1.5pt}
\begin{tabular}{@{}r@{\hspace{4pt}}cccccc@{}}
\qualrow{LM}
  {{lm_q01191}{\NCD 0.53\hspace{3pt}\IoUthreeD 0.07}{\NCD 0.47\hspace{3pt}\IoUthreeD 0.04}{\NCD 0.96\hspace{3pt}\IoUthreeD 0.00}{the small decorative animal figurines}}  %
  {{lm_q01109}{\NCD 0.17\hspace{3pt}\IoUthreeD 0.33}{\NCD 0.26\hspace{3pt}\IoUthreeD 0.55}{\NCD 0.65\hspace{3pt}\IoUthreeD 0.00}{the cardboard box located in the center of the board}}  %
  {{lm_q01122}{\NCD 0.34\hspace{3pt}\IoUthreeD 0.19}{\NCD 1.19\hspace{3pt}\IoUthreeD 0.00}{\NCD 0.37\hspace{3pt}\IoUthreeD 0.05}{the object just behind the duck}}  %
  {{lm_q01173}{\NCD 0.27\hspace{3pt}\IoUthreeD 0.32}{\NCD 0.76\hspace{3pt}\IoUthreeD 0.00}{\NCD 1.03\hspace{3pt}\IoUthreeD 0.00}{the item you would use to water a plant}}  %
  {{lm_q01132}{\NCD 0.83\hspace{3pt}\IoUthreeD 0.00}{\NCD 0.20\hspace{3pt}\IoUthreeD 0.41}{\NCD 0.50\hspace{3pt}\IoUthreeD 0.00}{the blue iron}}  %
  {{lm_q01158}{\NCD 0.36\hspace{3pt}\IoUthreeD 0.21}{\NCD 0.36\hspace{3pt}\IoUthreeD 0.34}{\NCD 0.74\hspace{3pt}\IoUthreeD 0.00}{the red figurine located behind the yellow rubber duck}}  %
\noalign{\vskip\qualrowgap}
\qualrow{HB}
  {{hb_q00320}{\NCD 0.08\hspace{3pt}\IoUthreeD 0.63}{\NCD 0.38\hspace{3pt}\IoUthreeD 0.01}{\NCD 0.38\hspace{3pt}\IoUthreeD 0.00}{the red toy car}}  %
  {{hb_q00365}{\NCD 0.17\hspace{3pt}\IoUthreeD 0.43}{\NCD 0.59\hspace{3pt}\IoUthreeD 0.10}{\NCD 0.57\hspace{3pt}\IoUthreeD 0.05}{the rectangular object located to the left of the brown ring}}  %
  {{hb_q00327}{\NCD 0.78\hspace{3pt}\IoUthreeD 0.00}{\NCD 0.57\hspace{3pt}\IoUthreeD 0.00}{\NCD 1.00\hspace{3pt}\IoUthreeD 0.00}{the smallest industrial component on the table}}  %
  {{hb_q00316}{\NCD 0.13\hspace{3pt}\IoUthreeD 0.57}{\NCD 0.73\hspace{3pt}\IoUthreeD 0.00}{\NCD 0.30\hspace{3pt}\IoUthreeD 0.04}{the object that is the closest to the camera}}  %
  {{hb_q00378}{\NCD 0.88\hspace{3pt}\IoUthreeD 0.00}{\NCD 0.65\hspace{3pt}\IoUthreeD 0.00}{\NCD 1.22\hspace{3pt}\IoUthreeD 0.00}{the figurine with tusks and a charcoal-colored body}}  %
  {{hb_q00353}{\NCD 0.39\hspace{3pt}\IoUthreeD 0.33}{\NCD 0.50\hspace{3pt}\IoUthreeD 0.00}{\NCD 0.59\hspace{3pt}\IoUthreeD 0.04}{all the objects placed on the blue surface}}  %
\noalign{\vskip\qualrowgap}
\qualrow{HB}
  {{hb_q00349}{\NCD 0.28\hspace{3pt}\IoUthreeD 0.53}{\NCD 0.42\hspace{3pt}\IoUthreeD 0.13}{\NCD 0.61\hspace{3pt}\IoUthreeD 0.05}{the two objects unoccluded in the front}}  %
  {{hb_q00380}{\NCD 0.25\hspace{3pt}\IoUthreeD 0.48}{\NCD 0.31\hspace{3pt}\IoUthreeD 0.16}{\NCD 0.68\hspace{3pt}\IoUthreeD 0.00}{the office tool used for creating holes in paper}}  %
  {{hb_q00373}{\NCD 1.85\hspace{3pt}\IoUthreeD 0.00}{\NCD 2.13\hspace{3pt}\IoUthreeD 0.00}{\NCD 1.66\hspace{3pt}\IoUthreeD 0.00}{the industrial component that is southeast of the central ring}}  %
  {{hb_q00370}{\NCD 0.86\hspace{3pt}\IoUthreeD 0.00}{\NCD 0.76\hspace{3pt}\IoUthreeD 0.00}{\NCD 0.52\hspace{3pt}\IoUthreeD 0.00}{all the objects on the white tabletop}}  %
  {{hb_q00313}{\NCD 0.19\hspace{3pt}\IoUthreeD 0.46}{\NCD 0.20\hspace{3pt}\IoUthreeD 0.39}{\NCD 0.18\hspace{3pt}\IoUthreeD 0.42}{all objects placed on the white table surface}}  %
  {{hb_q00388}{\NCD 0.08\hspace{3pt}\IoUthreeD 0.56}{\NCD 1.50\hspace{3pt}\IoUthreeD 0.00}{\NCD 0.40\hspace{3pt}\IoUthreeD 0.24}{the yellow object in the front}}  %
\end{tabular}
\vspace{-3pt}
\caption{\textbf{3D predictions of three strong VLMs on BOP-Refer} (continued from Fig.~\ref{fig:qual1}; 4/6).}
\label{fig:qual4}
\end{figure*}

\begin{figure*}[tp!]
\centering
\renewcommand{\qualimgdir}{figures/qual_val_3d}
\setlength{\tabcolsep}{1.5pt}
\begin{tabular}{@{}r@{\hspace{4pt}}cccccc@{}}
\qualrow{T-LESS}
  {{tless_q01334}{\NCD 0.18\hspace{3pt}\IoUthreeD 0.58}{\NCD 0.44\hspace{3pt}\IoUthreeD 0.13}{\NCD 0.17\hspace{3pt}\IoUthreeD 0.53}{the small ceramic insert nearest the flashlight}}  %
  {{tless_q01291}{\NCD 0.33\hspace{3pt}\IoUthreeD 0.20}{\NCD 1.45\hspace{3pt}\IoUthreeD 0.00}{\NCD 3.16\hspace{3pt}\IoUthreeD 0.00}{the front circular-flanged screw-in piece}}  %
  {{tless_q01269}{\NCD 0.85\hspace{3pt}\IoUthreeD 0.00}{\NCD 0.70\hspace{3pt}\IoUthreeD 0.01}{\NCD 0.88\hspace{3pt}\IoUthreeD 0.00}{the rectangular adapter standing on its shorter side}}  %
  {{tless_q01201}{\NCD 0.89\hspace{3pt}\IoUthreeD 0.00}{\NCD 0.97\hspace{3pt}\IoUthreeD 0.00}{\NCD 0.37\hspace{3pt}\IoUthreeD 0.07}{the front one of the two single-socket ceramic blocks}}  %
  {{tless_q01327}{\NCD 1.67\hspace{3pt}\IoUthreeD 0.00}{\NCD 4.72\hspace{3pt}\IoUthreeD 0.00}{\NCD 3.28\hspace{3pt}\IoUthreeD 0.00}{the tallest lamp holder on the book}}  %
  {{tless_q01241}{\NCD 0.19\hspace{3pt}\IoUthreeD 0.52}{\NCD 0.21\hspace{3pt}\IoUthreeD 0.60}{\NCD 3.34\hspace{3pt}\IoUthreeD 0.00}{the leftmost single-socket ceramic base}}  %
\noalign{\vskip\qualrowgap}
\qualrow{T-LESS}
  {{tless_q01305}{\NCD 0.74\hspace{3pt}\IoUthreeD 0.06}{\NCD 0.33\hspace{3pt}\IoUthreeD 0.30}{\NCD 0.36\hspace{3pt}\IoUthreeD 0.30}{the rounded three-way adapters on the left}}  %
  {{tless_q01252}{\NCD 0.21\hspace{3pt}\IoUthreeD 0.34}{\NCD 0.46\hspace{3pt}\IoUthreeD 0.14}{\NCD 0.72\hspace{3pt}\IoUthreeD 0.00}{the splitter that is on top of another splitter}}  %
  {{tless_q01206}{\NCD 0.32\hspace{3pt}\IoUthreeD 0.42}{\NCD 1.63\hspace{3pt}\IoUthreeD 0.00}{\NCD 0.40\hspace{3pt}\IoUthreeD 0.18}{the compact white ceramic piece right of the longer one}}  %
  {{tless_q01317}{\NCD 0.35\hspace{3pt}\IoUthreeD 0.18}{\NCD 2.03\hspace{3pt}\IoUthreeD 0.00}{\NCD 1.34\hspace{3pt}\IoUthreeD 0.00}{the widest of the socket holders on top of the book}}  %
  {{tless_q01240}{\NCD 0.32\hspace{3pt}\IoUthreeD 0.24}{\NCD 0.35\hspace{3pt}\IoUthreeD 0.18}{\NCD 0.96\hspace{3pt}\IoUthreeD 0.00}{the two matching ring-shaped ceramic pieces on the right}}  %
  {{tless_q01349}{\NCD 0.66\hspace{3pt}\IoUthreeD 0.03}{\NCD 5.51\hspace{3pt}\IoUthreeD 0.00}{\NCD 1.89\hspace{3pt}\IoUthreeD 0.00}{the smooth round-rimmed piece on the far right}}  %
\noalign{\vskip\qualrowgap}
\qualrow{ITODD}
  {{itodd_q01000}{\NCD 0.45\hspace{3pt}\IoUthreeD 0.05}{\NCD 4.03\hspace{3pt}\IoUthreeD 0.00}{\NCD 0.88\hspace{3pt}\IoUthreeD 0.00}{the top cap}}  %
  {{itodd_q01012}{\NCD 0.22\hspace{3pt}\IoUthreeD 0.27}{\NCD 3.78\hspace{3pt}\IoUthreeD 0.00}{\NCD 4.27\hspace{3pt}\IoUthreeD 0.00}{the screw at the top left}}  %
  {{itodd_q01008}{\NCD 0.09\hspace{3pt}\IoUthreeD 0.64}{\NCD 1.24\hspace{3pt}\IoUthreeD 0.00}{\NCD 0.79\hspace{3pt}\IoUthreeD 0.00}{the right shroud}}  %
  {{itodd_q00996}{\NCD 3.21\hspace{3pt}\IoUthreeD 0.00}{\NCD 0.84\hspace{3pt}\IoUthreeD 0.00}{\NCD 1.09\hspace{3pt}\IoUthreeD 0.00}{the matching pair of finned cooler blocks}}  %
  {{itodd_q01088}{\NCD 1.46\hspace{3pt}\IoUthreeD 0.00}{\NCD 0.76\hspace{3pt}\IoUthreeD 0.00}{\NCD 1.26\hspace{3pt}\IoUthreeD 0.00}{the carburetor below the other two}}  %
  {{itodd_q01014}{\NCD 0.20\hspace{3pt}\IoUthreeD 0.00}{\NCD 3.19\hspace{3pt}\IoUthreeD 0.00}{\NCD 1.44\hspace{3pt}\IoUthreeD 0.00}{the pair farthest apart horizontally}}  %
\end{tabular}
\vspace{-3pt}
\caption{\textbf{3D predictions of three strong VLMs on BOP-Refer} (continued from Fig.~\ref{fig:qual1}; 5/6).}
\label{fig:qual5}
\end{figure*}

\begin{figure*}[tp!]
\centering
\renewcommand{\qualimgdir}{figures/qual_val_3d}
\setlength{\tabcolsep}{1.5pt}
\begin{tabular}{@{}r@{\hspace{4pt}}cccccc@{}}
\qualrow{ITODD}
  {{itodd_q01025}{\NCD 0.32\hspace{3pt}\IoUthreeD 0.00}{\NCD 3.86\hspace{3pt}\IoUthreeD 0.00}{\NCD 2.16\hspace{3pt}\IoUthreeD 0.00}{the pair of triangular plates on the right}}  %
  {{itodd_q00956}{\NCD 0.16\hspace{3pt}\IoUthreeD 0.58}{\NCD 0.13\hspace{3pt}\IoUthreeD 0.60}{\NCD 0.05\hspace{3pt}\IoUthreeD 0.76}{the top-right breaker}}  %
  {{itodd_q01034}{\NCD 0.10\hspace{3pt}\IoUthreeD 0.52}{\NCD 1.00\hspace{3pt}\IoUthreeD 0.00}{\NCD 1.23\hspace{3pt}\IoUthreeD 0.00}{the top-left cartridge cap}}  %
  {{itodd_q01017}{\NCD 0.97\hspace{3pt}\IoUthreeD 0.00}{\NCD 1.03\hspace{3pt}\IoUthreeD 0.00}{\NCD 0.28\hspace{3pt}\IoUthreeD 0.44}{all the wheel rims}}  %
  {{itodd_q01046}{\NCD 0.17\hspace{3pt}\IoUthreeD 0.19}{\NCD 0.38\hspace{3pt}\IoUthreeD 0.00}{\NCD 1.78\hspace{3pt}\IoUthreeD 0.00}{the two ring-shaped parts on the left}}  %
  {{itodd_q01086}{\NCD 0.07\hspace{3pt}\IoUthreeD 0.73}{\NCD 2.63\hspace{3pt}\IoUthreeD 0.00}{\NCD 3.05\hspace{3pt}\IoUthreeD 0.00}{the rightmost box}}  %
\noalign{\vskip\qualrowgap}
\qualrow{IPD}
  {{ipd_q00934}{\NCD 0.35\hspace{3pt}\IoUthreeD 0.00}{\NCD 1.21\hspace{3pt}\IoUthreeD 0.00}{\NCD 1.25\hspace{3pt}\IoUthreeD 0.00}{the U-shaped clamp near the lower middle}}  %
  {{ipd_q00922}{no box}{\NCD 0.75\hspace{3pt}\IoUthreeD 0.00}{\NCD 2.53\hspace{3pt}\IoUthreeD 0.00}{the T-shaped bracket on the left}}  %
  {{ipd_q00904}{\NCD 3.37\hspace{3pt}\IoUthreeD 0.00}{\NCD 0.67\hspace{3pt}\IoUthreeD 0.00}{\NCD 0.82\hspace{3pt}\IoUthreeD 0.00}{the long rectangular piece with rows of round holes}}  %
  {{ipd_q00906}{\NCD 4.15\hspace{3pt}\IoUthreeD 0.00}{\NCD 4.39\hspace{3pt}\IoUthreeD 0.00}{\NCD 1.53\hspace{3pt}\IoUthreeD 0.00}{the horizontally oriented metal piece with a row of circular holes}}  %
  {{ipd_q00918}{\NCD 2.64\hspace{3pt}\IoUthreeD 0.00}{\NCD 4.49\hspace{3pt}\IoUthreeD 0.00}{\NCD 2.59\hspace{3pt}\IoUthreeD 0.00}{the metallic part with two parallel elongated holes in the bottom-left part of the table}}  %
  {{ipd_q00902}{\NCD 2.22\hspace{3pt}\IoUthreeD 0.00}{\NCD 1.51\hspace{3pt}\IoUthreeD 0.00}{\NCD 0.82\hspace{3pt}\IoUthreeD 0.00}{the elongated block with multiple visible round openings}}  %
\noalign{\vskip\qualrowgap}
\qualrow{IPD}
  {{ipd_q00912}{\NCD 0.48\hspace{3pt}\IoUthreeD 0.06}{\NCD 1.31\hspace{3pt}\IoUthreeD 0.00}{\NCD 1.63\hspace{3pt}\IoUthreeD 0.00}{the pair of handrail brackets on the left}}  %
  {{ipd_q00914}{\NCD 3.54\hspace{3pt}\IoUthreeD 0.00}{\NCD 0.47\hspace{3pt}\IoUthreeD 0.04}{\NCD 3.44\hspace{3pt}\IoUthreeD 0.00}{the five metal handrail mounting components on the table}}  %
  {{ipd_q00923}{\NCD 1.40\hspace{3pt}\IoUthreeD 0.00}{\NCD 1.95\hspace{3pt}\IoUthreeD 0.00}{\NCD 2.05\hspace{3pt}\IoUthreeD 0.00}{all the T-shaped pieces}}  %
  {{ipd_q00944}{no box}{\NCD 1.33\hspace{3pt}\IoUthreeD 0.00}{\NCD 2.01\hspace{3pt}\IoUthreeD 0.00}{the U-shaped pipe clamps}}  %
  {{ipd_q00941}{\NCD 2.10\hspace{3pt}\IoUthreeD 0.00}{\NCD 0.77\hspace{3pt}\IoUthreeD 0.00}{\NCD 2.28\hspace{3pt}\IoUthreeD 0.00}{all the industrial fasteners designed for securing pipes}}  %
  {{ipd_q00947}{\NCD 9.29\hspace{3pt}\IoUthreeD 0.00}{\NCD 5.73\hspace{3pt}\IoUthreeD 0.00}{\NCD 3.60\hspace{3pt}\IoUthreeD 0.00}{all the spur gears}}  %
\end{tabular}
\vspace{-3pt}
\caption{\textbf{3D predictions of three strong VLMs on BOP-Refer} (continued from Fig.~\ref{fig:qual1}; 6/6).}
\label{fig:qual6}
\end{figure*}

\ifdefined\qualcard\else\fi

\begin{figure*}[tp!]
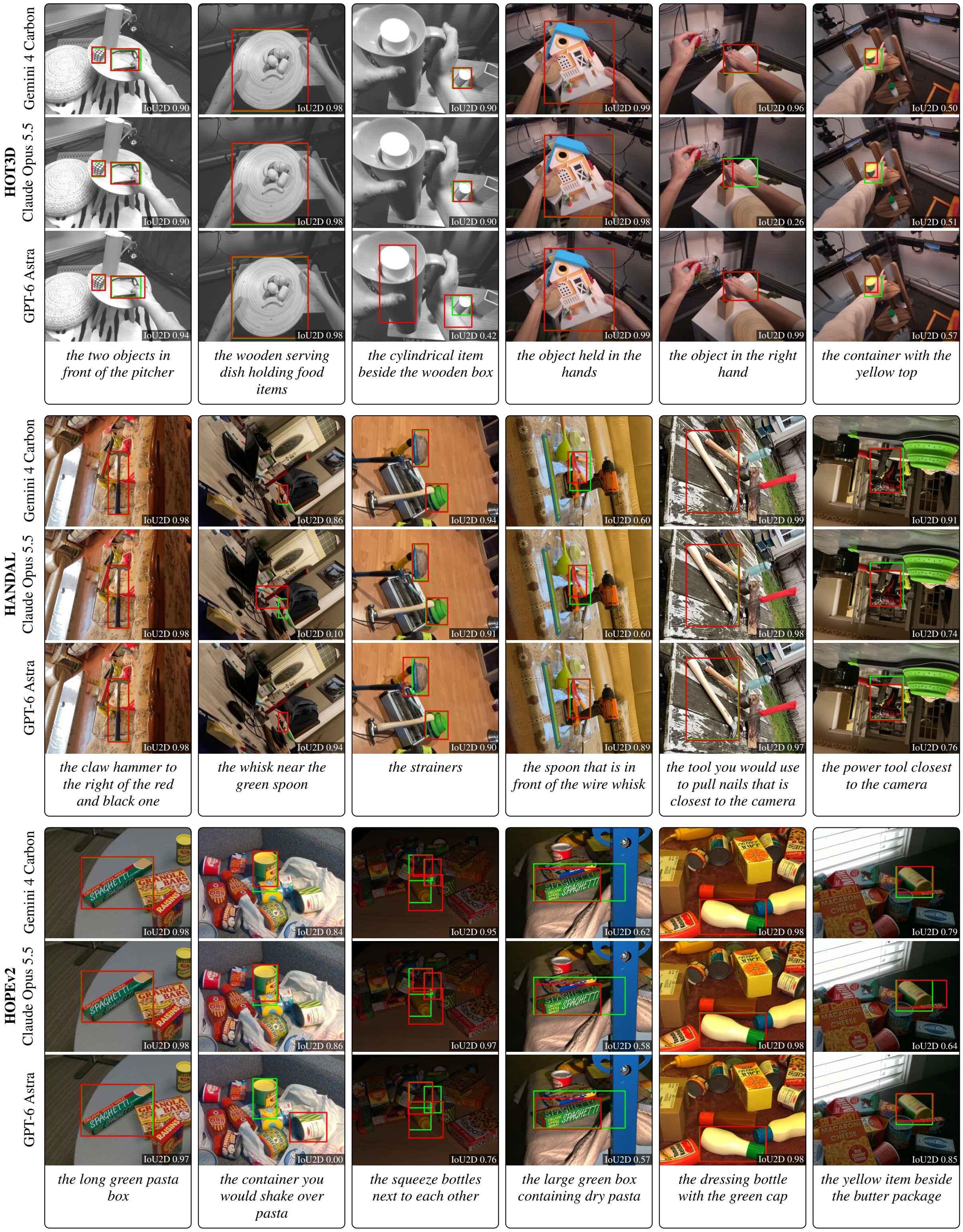

\centering
\renewcommand{\qualimgdir}{figures/qual_val_2d}
\setlength{\tabcolsep}{1.5pt}
\begin{tabular}{@{}r@{\hspace{4pt}}cccccc@{}}
\qualrow{HOT3D}
  {{hot3d_quest_q00626}{\IoUtwoD 0.90}{\IoUtwoD 0.90}{\IoUtwoD 0.94}{the two objects in front of the pitcher}}  %
  {{hot3d_quest_q00641}{\IoUtwoD 0.98}{\IoUtwoD 0.98}{\IoUtwoD 0.98}{the wooden serving dish holding food items}}  %
  {{hot3d_quest_q00607}{\IoUtwoD 0.90}{\IoUtwoD 0.90}{\IoUtwoD 0.42}{the cylindrical item beside the wooden box}}  %
  {{hot3d_aria_q00793}{\IoUtwoD 0.99}{\IoUtwoD 0.98}{\IoUtwoD 0.99}{the object held in the hands}}  %
  {{hot3d_aria_q00795}{\IoUtwoD 0.96}{\IoUtwoD 0.26}{\IoUtwoD 0.99}{the object in the right hand}}  %
  {{hot3d_aria_q00791}{\IoUtwoD 0.50}{\IoUtwoD 0.51}{\IoUtwoD 0.57}{the container with the yellow top}}  %
\noalign{\vskip\qualrowgap}
\qualrow{HANDAL}
  {{handal_q00193}{\IoUtwoD 0.98}{\IoUtwoD 0.98}{\IoUtwoD 0.98}{the claw hammer to the right of the red and black one}}  %
  {{handal_q00021}{\IoUtwoD 0.86}{\IoUtwoD 0.10}{\IoUtwoD 0.94}{the whisk near the green spoon}}  %
  {{handal_q00062}{\IoUtwoD 0.94}{\IoUtwoD 0.91}{\IoUtwoD 0.90}{the strainers}}  %
  {{handal_q00279}{\IoUtwoD 0.60}{\IoUtwoD 0.60}{\IoUtwoD 0.89}{the spoon that is in front of the wire whisk}}  %
  {{handal_q00072}{\IoUtwoD 0.99}{\IoUtwoD 0.98}{\IoUtwoD 0.97}{the tool you would use to pull nails that is closest to the camera}}  %
  {{handal_q00142}{\IoUtwoD 0.91}{\IoUtwoD 0.74}{\IoUtwoD 0.76}{the power tool closest to the camera}}  %
\noalign{\vskip\qualrowgap}
\qualrow{HOPEv2}
  {{hopev2_q00431}{\IoUtwoD 0.98}{\IoUtwoD 0.98}{\IoUtwoD 0.97}{the long green pasta box}}  %
  {{hopev2_q00401}{\IoUtwoD 0.84}{\IoUtwoD 0.86}{\IoUtwoD 0.00}{the container you would shake over pasta}}  %
  {{hopev2_q00458}{\IoUtwoD 0.95}{\IoUtwoD 0.97}{\IoUtwoD 0.76}{the squeeze bottles next to each other}}  %
  {{hopev2_q00525}{\IoUtwoD 0.62}{\IoUtwoD 0.58}{\IoUtwoD 0.57}{the large green box containing dry pasta}}  %
  {{hopev2_q00597}{\IoUtwoD 0.98}{\IoUtwoD 0.98}{\IoUtwoD 0.98}{the dressing bottle with the green cap}}  %
  {{hopev2_q00423}{\IoUtwoD 0.79}{\IoUtwoD 0.64}{\IoUtwoD 0.85}{the yellow item beside the butter package}}  %
\end{tabular}
\vspace{-3pt}
\caption{\textbf{2D predictions of three strong VLMs on BOP-Refer.} Each card shows a sample from the validation split (query at the bottom) and the 2D boxes predicted by Gemini~4 Carbon, Claude Opus~5.5, and GPT-6 Astra (top to bottom, red), together with the ground-truth boxes (green). The samples differ from those of the 3D predictions in Figs.~\ref{fig:qual1}--\ref{fig:qual6}. The corner of each image gives the \IoUtwoD of the prediction. For samples with multiple targets, only the score of the best prediction is shown, i.e., the highest \IoUtwoD among all pairs of predicted and ground-truth boxes. Images are cropped around the boxes. Continued in Figs.~\ref{fig:qualtwod2}--\ref{fig:qualtwod3}.}
\label{fig:qualtwod1}
\end{figure*}

\begin{figure*}[tp!]
\centering
\renewcommand{\qualimgdir}{figures/qual_val_2d}
\setlength{\tabcolsep}{1.5pt}
\begin{tabular}{@{}r@{\hspace{4pt}}cccccc@{}}
\qualrow{YCB-V}
  {{ycbv_q01369}{\IoUtwoD 0.99}{\IoUtwoD 0.97}{\IoUtwoD 0.99}{the blue object with a handle and pouring spout}}  %
  {{ycbv_q01360}{\IoUtwoD 0.94}{\IoUtwoD 0.95}{\IoUtwoD 0.89}{the can left of the red mug}}  %
  {{ycbv_q01383}{\IoUtwoD 0.95}{\IoUtwoD 0.92}{\IoUtwoD 0.96}{the rectangular cardboard boxes on the desk}}  %
  {{ycbv_q01371}{\IoUtwoD 0.65}{\IoUtwoD 0.66}{\IoUtwoD 0.70}{the yellow fruit}}  %
  {{ycbv_q01380}{\IoUtwoD 0.97}{\IoUtwoD 0.96}{\IoUtwoD 0.95}{the white bottle}}  %
  {{ycbv_q01358}{\IoUtwoD 0.93}{\IoUtwoD 0.93}{\IoUtwoD 0.95}{the cardboard box that is in front of the drill}}  %
\noalign{\vskip\qualrowgap}
\qualrow{LM}
  {{lm_q01137}{\IoUtwoD 0.93}{\IoUtwoD 0.96}{\IoUtwoD 0.96}{the bath toy near the hole punch}}  %
  {{lm_q01185}{\IoUtwoD 0.82}{\IoUtwoD 0.00}{\IoUtwoD 0.00}{the white cardboard carton}}  %
  {{lm_q01155}{\IoUtwoD 0.92}{\IoUtwoD 0.89}{\IoUtwoD 0.86}{the toy animal figures}}  %
  {{lm_q01121}{\IoUtwoD 0.90}{\IoUtwoD 0.43}{\IoUtwoD 0.43}{the white object to the right of the bench vise}}  %
  {{lm_q01114}{\IoUtwoD 0.89}{\IoUtwoD 0.90}{\IoUtwoD 0.88}{the white object in between the ape and cat}}  %
  {{lm_q01148}{\IoUtwoD 0.81}{\IoUtwoD 0.02}{\IoUtwoD 0.03}{the blue object adjacent to the bench vise}}  %
\noalign{\vskip\qualrowgap}
\qualrow{HB}
  {{hb_q00357}{\IoUtwoD 0.97}{\IoUtwoD 0.98}{\IoUtwoD 0.98}{the object with two control knobs on the front}}  %
  {{hb_q00396}{\IoUtwoD 0.94}{\IoUtwoD 0.67}{\IoUtwoD 0.72}{the black part behind the minion}}  %
  {{hb_q00318}{\IoUtwoD 0.87}{\IoUtwoD 0.86}{\IoUtwoD 0.91}{the electrical components near each other}}  %
  {{hb_q00387}{\IoUtwoD 0.57}{\IoUtwoD 0.79}{\IoUtwoD 0.81}{the character with goggles}}  %
  {{hb_q00303}{\IoUtwoD 0.96}{\IoUtwoD 0.96}{\IoUtwoD 0.95}{the tool used for boring holes and driving screws}}  %
  {{hb_q00355}{\IoUtwoD 0.96}{\IoUtwoD 0.94}{\IoUtwoD 0.86}{the dark part to the right of the white cylinder}}  %
\end{tabular}
\vspace{-3pt}
\caption{\textbf{2D predictions of three strong VLMs on BOP-Refer} (continued from Fig.~\ref{fig:qualtwod1}; 2/3).}
\label{fig:qualtwod2}
\end{figure*}

\begin{figure*}[tp!]
\centering
\renewcommand{\qualimgdir}{figures/qual_val_2d}
\setlength{\tabcolsep}{1.5pt}
\begin{tabular}{@{}r@{\hspace{4pt}}cccccc@{}}
\qualrow{T-LESS}
  {{tless_q01225}{\IoUtwoD 0.97}{\IoUtwoD 0.97}{\IoUtwoD 0.97}{the white switch box sitting on top of the larger ceramic block}}  %
  {{tless_q01285}{\IoUtwoD 0.07}{\IoUtwoD 0.91}{\IoUtwoD 0.97}{the smaller component you would mount on a rail}}  %
  {{tless_q01232}{\IoUtwoD 0.97}{\IoUtwoD 0.97}{\IoUtwoD 0.97}{the ceramic socket rings in front of the long three-socket block}}  %
  {{tless_q01335}{\IoUtwoD 0.62}{\IoUtwoD 0.64}{\IoUtwoD 0.70}{the white fuse with the black circle on top}}  %
  {{tless_q01272}{\IoUtwoD 0.96}{\IoUtwoD 0.99}{\IoUtwoD 0.96}{the rounded adapter to the right of the long three-outlet strip}}  %
  {{tless_q01316}{\IoUtwoD 0.00}{\IoUtwoD 0.00}{\IoUtwoD 0.00}{the socket holder with side ventilation slots}}  %
\noalign{\vskip\qualrowgap}
\qualrow{ITODD}
  {{itodd_q01039}{\IoUtwoD 0.99}{\IoUtwoD 0.99}{\IoUtwoD 0.98}{the metal corner support near the center}}  %
  {{itodd_q00957}{\IoUtwoD 0.97}{\IoUtwoD 0.10}{\IoUtwoD 0.88}{the breaker on the right}}  %
  {{itodd_q01062}{\IoUtwoD 0.92}{\IoUtwoD 0.90}{\IoUtwoD 0.99}{the two adjacent perforated plates in the upper half}}  %
  {{itodd_q01055}{\IoUtwoD 0.75}{\IoUtwoD 0.30}{\IoUtwoD 0.74}{the partially visible bracket screw on the right}}  %
  {{itodd_q00968}{\IoUtwoD 0.99}{\IoUtwoD 0.97}{\IoUtwoD 0.99}{the circular piece to the right of the other one}}  %
  {{itodd_q00984}{\IoUtwoD 0.94}{\IoUtwoD 0.93}{\IoUtwoD 0.03}{the rightmost washer}}  %
\noalign{\vskip\qualrowgap}
\qualrow{IPD}
  {{ipd_q00921}{\IoUtwoD 0.99}{\IoUtwoD 0.98}{\IoUtwoD 0.95}{the rightmost T-shaped bracket}}  %
  {{ipd_q00935}{\IoUtwoD 0.97}{\IoUtwoD 0.94}{\IoUtwoD 0.00}{the leftmost U-shaped clamp}}  %
  {{ipd_q00901}{\IoUtwoD 0.99}{\IoUtwoD 0.91}{\IoUtwoD 0.95}{the two perforated mounting blocks}}  %
  {{ipd_q00948}{\IoUtwoD 0.95}{\IoUtwoD 0.94}{\IoUtwoD 0.92}{the leftmost L-shaped piece}}  %
  {{ipd_q00928}{\IoUtwoD 0.97}{\IoUtwoD 0.98}{\IoUtwoD 0.97}{the two T-shaped brackets on the left side}}  %
  {{ipd_q00909}{\IoUtwoD 0.96}{\IoUtwoD 0.77}{\IoUtwoD 0.95}{the perforated metal piece on the right}}  %
\end{tabular}
\vspace{-3pt}
\caption{\textbf{2D predictions of three strong VLMs on BOP-Refer} (continued from Fig.~\ref{fig:qualtwod1}; 3/3).}
\label{fig:qualtwod3}
\end{figure*}

\end{document}